\documentclass{article} % For LaTeX2e
\usepackage{iclr2026_conference,times}

\usepackage{amsmath,amsfonts,bm}

\def\eqref#1{equation~\ref{#1}}
\def\1{\bm{1}}

\DeclareMathAlphabet{\mathsfit}{\encodingdefault}{\sfdefault}{m}{sl}
\SetMathAlphabet{\mathsfit}{bold}{\encodingdefault}{\sfdefault}{bx}{n}
\newcommand{\tens}[1]{\bm{\mathsfit{#1}}}

\newcommand{\etens}[1]{\mathsfit{#1}}

\usepackage{hyperref}
\usepackage{url}
\definecolor{darkblue}{rgb}{0, 0, 0.5}
\hypersetup{colorlinks=true, citecolor=darkblue, linkcolor=darkblue, urlcolor=darkblue}

\usepackage{custom}

\usepackage{microtype}

\usepackage{amssymb,bm,mathtools,color}

\usepackage[normalem]{ulem}
\usepackage{enumitem}

\usepackage{booktabs}
\usepackage{multirow}
\usepackage{tabularx}
\usepackage{multicol}
\usepackage{multirow}
\usepackage{colortbl}
\usepackage{hhline}
\usepackage{stfloats}

\newcolumntype{Y}{>{\centering\arraybackslash}X}

\usepackage{xcolor}
\definecolor{red}{rgb}{0.74,0.08,0.10}
\definecolor{green}{rgb}{0.26,0.49,0.18}
\definecolor{blue}{rgb}{0.22,0.53,0.75}

\usepackage{colortbl}
\definecolor{Gray}{gray}{0.9}
\definecolor{LightCyan}{rgb}{0.75,1,1}

\makeatletter
\newcommand\notsotiny{\@setfontsize\notsotiny\@viiipt\@ixpt}
\makeatother

\newcommand{\ie}{\emph{i.e.}, }
\newcommand{\eg}{\emph{e.g.}, }

\newcommand{\makename}[3][s]{%
  \expandafter\newcommand\csname #2\endcsname{#3\xspace}%
  \expandafter\newcommand\csname #2s\endcsname{#3#1\xspace}%
}
\usepackage[]{todonotes}

\usepackage[]{todonotes}  % use option "disable" to suppress notes
\makeatletter
\newcommand*\iftodonotes{\if@todonotes@disabled\expandafter\@secondoftwo\else\expandafter\@firstoftwo\fi}  % defines \iftodonotes{<true>}{<false>}, thanks to https://tex.stackexchange.com/questions/126559/conditional-based-on-packageoption
\makeatother

\usepackage{listings}

\lstdefinestyle{simple_lst_style}{
    columns=flexible,
    keywordstyle=\color{red},
    numberstyle=\color{gray},
    stringstyle=\color{green},
    basicstyle=\ttfamily\small,
    identifierstyle=\color{black},
    commentstyle=\color{blue},
    breakatwhitespace=false,
    breaklines=true,
    captionpos=b,
    keepspaces=false,
    numbersep=5pt,
    showspaces=false,
    showstringspaces=false,
    showtabs=false,
    tabsize=2,
    frame=single,
}

\lstdefinelanguage{Prompt}{
    morekeywords={Human, Computer},
    sensitive=true,
    morestring=[b]",
}

\usepackage{booktabs}
\usepackage{multirow}
\usepackage{graphicx}

\usepackage{pifont}
\usepackage{amsmath}
\usepackage{amssymb}
\usepackage{ifthen}
\usepackage{dsfont}

\usepackage{dcolumn}
\usepackage{siunitx}
\usepackage{makecell,rotating,array}

\usepackage{listings}
\usepackage{xcolor}
\usepackage{ltablex}
\usepackage{longtable}
\usepackage{multicol}
\usepackage{floatrow}

\usepackage{tikz}
\usetikzlibrary{bayesnet}
\usepackage[linguistics]{forest}

\usepackage[ruled,vlined]{algorithm2e}

\usepackage{caption}

\usepackage{cancel}

\usepackage{xspace}
\usepackage{comment}
\usepackage{color,soul}

\usepackage{float}

\usepackage[noabbrev,capitalize]{cleveref} % http://tug.ctan.org/tex-archive/macros/latex/contrib/cleveref/cleveref.pdf

\crefname{page}{page}{pages}
\crefname{footnote}{footnote}{footnotes}   % "footnote" is lowercased, overriding capitalize option
\crefname{equation}{equation}{equations}   % "equation" is lowercased, overriding capitalize option; note that \labelcref drops this word if you want to say something like "the divergence (3)"
\crefname{line}{line}{lines}               % "line" is lowercased, overriding capitalize option
\crefrangeformat{line}{lines #3#1#4--#5#2#6}
\crefname{lstlsting}{Listing}{Listings}   
\crefname{section}{\S}{\S\S}
\Crefname{section}{\S}{\S\S}    % must define start-of-sentence version explicitly since \S isn't a letter
\crefformat{section}{#2\S#1#3}  % remove space between section symbol and the number, and include \S in the hyperlink
\Crefformat{section}{#2\S#1#3}
\crefrangeformat{section}{\S\S#3#1#4--#5#2#6}
\Crefrangeformat{section}{\S\S#3#1#4--#5#2#6}
\crefmultiformat{section}{\S#2#1#3}{ and~\S#2#1#3}{, \S#2#1#3}{ and~\S#2#1#3}
\Crefmultiformat{section}{\S#2#1#3}{ and~\S#2#1#3}{, \S#2#1#3}{ and~\S#2#1#3}
\crefrangemultiformat{section}{\S\S#3#1#4--#5#2#6}{ and~\S\S#3#1#4--#5#2#6}{, \S\S#3#1#4--#5#2#6}{ and~\S\S#3#1#4--#5#2#6}
\Crefrangemultiformat{section}{\S\S#3#1#4--#5#2#6}{ and~\S\S#3#1#4--#5#2#6}{, \S\S#3#1#4--#5#2#6}{ and~\S\S#3#1#4--#5#2#6}

\title{LLM Microscope: What Model Internals Reveal About Answer Correctness and Context Utilization}

\author{
Jiarui Liu$^{*}$, Jivitesh Jain\thanks{Equal contribution}~, Mona Diab, Nishant Subramani \\
Carnege Mellon University \\
\texttt{\{jiaruil5, jivitesj, mdiab, nishant2\}@andrew.cmu.edu}
}

\iclrpreprinttrue
\begin{document}

\maketitle

\begin{abstract}
Although large language models (LLMs) have tremendous utility, trustworthiness is still a chief concern: models often generate incorrect information with high confidence. While contextual information can help guide generation, identifying when a query would benefit from retrieved context and assessing the effectiveness of that context remains challenging. In this work, we operationalize interpretability methods to ascertain whether we can predict the correctness of model outputs from the model’s activations alone. We also explore whether model internals contain signals about the efficacy of external context. We consider correct, incorrect, and irrelevant context and introduce metrics to distinguish amongst them. Experiments on six different models reveal that a simple classifier trained on intermediate layer activations of the first output token can predict output correctness with about 75\% accuracy, enabling early auditing. Our model-internals-based metric significantly outperforms prompting baselines at distinguishing between correct and incorrect context, guarding against inaccuracies introduced by polluted context. These findings offer a lens to better understand the underlying decision-making processes of LLMs.\footnote{Our code is publicly available at \url{https://github.com/jiarui-liu/LLM-Microscope}.}%~\nishant{put the code location (github repo in here either in the abstract or as a footnote i dont really have a preference)}
\end{abstract}
\section{Introduction}

Large language models (LLMs) have shown tremendous utility in many domains, including those that require accurately answering factual queries~\citep{bloom, grattafiori2024llama, groeneveld-etal-2024-olmo}. However, trustworthiness remains a chief concern: LLMs often generate convincing, but thoroughly incorrect and non-factual responses, termed \textit{hallucinations}~\citep{bang-etal-2023-multitask, survey_hallucination, guerreiro-etal-2023-hallucinations}. 

Recently, retrieval-augmented generation (RAG) has been proposed to mitigate this problem \citep{lewis2020retrieval}. 
Although RAG is effective, two challenges remain: confidence estimation to identify uncertain examples where an LLM requires external context and efficacy evaluation to score the utility of the retrieved external context.
Confidence estimation is challenging as LLMs are poorly calibrated: models often assign high probabilities to incorrect generations, making it difficult to detect when retrieval is needed~\citep{jiang-etal-2021-know,mielke-etal-2022-reducing, kadavath_language_2022, yin-etal-2023-large}. %~\nishant{needs more here} 
However, existing approaches either rely on fragile self-evaluation~\citep{yin2023large,chen2024insidellmsinternalstates} or focus on narrow tasks and require complex setups~\citep{azaria-mitchell-2023-internal,burns2024discoveringlatentknowledgelanguage}.

For context evaluation, although there exist methods to estimate efficacy of retrieved context such as SelfRAG~\citep{asai2023self}, they rely on fine-tuning models and prompting external models to gauge the utility of external context. A lightweight way to judge efficacy directly from model internals remains missing. %~\nishant{need more background here too; motivation isnt as comprehensive as it needs to be}
Instead, we look at these questions using our \emph{LLM microscope}, through the \emph{lens of mechanistic interpretability}, and study whether model internals contain signals about the correctness of responses and the efficacy of specific auxiliary context when answering a query. 
Concretely, we study the following research questions:

\begin{enumerate}[leftmargin=*, itemsep=0pt, topsep=0pt]
    \item \textbf{RQ1:} \textit{Can we estimate the \textit{correctness of a model generation} from its model internals alone?}
    \item \textbf{RQ2:} \textit{Can we estimate the \textit{efficacy of a given context} directly from model internals?}
\end{enumerate}
% \textbf{RQ1:} \textit{Can we estimate the \textit{correctness of a model generation} from its model internals alone?}\\
% \textbf{RQ2:} \textit{Can we estimate the \textit{efficacy of a given context} directly from model internals?}
% \nishant{the spacing isn't ideal here}

To answer these questions, we study six LLMs across three %~\nishant{how many model families?}
model families and sizes in an open-domain factual question-answering setting on the TriviaQA~\citep{joshi2017triviaqa} and MMLU~\citep{hendryckstest2021} datasets. We train classifiers on model internals to predict generation correctness and context relevance.
To operationalize contextual relevance, we introduce two novel measures: \textit{contextual log-likelihood gain} and \textit{contextual relative utility} and study whether model internals can discriminate between contexts across two axes: correctness and relevance. We find that:
%Additionally, to operationalize the concept of contextual relevance, we introduce two novel measures: \textit{contextual log-likelihood gain} and \textit{contextual relative utility} and study whether model internals can discriminate between contexts across two axes: correctness and relevance. We find that:

\begin{enumerate}[leftmargin=*, noitemsep, topsep=0pt]
\item We can estimate correctness of a model generation to open-domain questions from model internals alone with over 75\% accuracy and 70\% AUC-ROC.%~\nishant{should we report an AUC number or is this distracting? My first thought would be ok but whats the majority class baseline. But I'm not the representative reviewer/reader.}. 
\item Using our model internals-based contextual log-likelihood gain metric, we can effectively discriminate between contexts across both the axes of correctness and relevance.
% \item Using just the single-token activations after processing the input at the first layer, we can estimate correctness with over 85\% accuracy.
\end{enumerate}

\begin{figure*}[t!]
\centering
\includegraphics[width=\linewidth]{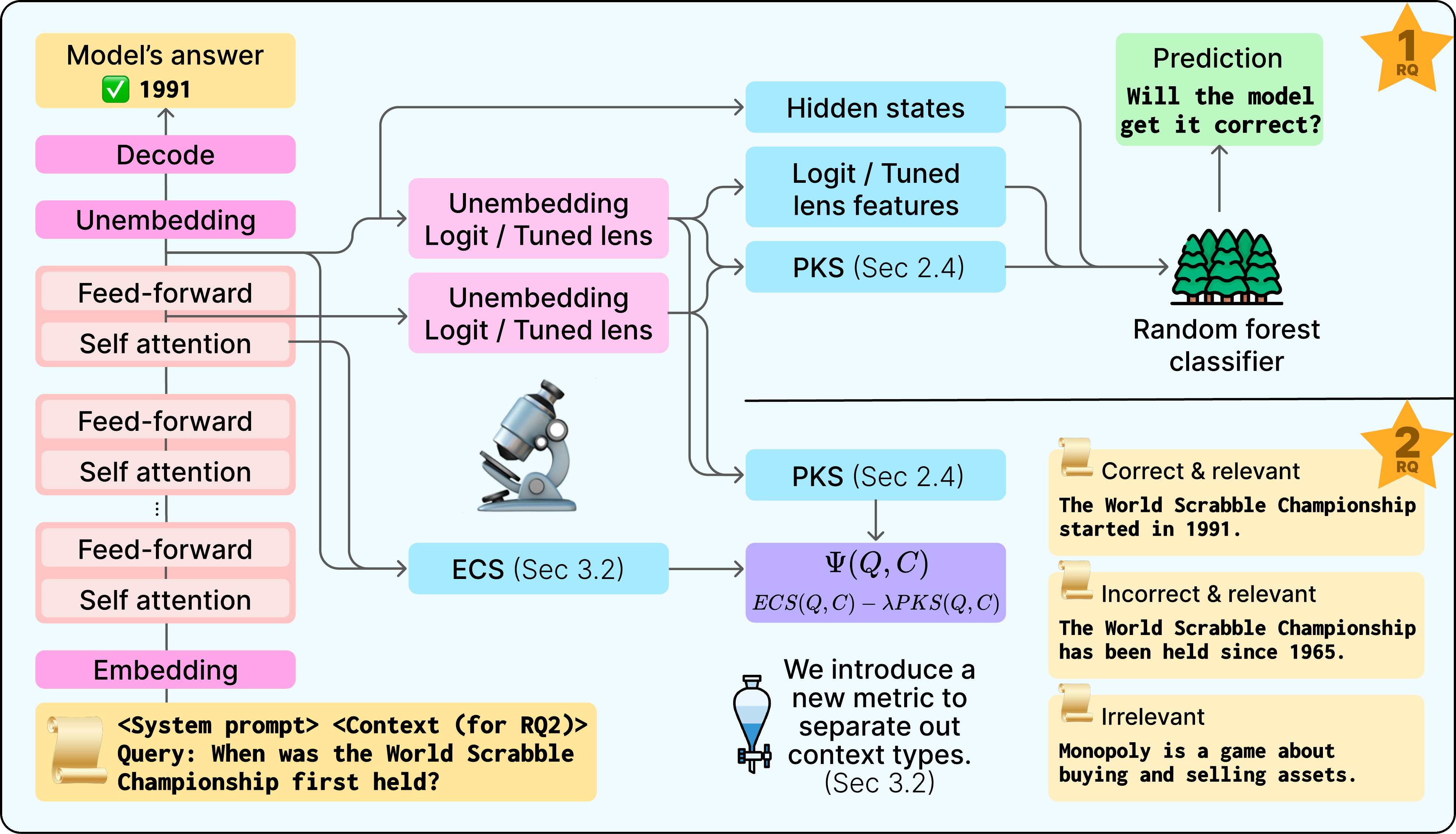}
\caption{Overview of our framework. For RQ1, we use model internals, including hidden states, Logit/Tuned Lens-based features, and parametric knowledge score (PKS) to train classifiers that predict the correctness of a model’s output when answering a question. For RQ2, we analyze how internal signals like external context score (ECS) and PKS respond to different types of external context (correct, incorrect, irrelevant) in order to assess the model’s sensitivity to context when generating answers.}
\label{fig:framework_overview}
\end{figure*}
\section{Related Work}

\paragraph{Confidence Estimation}

Confidence estimation methods are typically categorized as \textit{closed-box} or \textit{open-box}. Closed-box methods prompt the model to assess its own correctness, either by estimating the likelihood that its answer is correct or by judging whether it knows the answer~\citep{kadavath2022language, tian2023just, yin2023large}. Some use linguistic cues~\citep{mielke2022reducing} or generation consistency~\citep{manakul2023selfcheckgpt, zhang-etal-2023-sac3, chen2024insidellmsinternalstates}. Open-box methods instead analyze the model’s internals, either through logit-based uncertainty estimates~\citep{murray2018correcting,kadavath2022language,huang2023look,vazhentsev2023efficient} or by probing hidden activations directly~\citep{li2023inferencetime,orgad2024llmsknowshowintrinsic,burns2024discoveringlatentknowledgelanguage,mice_for_cats}.%~\nishant{this doesnt read well, try to rewrite} 

Our work builds on this line of open-box internal-state approaches but differs in several key ways. First, we focus on predicting the correctness of generated answers in open-domain QA, rather than limiting to binary statements or questions~\citep{azaria-mitchell-2023-internal, burns2024discoveringlatentknowledgelanguage}. Second, we avoid complex proxy objectives (\eg predicting other tokens unrelated to QA%~\nishant{i dont know what this means}
) and instead directly estimate generation correctness~\citep{kadavath2022language}. We use a simple random forest classifier to both preserve feature interpretability and avoid entangling effects of classifier complexity~\citep{orgad2024llmsknowshowintrinsic}. This simple yet effective setup allows us to demonstrate that a variety of internal-state features derived from diverse interpretability techniques can support robust confidence estimation in generative tasks.~\footnote{See~\citet{geng-etal-2024-survey} for a broader overview.}

\paragraph{Parametric and Contextual Knowledge}

The activation spaces of LLMs encode structured, editable knowledge through mechanisms like induction heads~\citep{circuits}, knowledge neurons~\citep{dai2022knowledgeneuronspretrainedtransformers}, and feed-forward layers~\citep{geva2022transformer}. Probing tools such as Logit-Lens~\citep{logitlens}, Tuned-Lens~\citep{tunedlens}, Future-Lens~\citep{pal2023future}, and Backward-Lens~\citep{katz2024backward}, and methods like causal tracing~\citep{meng2022rome}, attention analysis~\citep{geva2023attnknockout, yuksekgonul2023attention}, and steering~\citep{subramani2020discovering, subramani2022extracting, turner2023steering}, reveal how predictions evolve across layers.
Additionally, different layers of transformer-based language models encode different linguistic properties~\citep{tenney2019bert, ethayarajh-2019-contextual, Li2025ModelIS}.%~\nishant{TODO change the Li2025 citation to the updated one once we push it to arXiv}
To improve factuality, retrieval-augmented approaches incorporate external context~\citep{roller2020recipes, shuster2021retrieval, ravichander2025halogenfantasticllmhallucinations}. Recent work~\citep{Wu2024ClashEvalQT, rags_riches, redeep, li2025taming} examine clashes between internal and retrieved knowledge, identify model over-reliance on retrieved context, and develop metrics to diagnose and treat that reliance. %\citet{rags_riches} find models may overly rely on retrieved context (“short-circuiting”), while \citet{redeep} introduce metrics (ReDeEP, AARF) to diagnose and rebalance this reliance.

\section{RQ1: Estimating Correctness}
\label{sec:methods_rq1}

To study RQ1, we formulate a binary classification task to predict whether a generated answer is correct. %\footnote{This is a binary form of the confidence estimation task.} 
Each instance in this task consists of a factual question $Q$, the ground-truth answer $A^*$, an LLM-generated answer $A$, and all activations and outputs $\mathcal{H}_{Q,A}$ produced by a model $M$ when attempting to answer $Q$. 
We train a simple classifier $f$ that takes $Q$, $A$, and $\mathcal{H}_{Q,A}$ as input and predicts $\mathbb{I}[A=A*]$.
The classifier does not have access to the ground-truth answer $A*$ or any external knowledge source; $A*$ is only used to create the ground-truth label for training. We explore several choices for $\mathcal{H}_{Q,A}$ as input features, which we describe below.

\subsection{Asking LLMs Directly}
\label{sec:asking_llms_directly}
We explore two prompting strategies to measure how well LLMs can express their confidence in an answer: \textit{Prompting without Answers} and \textit{Prompting with Answers}.
In \textit{Prompting without Answers}, the model $M$ is given a question $Q$, and asked to output a confidence score between 0 and 100 indicating how confident it is about answering the question without actually generating an answer. 
In \textit{Prompting with Answers}, $M$ first generates a candidate answer $A$ for $Q$ and then is asked to output a confidence score based on how certain it is about the generated answer (see~\cref{appn:prompting_details} for prompting setup details).

%To assess whether models can express their confidence in answering questions through output tokens, we explore two prompting strategies: \textit{Prompting with Answers} and \textit{Prompting without Answers}. In the former, the model is given a question and its answer, and asked to output a confidence score between 0 and 100 indicating how certain it is about the provided answer. In the latter, the model is prompted to express its confidence score in answering the question without revealing the answer itself. See \cref{appn:prompting_details} for further details on the prompting setups.

\subsection{Decoding from Intermediate Layers}
\label{sec: decoding_from_intermediate_layers}

To test whether model internals contain signals for estimating answer correctness, we look at two techniques that facilitate decoding from intermediate hidden states: \textit{Logit Lens} and \textit{Tuned Lens}~\citep{logitlens, tunedlens}.
Both of these methods convert intermediate hidden states into the vocabulary space. We obtain the following input features $\mathcal{H}_{Q,A}$ from these probability distributions. These features are computed per layer and sequence position and provided at once to the classifier. See \cref{appn:logit_lens_tuned_lens} and \cref{appn:input_features} for additional details on \textit{Logit/Tuned Lens} and these features.

% \paragraph{Features:} We convert the resulting probability distributions from both \textit{Logit Lens} and \textit{Tuned Lens} into a set of relevant input features $\mathcal{H}_{Q,A}$ to pass into our correctness classifier $f$.
% We use entropy, rank of the correct token, top-$p$ presence, and cross entropy as relevant features.\footnote{See~\cref{appn:input_features} for details on these features.}

\begin{itemize}[leftmargin=*, noitemsep, topsep=0pt]
    \item The \textbf{Shannon's entropy} measures the uncertainty in the model's probability distribution over the next token at each layer and sequence position \citep{shannon1948mathematical}.
    \item The \textbf{output token rank} is the position, in descending order of log-probability, of the token selected by the model at each layer. While this rank is related to probability, it serves as a more direct proxy for the token a decoding algorithm is likely to generate.%~\nishant{rewrite this "the distribution" is very vague}
    \item The \textbf{top-$p$ presence} of the output token is a binary indicator of whether the token generated by the model appears in the top-$p$ nucleus set at each layer ($p\in \{ 0.5, 0.9, 0.95, 0.99 \}$~\citet{Holtzman2019TheCC}).
    \item The \textbf{cross-entropy} quantifies how similar model predictions are across layers and is computed as the negative log probability of the generated token under the distribution at each layer.%~\nishant{should we use likelihood rather than probability here?}
\end{itemize}

\subsection{Hidden States}
Rather than transforming the hidden states and decoding from them, we experiment with using the model's hidden states directly as input features to $f$. 
Formally, we use $\mathbf{h} \in \mathbb{R}^d$ at layer $\ell \in \{1, ..., L\}$ from the first token position of the forward pass that generated $A$ in response to $Q$ to train $f$.
Since activations could have different scales due to differences in parameter norms throughout training~\citep{merrill-etal-2021-effects}, we convert the value in every dimension $d$ of the activation $h$ to a z-score using means and variances computed from an auxiliary dataset.%~\nishant{do we do this for both mmlu and triviaQA?} 
% We perform an ablation study on the effect of this normalization in \cref{sec:discussion_hidden_state}. 

%We note that $\mathbf{h}$ has a large dimensionality (ranging from $3584$ to $5120$ across the models we experiment with), which can introduce noise leading to overfitting as well as make the classifiers computationally expensive to train. 
%To address this, we experiment with the use of principal-component analysis (PCA) for dimensionality reduction in \cref{sec:discussion_hidden_state}.

%To eliminate scale disparities\jivitesh{@nishant: what's the motivation for using z-score?}, we z-score each hidden state dimension independently. We perform an ablation study on the effect of this normalization in \cref{sec:discussion_hidden_state}. We note that $\mathbf{h}$ has a large dimensionality (ranging from $3584$ to $5120$ across the models we experiment with), which can introduce noise leading to overfitting as well as make the classifiers computationally expensive to train. We ablate the use of principal-component analysis for dimensionality reduction in \cref{sec:discussion_hidden_state}.

\subsection{Parametric Knowledge Scores (PKS)}
\label{sec:pks_defn}
We expect the amount of parametric knowledge used to answer a question $Q$ positively correlates with the confidence the model has about its generated answer $A$.
In other words, the more parametric knowledge used to answer the question the more confidence it should have about that answer.
To quantify the utilization of parametric knowledge, we use the \textit{Parametric Knowledge Score (PKS)} from \citet{redeep}, which measures how much the feedforward networks (FFN) contribute to the activations.
For each token at layer $\ell$, the activations before and after the FFN layer are transformed into a probability distribution over the vocabulary via \textit{Logit Lens}. %: $q(\mathbf{x}_{\text{before}, \ell})$, $q(\mathbf{x}_{\text{after}, \ell})$. 
%For each token at layer $\ell$, the activations before and after the FFN layer are transformed into a probability distribution over the vocabulary via \textit{Logit Lens}: $q(\mathbf{x}_{\text{before}, \ell})$, $q(\mathbf{x}_{\text{after}, \ell})$. 
PKS is the Jensen-Shannon divergence (JSD) between these two distributions, loosely capturing the amount of information imparted by FFN weights into the activations. Formally, the token-level PKS is given by $P_\ell = \mathrm{JSD}\left(q(\mathbf{x}_{\text{before},\ell}) \, \| \, q(\mathbf{x}_{\text{after}, \ell})\right)$, where $q(\cdot) = \mathrm{softmax}(\text{LogitLens}(\cdot))$.
%\footnote{Formally, the token-level PKS is given by $P_\ell = \mathrm{JSD}\left(q(\mathbf{x}_{\text{before},\ell}) \, \| \, q(\mathbf{x}_{\text{after}, \ell})\right)$, where $q(\cdot) = \mathrm{softmax}(\text{LogitLens}(\cdot))$.}
%A higher PKS value corresponds to a stronger reliance on parametric knowledge, but can vary depending on the input query.

%To quantify the utilization of parametric knowledge in transformer-based LLMs, we use the \textit{Parametric Knowledge Score (PKS)} following \citep{redeep}, which measures the contribution of FFN modules to the residual stream. Specifically, for each token at layer $\ell$, the hidden states before and after the FFN layer are decoded into vocabulary distributions using the Logit Lens. The PKS is then defined as the Jensen-Shannon divergence between these two distributions, capturing the information injected by the FFN. Formally, the token-level PKS is given by $P_\ell = \mathrm{JSD}\left(q(\mathbf{x}_{\text{mid},\ell}) \, \| \, q(\mathbf{x}_\ell)\right)$, where $q(\cdot) = \mathrm{softmax}(\text{LogitLens}(\cdot))$. A higher PKS reflects a stronger reliance on parametric knowledge and varies depending on the input query.

\section{RQ2: Estimating Efficacy of Context}

Our goal is to see whether we can measure the efficacy of a given context $C$ from either internal (via model internals) or external (via prompting) features. 
We look at two attributes of context: correctness and relevance and define three types of context $C$: 
\begin{itemize}[leftmargin=*, noitemsep, topsep=0pt]
    \item Correct and relevant ($C_{\text{correct}}$): aligns with the gold answer and has essential information.
    \item Incorrect but relevant ($C_{\text{incorrect}}$): structurally similar and topically aligned, but contains incorrect or misleading information.
    \item Irrelevant ($C_{\text{irrelevant}}$): topically unrelated to and unhelpful for answering the question.
\end{itemize}

\noindent We define two lenses with which we can observe the effect of contexts: \textit{contextual log-likelihood gain} and \textit{contextual relative utility}.

\textit{Contextual log-likelihood gain} measures how much incorporating a context $C$ improves (positive) or degrades (negative) the model's confidence in generating the correct answer. 
In RAG, it quantifies the utility of the retrieved context to the model to generate the correct answer.
For a question $Q$, context $C$, and ground-truth answer tokens $\mathbf{y} = (y_1, \ldots, y_T)$, the \textit{contextual log-likelihood gain} is defined as:
\begin{align}
\label{eq:contextual_LL_gain}
\text{LL}(Q, C) = & \sum_{t=1}^T \log p(y_t \mid y_{<t}, Q, C) - \sum_{t=1}^T \log p(y_t \mid y_{<t}, Q) 
\end{align}%\nishant{todo rewrite such that its shorter}
%This score measures how much the presence of context $C$ improves or degrades the model’s confidence in generating the correct answer. In retrieval-augmented generation, it quantifies the utility of the retrieved context in helping the model produce accurate outputs.

\textit{Contextual relative utility} compares two different contexts $C_1$ and $C_2$ and measures whether $C_1$ is more helpful (positive) or harmful (negative) than $C_2$ for a model to produce the correct answer.
% For a question $Q$ and contexts $C_1$ and $C_2$, 
We formally define \textit{contextual relative utility} as $\Delta\text{LL}(Q, C_1, C_2) = \text{LL}(Q, C_1) - \text{LL}(Q, C_2)$.

\subsection{Prompting-Based Confidence Estimation}
% We evaluate whether we can ask LLMs directly to generate confidence scores.
We analyze whether prompting-based methods can approximate the model's contextual log-likelihood gain through asking the model to generate confidence scores.
For each question $Q$, we prompt the model to output a confidence score on a 0–100 scale under the two different context conditions discussed in~\cref{sec:asking_llms_directly}: \textit{Prompting without Answers} and \textit{Prompting with Answers}.
%In \textit{Prompting without Answers}, the model is asked to report confidence while being shown only the question $Q$ and context $C$. In \textit{Prompting with Answers}, the LLM is additionally given its candidate answer $A$ (generated directly from the model using $Q$ and $C$) and asked to report confidence scores.

Let $\text{Conf}(Q, C)$ denote the model’s confidence score when answering $Q$ with context $C$, and $\text{Conf}(Q)$ denote its confidence when answering without any context. 
We define the \textit{prompting-based contextual gain} as: 
\begin{equation}
\label{eq:prompting_based_contextual_gain}
{\Omega}(Q, C) = \text{Conf}(Q, C) - \text{Conf}(Q). 
\end{equation}
To compare two contexts $C_1$ and $C_2$, we define the \textit{prompting-based relative utility} as $\Delta{\Omega}(Q, C_1, C_2) = {\Omega}(Q, C_1) - {\Omega}(Q, C_2)$,
% \begin{equation}
% \label{eq:prompting_based_relative_utility}
% \Delta{\Omega}(Q, C_1, C_2) = {\Omega}(Q, C_1) - {\Omega}(Q, C_2)
% \end{equation}
which estimates the relative helpfulness of $C_1$ over $C_2$. If the model can accurately distinguish context quality, we expect:
$\Delta{\Omega}(Q, C_{\text{correct}}, C_{\text{incorrect}}) > 0,$ and $ \Delta{\Omega}(Q, C_{\text{correct}}, C_{\text{irrelevant}}) > 0.$\footnote{See \cref{appn:prompt_rq2_explanation} for the reasoning, and \cref{appn:prompting_rq2} for details on the prompting setup.}

\subsection{Internals-Based Confidence Estimation}
To measure how well additional context can affect question-answering ability along the two axes of external context utilization and internal parametric knowledge reliance, we use \textit{external context score} (ECS) and \textit{parametric knowledge score} (PKS) from~\citet{redeep}.
\textit{External Context Score (ECS)} captures the extent to which a language model relies on external context during generation. For each output token \( t_n \), ECS is computed as the cosine similarity between the token’s final-layer hidden state \( x^L_n \) and the mean-pooled embedding \( e \) of the top-\(k\%\) most attended context tokens, selected based on attention weights. %\footnote{See \cref{appn:ecs} for more details on ECS.} 
The token-level ECS is defined as:
\begin{equation}
\text{ECS}^l_h(t_n) = \frac{e \cdot x^L_n}{\|e\| \|x^L_n\|}
\end{equation}

We can obtain the ECS score across a multi-token generation by simply averaging token-wise ECS scores over all output tokens.
A higher ECS indicates stronger alignment between the generated output and the retrieved context, suggesting that the model is effectively utilizing external information.
%For a description of PKS see~\cref{sec:pks_defn}. 

In our setting, parametric knowledge and external context can be thought of as orthogonal. %since one often would either use parametric knowledge or external context to answer a question.
We define a proxy for \textit{contextual log-likelihood gain} using model internals as:
\begin{equation}
\label{eq:LL_internal}
{\Psi}(Q,C) = \text{ECS}(Q,C) - \lambda\cdot\text{PKS}(Q,C),
%\text{LL}^{(\text{int})}(Q,C) = \text{ECS}(Q,C) - \lambda\cdot\text{PKS}(Q,C),
\end{equation}
%\[
%\text{LL}^{(\text{internal})}(Q, C) = \text{ECS}(Q, C) - \lambda \cdot \text{PKS}(Q, C),
%\]
where $\text{ECS}(Q, C)$ measures the model’s reliance on external context, $\text{PKS}(Q, C)$ quantifies the effect of parametric knowledge in the presence of context, and $\lambda$ is a scaling factor that rescales PKS to match ECS. In our experiment, we choose the value of $\lambda$ for each dataset such that the mean PKS score is rescaled to match the mean ECS score, computed across all examples and all context types. This normalization ensures that per-example differences in ${\Psi}(Q,C)$ reflect shifts in context reliance without introducing category-specific bias.
To compare two contexts $C_1$ and $C_2$, we define \textit{internals-based relative utility} as:
\begin{align}
\label{eq:internals_based_relative_utility}
%& \Delta\text{LL}^{(\text{int})}(Q, C_1, C_2) \notag \\
%& = \text{LL}^{(\text{int})}(Q, C_1) - \text{LL}^{(\text{int})}(Q, C_2)
\Delta{\Psi}(Q, C_1, C_2) &= {\Psi}(Q, C_1) - {\Psi}(Q, C_2)
\end{align}
Our formulation reflects the tradeoff between contextual and parametric knowledge: higher values of $\Psi(Q, C)$ suggest stronger reliance on external context, while lower values indicate dominance of parametric knowledge or potential confusion from misleading context.

In addition, we directly compute contextual relative utility using the actual log-likelihood, denoted as $\Delta{\Psi}^{(\text{log likelihood})}$. Specifically, we apply the log softmax over the vocabulary to obtain token-level log probabilities, average these over the ground-truth answer tokens, and then take the difference between two context types.

\section{Experimental Setup}
\label{sec: experimental_setup}
\paragraph{Datasets}

We use factual short-question-answering datasets TriviaQA~\citep{joshi2017triviaqa} and MMLU~\citep{hendryckstest2021} for our experiments. We report results on both datasets for RQ1. As TriviaQA is open-ended while MMLU is multiple-choice, this ensures generalizability of our results across question types. As RQ2 requires accompanying context, we only use TriviaQA for this research question as MMLU does not provide context passages or evidence documents.
To ensure our experiments remain computationally viable, we only consider the validation subset of TriviaQA, from which we retain 6,557 examples after quality filtering (see \cref{appn:additional_dataset_details} for details). These are split 80\%-20\% for training and testing our classifiers. To account for possible variations in referring to the same entity, we use GPT-4o~\citep{openai2024gpt4ocard} as an LLM-judge to evaluate the correctness of model-generated answers against all the ground-truth answer variations present in TriviaQA.
We use the 14,042 test examples of the ``all" subset of MMLU as our training set and the 1,531 validation examples as our test set. As MMLU questions are multiple-choice, we use regex-based answer extraction and verification.

\paragraph{External Context:}
We use the original evidence document provided by TriviaQA as the correct context, or a summarized version if the original exceeds 500 tokens.
% As MMLU does not contain evidence documents, we synthetically generate them by prompting an LLM with the question and ground-truth answer.
We then construct incorrect and irrelevant contexts by modifying or substituting these documents in controlled ways: for incorrect context, we ask an LLM to replace all references to the ground-truth answers with incorrect alternatives; for irrelevant context, we sample the correct context of a different example with low textual similarity (see \cref{appn:external_context} for details and examples).

%Additionally, we require that the SentenceBERT~\citep{reimers-2019-sentence-bert} embeddings using all-MiniLM-L6-v2%\footnote{\url{https://huggingface.co/sentence-transformers/all-MiniLM-L6-v2}} of the candidate context have a cosine similarity $<0.3$ with the ``correct'' context for that question $Q$.

% \paragraph{Formatting of Prompts for RQ1 and RQ2:} For our model external or prompting based strategies described in~\cref{sec:asking_llms_directly}, we prepend a short instruction to each question $Q$ for RQ1.
% For RQ2, we add the instruction to the question along with either the ``correct,'' ``incorrect,'' or ``irrelevant'' context. 
% All prompts used during the dataset preparation process can be found in~\cref{appn:prompting_details}.

\paragraph{Models:}
We experiment with 6 models across families and sizes: LLaMA-3-8B-Instruct~\citep{grattafiori2024llama}, LLaMA-2-7B-Chat-HF and LLaMA-2-13B-Chat-HF~\citep{touvron2023llama}, Qwen-2.5-3B-Chat and Qwen-2.5-7B-Chat~\citep{yang2024qwen2}, and Gemma-2-9B-It~\citep{team2024gemma}.
%To balance model family diversity and size variation, we conduct experiments using the following models: LLaMA-3-8B-Instruct \citep{grattafiori2024llama}, LLaMA-2-7B-Chat-HF and LLaMA-2-13B-Chat-HF \citep{touvron2023llama}, Qwen-2.5-3B-Chat and Qwen-2.5-7B-Chat \citep{yang2024qwen2}, and Gemma-2-9B-It \citep{team2024gemma}.

\paragraph{Methodology Details:}
We choose random forest classifiers for their interpretability, which facilitates feature importance analysis, while also capturing non-linear patterns in the data. These classifiers are trained on the features discussed in \cref{sec:methods_rq1} (see~\cref{appn:methodology_rq1} for hyper-parameters and training speifications). For RQ2, we estimate \textit{contextual log-likelihood gain} using ECS and PKS using~\cref{eq:LL_internal}.
For ECS, we average across both attention heads and output tokens and for PKS, we average across output tokens to capture the overall effect of parametric knowledge on the entire output. 
PKS and ECS are calculated per layer and then averaged across layers. We present a layer-wise analysis in \cref{sec:discussion_rq2}. We calculate one $\lambda$ for each model by rescaling the relative magnitudes of PKS and ECS, averaged over all examples and context types in the training set.

% \paragraph{Methodology Details for RQ1:}
% We choose random forest classifiers for their interpretability, which facilitates feature importance analysis, while also capturing non-linear patterns in the data. These classifiers are trained on features extracted from intermediate model representations. Specifically, we use decoded logits from Logit-Lens and Tuned-Lens, as well as hidden states from each transformer layer at the first output token.\footnote{See~\cref{appn:methodology_rq1} for details on feature extraction, hidden state normalization, and training procedure.} %We use a random forest classifier for its strong performance and interpretability.\footnote{See~\cref{appn:methodology_rq1} for details on feature extraction, hidden state normalization, and training procedure.}
%To address RQ1, we train classifiers using features derived from intermediate model representations. Specifically, we leverage decoded logits from Logit-Lens and Tuned-Lens, as well as hidden states from each transformer layer at the first output token. We use a random forest classifier for its strong performance and interpretability. Full details of feature extraction, hidden state normalization, and training procedure are provided in \cref{appn:methodology_rq1}.

\begin{table*}[t!]
\centering
\newcommand{\rot}[1]{\rotatebox[origin=c]{90}{#1}}
\resizebox{\textwidth}{!}{%
\begin{tabular}{llcccccc}
\toprule
 Dataset & Context Type & LLaMA 3 8B & LLaMA 2 13B & LLaMA 2 7B & Gemma 2 9B & Qwen 2.5 7B & Qwen 2.5 3B \\
\midrule
\multirow{4}{*}{TriviaQA}
 & None       & 0.708 & 0.591 & 0.611 & 0.749 & 0.558 & 0.435 \\
 & Correct    & 0.888 & 0.833 & 0.845 & 0.906 & 0.856 & 0.815 \\
 & Incorrect  & 0.358 & 0.340 & 0.314 & 0.406 & 0.332 & 0.301 \\
 & Irrelevant & 0.579 & 0.516 & 0.465 & 0.727 & 0.488 & 0.293 \\
\midrule
 MMLU & None       & 0.611 & 0.484 & 0.462 & 0.709 & 0.693 & 0.634 \\
 % & Correct    & 0.938 & 0.883 & 0.832 & 0.976 & 0.973 & 0.954 \\
 % & Incorrect  & 0.418 & 0.419 & 0.419 & 0.426 & 0.461 & 0.437 \\
 % & Irrelevant & 0.525 & 0.504 & 0.454 & 0.722 & 0.692 & 0.569 \\
\bottomrule
\end{tabular}}
\caption{Model accuracies on TriviaQA and MMLU datasets with no context and with varying context conditions. MMLU does not contain context or evidence documents.}%~\Nishant{all the tables and figures are too shifted up. Lets try and move them all down a bit.}
\label{tab:model_acc}
\end{table*}

%%%%%%%%%% RQ1 RESULT TABLES %%%%%%%%%%%%
\begin{table*}[t!]

\begin{subtable}{\textwidth}
\centering
\newcommand{\rot}[1]{\rotatebox[origin=c]{90}{#1}}
\resizebox{\textwidth}{!}{
\begin{tabular}{lccccccc}
\toprule
 & Estimator & LLaMA 3 8B & LLaMA 2 13B & LLaMA 2 7B & Gemma 2 9B & Qwen 2.5 7B & Qwen 2.5 3B \\
\midrule
\multirow{7}{*}{\rot{Accuracy}}
 & Majority        & 0.699 & 0.591 & 0.621 & 0.735 & 0.567 & 0.565 \\
 & Prompt w/ A     & \textbf{0.789} & 0.599 & 0.634 & 0.762 & 0.727 & 0.652 \\
 & Prompt w/o A    & 0.699 & 0.611 & 0.621 & 0.768 & 0.637 & 0.605 \\
% & \cellcolor{cyan!20}Logit lens      & \cellcolor{cyan!20}0.925\rlap{$^{*\dagger\ddagger}$} & \cellcolor{cyan!20}\textbf{0.918}\rlap{$^{*\dagger\ddagger}$} & \cellcolor{cyan!20}\textbf{0.925}\rlap{$^{*\dagger\ddagger}$} & \cellcolor{cyan!20}\textbf{0.921}\rlap{$^{*\dagger\ddagger}$} & \cellcolor{cyan!20}0.918\rlap{$^{*\dagger\ddagger}$} & \cellcolor{cyan!20}\textbf{0.916}\rlap{$^{*\dagger\ddagger}$} \\
& \cellcolor{cyan!20}Logit lens       & \cellcolor{cyan!20}0.782\rlap{$^{*\ddagger}$} & \cellcolor{cyan!20}\textbf{0.779}\rlap{$^{*\dagger\ddagger}$} & \cellcolor{cyan!20}\textbf{0.776}\rlap{$^{*\dagger\ddagger}$} & \cellcolor{cyan!20}0.774\rlap{$^{*}$} & \cellcolor{cyan!20}0.751\rlap{$^{*\ddagger}$} & \cellcolor{cyan!20}0.729\rlap{$^{*\dagger\ddagger}$} \\
 % & \cellcolor{cyan!20}Tuned lens      & \cellcolor{cyan!20}\textbf{0.926}\rlap{$^{*\dagger\ddagger}$} & \cellcolor{cyan!20}0.914\rlap{$^{*\dagger\ddagger}$} & \cellcolor{cyan!20}0.920\rlap{$^{*\dagger\ddagger}$} & \cellcolor{cyan!20}-     & \cellcolor{cyan!20}-     & \cellcolor{cyan!20}-     \\
& \cellcolor{cyan!20}Tuned lens        & \cellcolor{cyan!20}0.779\rlap{$^{*\ddagger}$} & \cellcolor{cyan!20}0.775\rlap{$^{*\dagger\ddagger}$} & \cellcolor{cyan!20}0.775\rlap{$^{*\dagger\ddagger}$} & \cellcolor{cyan!20}- & \cellcolor{cyan!20}- & \cellcolor{cyan!20}- \\
 & \cellcolor{cyan!20}Hidden states (Best)  & \cellcolor{cyan!20}0.759\rlap{$^{*\ddagger}$} & \cellcolor{cyan!20}0.679\rlap{$^{*\dagger\ddagger}$} & \cellcolor{cyan!20}0.702\rlap{$^{*\dagger\ddagger}$} & \cellcolor{cyan!20}\textbf{0.785}\rlap{$^{*}$} & \cellcolor{cyan!20}\textbf{0.782}\rlap{$^{*\dagger\ddagger}$} & \cellcolor{cyan!20}\textbf{0.749}\rlap{$^{*\dagger\ddagger}$} \\
 % Hidden states l=1 skipped below
 % & \cellcolor{cyan!20}Hidden states ($\ell=1$)& \cellcolor{cyan!20}0.668\rlap{$^{*\dagger\ddagger}$} & \cellcolor{cyan!20}0.615\rlap{$^{*\dagger\ddagger}$} & \cellcolor{cyan!20}0.619\rlap{$^{*\dagger\ddagger}$} & \cellcolor{cyan!20}0.733\rlap{$^{*\dagger\ddagger}$} & \cellcolor{cyan!20}0.631\rlap{$^{*\dagger\ddagger}$} & \cellcolor{cyan!20}0.628\rlap{$^{*\dagger\ddagger}$} \\
 % & \cellcolor{cyan!20}PKS             & \cellcolor{cyan!20}0.914\rlap{$^{*\dagger\ddagger}$} & \cellcolor{cyan!20}0.898\rlap{$^{*\dagger\ddagger}$} & \cellcolor{cyan!20}0.885\rlap{$^{*\dagger\ddagger}$} & \cellcolor{cyan!20}0.913\rlap{$^{*\dagger\ddagger}$} & \cellcolor{cyan!20}0.900\rlap{$^{*\dagger\ddagger}$} & \cellcolor{cyan!20}0.906\rlap{$^{*\dagger\ddagger}$} \\
 & \cellcolor{cyan!20}PKS  & \cellcolor{cyan!20}0.733 & \cellcolor{cyan!20}0.725\rlap{$^{*\dagger\ddagger}$} & \cellcolor{cyan!20}0.709\rlap{$^{*\dagger\ddagger}$} & \cellcolor{cyan!20}0.743 & \cellcolor{cyan!20}0.650\rlap{$^{*}$} & \cellcolor{cyan!20}0.695\rlap{$^{*\dagger\ddagger}$} \\

\midrule
\multirow{7}{*}{\rot{AUC-ROC}}
 & Majority        & 0.500 & 0.500 & 0.500 & 0.500 & 0.500 & 0.500 \\
 & Prompt w/ A     & 0.783 & 0.512 & 0.537 & 0.592 & 0.736 & 0.687 \\
 & Prompt w/o A    & 0.591 & 0.582 & 0.542 & 0.742 & 0.653 & 0.631 \\
% & \cellcolor{cyan!20}Logit lens      & \cellcolor{cyan!20}\textbf{0.966}\rlap{$^{*\dagger\ddagger}$} & \cellcolor{cyan!20}\textbf{0.975}\rlap{$^{*\dagger\ddagger}$} & \cellcolor{cyan!20}0.975\rlap{$^{*\dagger\ddagger}$} & \cellcolor{cyan!20}\textbf{0.966}\rlap{$^{*\dagger\ddagger}$} & \cellcolor{cyan!20}\textbf{0.979}\rlap{$^{*\dagger\ddagger}$} & \cellcolor{cyan!20}\textbf{0.982}\rlap{$^{*\dagger\ddagger}$} \\
& \cellcolor{cyan!20}Logit lens       & \cellcolor{cyan!20}\textbf{0.790}\rlap{$^{*\ddagger}$} & \cellcolor{cyan!20}\textbf{0.847}\rlap{$^{*\dagger\ddagger}$} & \cellcolor{cyan!20}\textbf{0.835}\rlap{$^{*\dagger\ddagger}$} & \cellcolor{cyan!20}\textbf{0.747}\rlap{$^{*\dagger}$} & \cellcolor{cyan!20}\textbf{0.826}\rlap{$^{*\dagger\ddagger}$} & \cellcolor{cyan!20}\textbf{0.812}\rlap{$^{*\dagger\ddagger}$} \\
 % & \cellcolor{cyan!20}Tuned lens      & \cellcolor{cyan!20}0.964\rlap{$^{*\dagger\ddagger}$} & \cellcolor{cyan!20}0.973\rlap{$^{*\dagger\ddagger}$} & \cellcolor{cyan!20}\textbf{0.976}\rlap{$^{*\dagger\ddagger}$} & \cellcolor{cyan!20}-     & \cellcolor{cyan!20}-     & \cellcolor{cyan!20}-     \\
& \cellcolor{cyan!20}Tuned lens        & \cellcolor{cyan!20}0.782\rlap{$^{*\ddagger}$} & \cellcolor{cyan!20}0.846\rlap{$^{*\dagger\ddagger}$} & \cellcolor{cyan!20}0.829\rlap{$^{*\dagger\ddagger}$} & \cellcolor{cyan!20}- & \cellcolor{cyan!20}- & \cellcolor{cyan!20}- \\
 & \cellcolor{cyan!20}Hidden states (Best)   & \cellcolor{cyan!20}0.647\rlap{$^{*\ddagger}$} & \cellcolor{cyan!20}0.631\rlap{$^{*\dagger\ddagger}$} & \cellcolor{cyan!20}0.647\rlap{$^{*\dagger\ddagger}$} & \cellcolor{cyan!20}0.616\rlap{$^{*}$} & \cellcolor{cyan!20}0.774\rlap{$^{*\ddagger}$} & \cellcolor{cyan!20}0.735\rlap{$^{*\dagger\ddagger}$} \\
 % Hidden states l=1 skipped below
 % & \cellcolor{cyan!20}Hidden states ($\ell=1$) & \cellcolor{cyan!20}0.498\rlap{$^{*\dagger\ddagger}$} & \cellcolor{cyan!20}0.549\rlap{$^{*\dagger\ddagger}$} & \cellcolor{cyan!20}0.535\rlap{$^{*\dagger\ddagger}$} & \cellcolor{cyan!20}0.501\rlap{$^{*\dagger\ddagger}$} & \cellcolor{cyan!20}0.610\rlap{$^{*\dagger\ddagger}$} & \cellcolor{cyan!20}0.595\rlap{$^{*\dagger\ddagger}$} \\
 % & \cellcolor{cyan!20}PKS             & \cellcolor{cyan!20}0.962\rlap{$^{*\dagger\ddagger}$} & \cellcolor{cyan!20}0.962\rlap{$^{*\dagger\ddagger}$} & \cellcolor{cyan!20}0.967\rlap{$^{*\dagger\ddagger}$} & \cellcolor{cyan!20}0.965\rlap{$^{*\dagger\ddagger}$} & \cellcolor{cyan!20}0.975\rlap{$^{*\dagger\ddagger}$} & \cellcolor{cyan!20}0.977\rlap{$^{*\dagger\ddagger}$} \\
 & \cellcolor{cyan!20}PKS & \cellcolor{cyan!20}0.729\rlap{$^{*\ddagger}$} & \cellcolor{cyan!20}0.768\rlap{$^{*\dagger\ddagger}$} & \cellcolor{cyan!20}0.743\rlap{$^{*\dagger\ddagger}$} & \cellcolor{cyan!20}0.723\rlap{$^{*\dagger}$} & \cellcolor{cyan!20}0.715\rlap{$^{*\ddagger}$} & \cellcolor{cyan!20}0.752\rlap{$^{*\dagger\ddagger}$} \\

\bottomrule
\end{tabular}}
\caption{TriviaQA}
\label{tab:main_results_rq1:triviaqa}
\end{subtable}

\vspace{5pt}

\begin{subtable}{\textwidth}
\centering
\newcommand{\rot}[1]{\rotatebox[origin=c]{90}{#1}}
\resizebox{\textwidth}{!}{
\begin{tabular}{lccccccc}
\toprule
 & Estimator & LLaMA 3 8B & LLaMA 2 13B & LLaMA 2 7B & Gemma 2 9B & Qwen 2.5 7B & Qwen 2.5 3B \\
\midrule
\multirow{7}{*}{\rot{Accuracy}}
 & Majority        & 0.604 & 0.538 & 0.548 & 0.717 & 0.692 & 0.627 \\
 & Prompt w/ A     & 0.603 & 0.497 & 0.547 & 0.718 & 0.705 & 0.627 \\
 & Prompt w/o A    & 0.605 & 0.535 & 0.529 & 0.718 & 0.704 & 0.629 \\
 & \cellcolor{cyan!20}Logit lens       & \cellcolor{cyan!20}0.705\rlap{$^{*\dagger\ddagger}$} & \cellcolor{cyan!20}\textbf{0.692}\rlap{$^{*\dagger\ddagger}$} & \cellcolor{cyan!20}\textbf{0.699}\rlap{$^{*\dagger\ddagger}$} & \cellcolor{cyan!20}0.769\rlap{$^{*\dagger\ddagger}$} & \cellcolor{cyan!20}\textbf{0.815}\rlap{$^{*\dagger\ddagger}$} & \cellcolor{cyan!20}0.673\rlap{$^{*\dagger\ddagger}$} \\
 & \cellcolor{cyan!20}Tuned lens        & \cellcolor{cyan!20}0.695\rlap{$^{*\dagger\ddagger}$} & \cellcolor{cyan!20}0.671\rlap{$^{*\dagger\ddagger}$} & \cellcolor{cyan!20}0.684\rlap{$^{*\dagger\ddagger}$} & \cellcolor{cyan!20}- & \cellcolor{cyan!20}- & \cellcolor{cyan!20}- \\
 & \cellcolor{cyan!20}Hidden states (Best)  & \cellcolor{cyan!20}\textbf{0.744}\rlap{$^{*\dagger\ddagger}$} & \cellcolor{cyan!20}0.686\rlap{$^{*\dagger\ddagger}$} & \cellcolor{cyan!20}0.672\rlap{$^{*\dagger\ddagger}$} & \cellcolor{cyan!20}\textbf{0.801}\rlap{$^{*\dagger\ddagger}$} & \cellcolor{cyan!20}0.777\rlap{$^{*\dagger\ddagger}$} & \cellcolor{cyan!20}\textbf{0.746}\rlap{$^{*\dagger\ddagger}$} \\
 % Hidden states l=1 skipped below
 % & \cellcolor{cyan!20}Hidden states ($\ell=1$)& 0.6558 & 0.6310 & 0.6395 & 0.7485 & 0.7178 & 0.6891 \\
 & \cellcolor{cyan!20}PKS  & \cellcolor{cyan!20}0.605 & \cellcolor{cyan!20}- & \cellcolor{cyan!20}0.543 & \cellcolor{cyan!20}0.705 & \cellcolor{cyan!20}0.691 & \cellcolor{cyan!20}0.618 \\

\midrule
\multirow{7}{*}{\rot{AUC-ROC}}
 & Majority        & 0.500 & 0.500 & 0.500 & 0.500 & 0.500 & 0.500 \\
 & Prompt w/ A     & 0.590 & 0.501 & 0.560 & 0.557 & 0.562 & 0.629 \\
 & Prompt w/o A    & 0.497 & 0.538 & 0.492 & 0.513 & 0.544 & 0.519 \\
 & \cellcolor{cyan!20}Logit lens       & \cellcolor{cyan!20}\textbf{0.798}\rlap{$^{*\dagger\ddagger}$} & \cellcolor{cyan!20}\textbf{0.771}\rlap{$^{*\dagger\ddagger}$} & \cellcolor{cyan!20}\textbf{0.767}\rlap{$^{*\dagger\ddagger}$} & \cellcolor{cyan!20}\textbf{0.843}\rlap{$^{*\dagger\ddagger}$} & \cellcolor{cyan!20}\textbf{0.939}\rlap{$^{*\dagger\ddagger}$} & \cellcolor{cyan!20}0.711\rlap{$^{*\dagger\ddagger}$} \\
 & \cellcolor{cyan!20}Tuned lens        & \cellcolor{cyan!20}0.770\rlap{$^{*\dagger\ddagger}$} & \cellcolor{cyan!20}0.752\rlap{$^{*\dagger\ddagger}$} & \cellcolor{cyan!20}0.760\rlap{$^{*\dagger\ddagger}$}  & \cellcolor{cyan!20}- & \cellcolor{cyan!20}- & \cellcolor{cyan!20}- \\
 & \cellcolor{cyan!20}Hidden states (Best)   & \cellcolor{cyan!20}0.740\rlap{$^{*\dagger\ddagger}$} & \cellcolor{cyan!20}0.684\rlap{$^{*\dagger\ddagger}$} & \cellcolor{cyan!20}0.576\rlap{$^{*\ddagger}$} & \cellcolor{cyan!20}0.736\rlap{$^{*\dagger\ddagger}$} & \cellcolor{cyan!20}0.728\rlap{$^{*\dagger\ddagger}$} & \cellcolor{cyan!20}\textbf{0.733}\rlap{$^{*\dagger\ddagger}$} \\
 % Hidden states l=1 skipped below
 % & \cellcolor{cyan!20}Hidden states ($\ell=1$) & 0.6296 & 0.6305 & 0.5483 & 0.6437 & 0.5935 & 0.6173 \\
 & \cellcolor{cyan!20}PKS & \cellcolor{cyan!20}0.537\rlap{$^{\ddagger}$} & \cellcolor{cyan!20}- & \cellcolor{cyan!20}0.543\rlap{$^{*\ddagger}$} & \cellcolor{cyan!20}0.555\rlap{$^{*\ddagger}$} & \cellcolor{cyan!20}0.541\rlap{$^{*}$} & \cellcolor{cyan!20}0.538 \\

\bottomrule
\end{tabular}}
\caption{MMLU}
\label{tab:main_results_rq1:mmlu}
\end{subtable}

\caption{Performance of various classifiers on the test sets using our proposed methods and baseline approaches. We include two prompting baselines: Prompt w/ A (prompting with answers) and Prompt w/o A (prompting without answers) and a simple majority class baseline (Majority). For Logit Lens, Tuned Lens, and PKS methods, we use all values across layers as input features, while for Hidden States, we choose the best layer. Hidden states are normalized using z-score normalization. Tuned Lens results are omitted for models whose weights are not publicly available \citep{tunedlens}. PKS scores are not available for LLaMA 2 13B on MMLU due to computational constraints. $^*$, $^\dagger$, and $^\ddagger$ indicate statistical significance compared to Majority, Prompt w/ A, and Prompt w/o A, respectively (p-value $<0.05$, two-sided permutation test).}
\label{tab:main_results_rq1}
\end{table*}

% \paragraph{Methodology Details for RQ2:}
% We estimate \textit{contextual log-likelihood gain} using ECS and PKS via~\cref{eq:LL_internal}.
% For ECS, we average across both attention heads and output tokens and for PKS, we average across output tokens to capture the overall effect of parametric knowledge on the entire output. 
% PKS and ECS are calculated per layer and then averaged across layers. We present a layer-wise analysis in \cref{sec:discussion_rq2}.\footnote{We calculate $\lambda$ by rescaling the relative magnitudes of PKS and ECS, averaged over all examples and context types in the training set. There is one $\lambda$ per LLM.} 
%To determine the value of $\lambda$ used in the log-likelihood score for internal-based confidence estimation, we compute a single $\lambda$ per model by rescaling the relative magnitudes of PKS and ECS, averaged over all examples and context types in the training set.
%For PKS, we average across output tokens to capture the overall effect of parametric knowledge on the full output. 
%For ECS, we average across both attention heads and output tokens.
%This results in one estimate for PKS and one estimate for ECS per layer of the underlying model.
%We can calculate \textit{contextual log-likelihood gain} using these estimates via~\cref{eq:LL_internal}.

\section{Results}

We present the accuracies of our six LLMs on both TriviaQA and MMLU in~\cref{tab:model_acc}.
%The question-answering performance of the six models on TriviaQA and MMLU is presented in~\cref{tab:model_acc}.
All models achieve moderate accuracy in the default no-context setting, with the larger and more recent models generally performing better. As expected, accuracy on TriviaQA increases with correct context and drops with incorrect or irrelevant context, highlighting that models are indeed sensitive to external context.
We focus on the no-context setting to study RQ1 in~\cref{sec:res_rq1} and examine different context types for RQ2 in~\cref{sec:res_rq2}.

%\subsection{RQ1: Can We Estimate the Correctness of a Model Generation from Its Model Internals Alone?}
\subsection{RQ1: Can We Estimate Correctness from Model Internals Alone?}
\label{sec:res_rq1}

% TODO: permutation test

\Cref{tab:main_results_rq1} shows that prompting-based baselines perform poorly across all models, barely outperforming the majority-class baseline, even when given access to its own generated answer.
%Whether the model is asked to estimate confidence with or without access to its own generated answer, the resulting predictions are only marginally better than the majority-class baseline.
This is particularly evident in the LLaMA models, suggesting that such prompting strategies fail to estimate correctness accurately and highlight that LLMs are poorly calibrated.

\begin{figure*}[t!]
\centering
% First row
% \includegraphics[width=\linewidth]{figs/layerwise_classifier_metrics_norm_area_under_ROC_curve.png}
\begin{subfigure}{0.49\linewidth}
    \includegraphics[width=\linewidth]{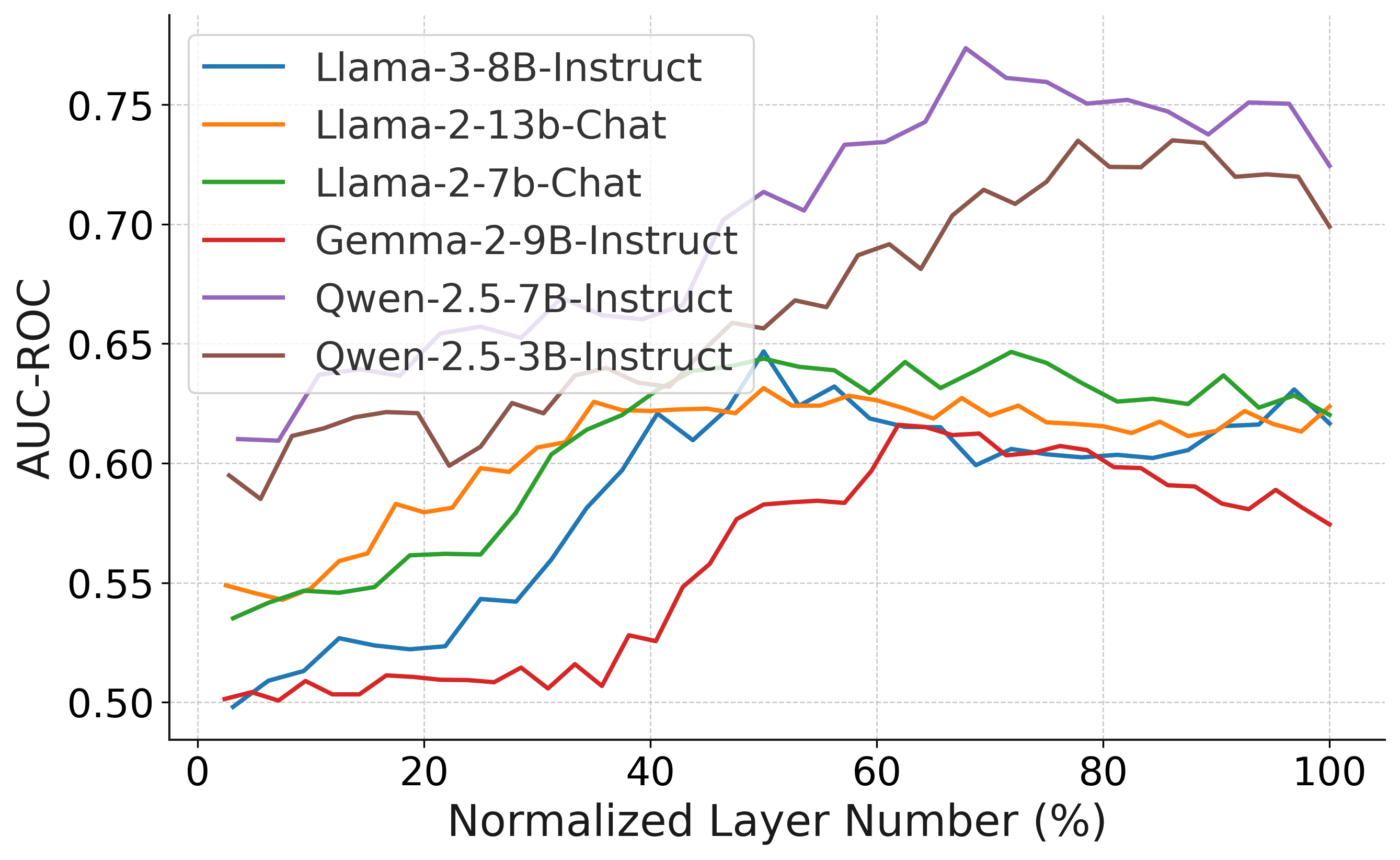}
    \caption{TriviaQA}
  \end{subfigure}
  \hfill
  % ------------ right panel -----------
  \begin{subfigure}{0.49\linewidth}
    \includegraphics[width=\linewidth]{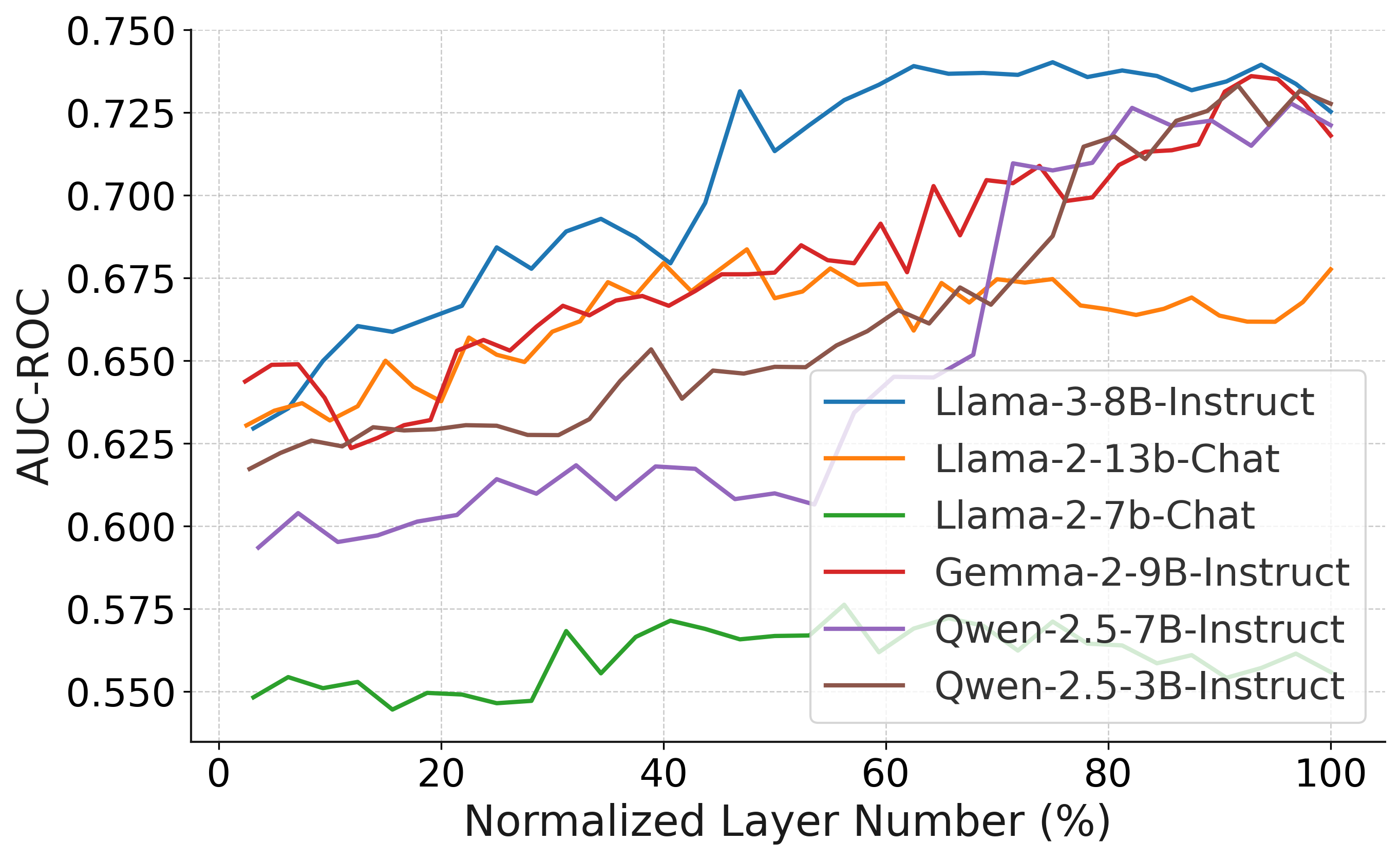}
    \caption{MMLU}           % produces “(b)”
  \end{subfigure}
\caption{Area under ROC curve for random forest classifiers trained on z-score normalized hidden states of each layer. Performance increases with layer depth, suggesting that later layers refine and consolidate decision-relevant signals.}
\label{fig:layerwise_hidden_states_auc_roc}%~\nishant{This figure looks horribly placed and seems to be taking up more space than it needs to}
\end{figure*}

In contrast, the classifier trained on features extracted with the \textit{Logit Lens} show strong performance, yielding the highest AUC-ROC scores on both datasets for most models. % ~\nishant{this is unclear; the classifier trained on the features extracted from the logit lens procedure shows strong performance, not the statistics themselves. Rewrite.}
\textit{Tuned Lens} performs comparably to \textit{Logit Lens}, despite being developed to improve upon \textit{Logit Lens} by better aligning intermediate hidden states with the output distribution through affine transformations. 
We hypothesize that the only distinction between the Tuned Lens and the Logit Lens is the affine transformation, and thus the corresponding scaling and shifting of activations do not contribute additional predictive information.
% We hypothesize that although Tuned Lens may make certain internal features easier to decode, it does not add new predictive information as it merely reshapes the existing representations without introducing additional structure.~\nishant{could say something weaker saying that the only diff between tuned lens and logit lens is the affine transformation and so appropriate scaling and shifting of the activations seems to not matter. It's another way of saying what you say though so idk.}

Surprisingly, classifiers trained directly on hidden states perform nearly as well as those based on Logit Lens features on both datasets.
This finding indicates that the vanilla activations already encode information about the correctness of the model's output.
In~\cref{fig:layerwise_hidden_states_auc_roc}, we observe that performance improves with depth, suggesting that later layers refine and consolidate decision-relevant signals.
This result implies that intermediate decoding methods may not be necessary to predict correctness and the strong performance of classifiers trained on first token hidden states across models may allow for correctness auditing of model responses early in the generation process.
% \jivitesh{Point to figure 1 and say how first layer is pretty good for classification already.}

Lastly, we find that PKS alone is strongly predictive of correctness on TriviaQA, despite being explicitly designed to just measure the influence of feedforward networks on token representations.
This indicates that the feedforward layers of an LLM impart information onto the activations differently based on how in-distribution a given query is. % to data it has seen.%in different ways based on when there exists relevant information versus not.
% In other words, the parametric confidence, as captured by PKS, seems to be able to distinguish between questions grounded in the model's parametric knowledge and those that are not.
However, PKS performs close to random chance for MMLU, suggesting that its discriminatory power may be limited to open-ended question answering and not extend to single-token multiple choice questions.

\subsection{RQ2: Can We Estimate Context Efficacy Directly from Model Internals?}
\label{sec:res_rq2}

% TODO: permutation test

\Cref{tab:main_results_rq2} shows that internals-based confidence estimation significantly outperforms prompting-based methods when distinguishing between correct and incorrect contexts. This result is striking: while models fail to express higher confidence when generating answers conditioned on correct context, their internal activations nonetheless reflect this difference. All models exceed the 50\% random baseline by a wide margin. We do not include results for LLaMA 2 13B due to computational constraints, nor for Gemma 2 9B due to the inapplicability of ECS to its architecture. Further explanation is provided in \cref{appn:res_analysis_rq2}.%~\nishant{is this footnoteable? Should we footnote this?}

When comparing correct and irrelevant contexts, \textit{Prompting without Answer} performs well, often on par with internal-based estimation and consistently outperforms the \textit{Prompting with Answer} baseline. 
This suggests that including the model's own answer in the prompt may mislead it, especially when the answer is incorrect due to misleading context. 
Without the generated answer, however, models appear better able to distinguish irrelevant contexts.
%, indicating a latent capacity to recognize unhelpful input. 
This is paradoxical: despite being able to differentiate between relevant and irrelevant contexts, the underlying LLM remains strongly influenced by irrelevant context (see \cref{tab:model_acc}), suggesting that recognition alone is insufficient to steer the model towards an accurate answer.
We suspect that during training, the model is rarely given irrelevant context and learns to implicitly trust contextual information.
%This presents an apparent contradiction: although both prompting and internal metrics can differentiate irrelevant contexts from correct ones, the model’s generations remain strongly influenced by irrelevant content as shown in \cref{tab:model_acc}. These results suggest that LLMs may internally recognize the quality of context, even if that recognition does not translate into more accurate output, which highlights a gap between internal awareness and surface behavior.
%This presents an apparent contradiction: although both prompting and internal metrics can differentiate irrelevant contexts from correct ones, the model’s generations remain strongly influenced by irrelevant content as shown in \cref{tab:model_acc}. These results suggest that LLMs may internally recognize the quality of context, even if that recognition does not translate into more accurate output, which highlights a gap between internal awareness and surface behavior.

\begin{table*}[t]
\centering
\resizebox{\textwidth}{!}{
\begin{tabular}{llcccccc}
\toprule
Context Comparison & Differentiator & LLaMA 3 8B & LLaMA 2 7B & Qwen 2.5 7B & Qwen 2.5 3B & Average \\
\midrule
% Correct > Incorrect & Prompt w/ A & 19.8 & 0.9 & 10.5 & 6.7 & 32.1 & 15.9 \\
% & Prompt w/o A & 20.8 & 12.3 & 11.2 & 29.8 & 29.3 & 23.0 \\
% & $\Delta LL$ & 84.6 & & 69.7 & 34.4 & 85.1 & 75.5 \\
Correct $>$ Incorrect & $\Delta\Omega^{(\text{prompt w/ A})}$ & 19.7 & 10.9 & 32.7 & 16.4 & 19.9 \\
& $\Delta\Omega^{(\text{prompt w/o A})}$ & 20.8 & 11.3 & 30.0 & 23.6 & 21.4 \\
% & \cellcolor{cyan!20}$\Delta\Psi^{(\text{model internals})}$ & \cellcolor{cyan!20}\textbf{84.6}\rlap{$^{\dagger\ddagger}$} & \cellcolor{cyan!20}\textbf{69.7}\rlap{$^{\dagger\ddagger}$} & \cellcolor{cyan!20}\textbf{34.4}\rlap{$^{\dagger\ddagger}$} & \cellcolor{cyan!20}\textbf{85.1}\rlap{$^{\dagger\ddagger}$} & \cellcolor{cyan!20}\textbf{75.5}\rlap{$^{\dagger\ddagger}$} & \cellcolor{cyan!20}\textbf{69.9}\rlap{$^{\dagger\ddagger}$} \\
& \cellcolor{cyan!20}$\Delta\Psi^{(\text{log likelihood})}$ & \cellcolor{cyan!20}83.9\rlap{$^{\dagger\ddagger}$} & \cellcolor{cyan!20}\textbf{89.5}\rlap{$^{\dagger\ddagger}$}  & \cellcolor{cyan!20}82.6\rlap{$^{\dagger\ddagger}$} & \cellcolor{cyan!20}\textbf{83.3}\rlap{$^{\dagger\ddagger}$} & \cellcolor{cyan!20}\textbf{84.8}\rlap{$^{\dagger\ddagger}$} \\
 & \cellcolor{cyan!20}$\Delta\Psi^{(\text{model internals})}$ & \cellcolor{cyan!20}\textbf{85.5}\rlap{$^{\dagger\ddagger}$} & \cellcolor{cyan!20}70.2\rlap{$^{\dagger\ddagger}$} & \cellcolor{cyan!20}\textbf{85.2}\rlap{$^{\dagger\ddagger}$} & \cellcolor{cyan!20}75.9\rlap{$^{\dagger\ddagger}$} & \cellcolor{cyan!20}79.2\rlap{$^{\dagger\ddagger}$} \\
\midrule

% Correct > Irrelevant & Prompt w/ A & 83.3 & 4.2 & 39.4 & 7.2 & 51.0 & 88.1 \\
% & Prompt w/o A & 94.1 & 86.5 & 68.9 & 96.6 & 79.9 & 71.7 \\
% & $\Delta LL$ & 86.1 & & 90.2 & 75.0 & 90.1 & 71.2 \\
Correct $>$ Irrelevant & $\Delta\Omega^{(\text{prompt w/ A})}$ & 83.2 & 39.7 & 52.2 & \textbf{88.5} & 65.9 \\
& $\Delta\Omega^{(\text{prompt w/o A})}$ & \textbf{93.6} & 68.4 & 79.9 & 72.4 & 78.6 \\
% & \cellcolor{cyan!20}$\Delta\Psi^{(\text{model internals})}$ & \cellcolor{cyan!20}86.1\rlap{$^{\dagger}$} & \cellcolor{cyan!20}\textbf{90.2}\rlap{$^{\dagger\ddagger}$} & \cellcolor{cyan!20}75.0\rlap{$^{\dagger}$} & \cellcolor{cyan!20}\textbf{90.1}\rlap{$^{\dagger\ddagger}$} & \cellcolor{cyan!20}71.2 & \cellcolor{cyan!20}\textbf{82.5}\rlap{$^{\dagger}$} \\
 & \cellcolor{cyan!20}$\Delta\Psi^{(\text{log likelihood})}$ & \cellcolor{cyan!20}76.7 & \cellcolor{cyan!20}74.7\rlap{$^{\dagger\ddagger}$} & \cellcolor{cyan!20}74.5\rlap{$^{\dagger}$} & \cellcolor{cyan!20}80.6\rlap{$^{\ddagger}$} & \cellcolor{cyan!20}76.6\rlap{$^{\dagger}$} \\
 & \cellcolor{cyan!20}$\Delta\Psi^{(\text{model internals})}$ & \cellcolor{cyan!20}86.3\rlap{$^{\dagger}$} & \cellcolor{cyan!20}\textbf{90.5}\rlap{$^{\dagger\ddagger}$} & \cellcolor{cyan!20}\textbf{89.6}\rlap{$^{\dagger\ddagger}$} & \cellcolor{cyan!20}70.6 & \cellcolor{cyan!20}\textbf{84.3}\rlap{$^{\dagger\ddagger}$} \\
% \midrule

% Incorrect > Irrelevant & Prompt w/ A & 79.0 & 4.2 & 35.9 & 7.0 & 44.7 & 87.1 \\
% & Prompt w/o A & 90.4 & 82.0 & 68.3 & 94.2 & 74.2 & 69.6 \\
% & $\Delta LL$ & 57.7 & & 86.9 & 80.0 & 66.9 & 52.8 \\

\bottomrule
\end{tabular}
}
\caption{Proportion of examples where $\Delta\text{LL}$ successfully distinguishes context quality using prompting-based and internal-based confidence estimators on TriviaQA. We compute $\Delta\text{LL}(Q, C_{\text{correct}}, C_{\text{incorrect}})$ and $\Delta\text{LL}(Q, C_{\text{correct}}, C_{\text{irrelevant}})$ separately for each example and report the fraction for which the result is greater than zero. $^\dagger$ and $^\ddagger$ indicate statistical significance compared to Prompt w/ A and Prompt w/o A, respectively (p-value $<$ 0.05, two-sided permutation test).}
\label{tab:main_results_rq2}
\end{table*}

% \begin{table*}[t]
% \centering
% \resizebox{\textwidth}{!}{
% \begin{tabular}{llccccccc}
% \toprule
% Context Comparison & Differentiator & LLaMA 3 8B & LLaMA 2 7B & Gemma 2 9B & Qwen 2.5 7B & Qwen 2.5 3B & Average \\
% \midrule
% Correct $>$ Incorrect & $\Delta\Omega^{(\text{prompt w/ A})}$ & 20.4 & 21.6 & 19.1 & 36.0 & 28.8 \\
% & $\Delta\Omega^{(\text{prompt w/o A})}$ & 18.7 & 23.6 & 21.9 & 33.0 & 22.3 \\
% & \cellcolor{cyan!20}$\Delta\Psi^{(\text{log likelihood})}$ &  &  &  &  &  &  \\
% & \cellcolor{cyan!20}$\Delta\Psi^{(\text{model internals})}$ &  &  &  &  &  &  \\
% \midrule
% Correct $>$ Irrelevant & $\Delta\Omega^{(\text{prompt w/ A})}$ & 56.8 & 45.2 & 39.7 & 68.8 & 92.1 \\
% & $\Delta\Omega^{(\text{prompt w/o A})}$ & 91.0 & 60.4 & 86.0 & 95.1 & 94.8 \\
% & \cellcolor{cyan!20}$\Delta\Psi^{(\text{log likelihood})}$ &  &  &  &  &  &  \\
% & \cellcolor{cyan!20}$\Delta\Psi^{(\text{model internals})}$ &  &  &  &  &  &  \\

% \bottomrule
% \end{tabular}
% }
% \caption{Proportion of examples where $\Delta\text{LL}$ successfully distinguishes context quality using prompting-based and internal-based confidence estimators on MMLU. We compute $\Delta\text{LL}(Q, C_{\text{correct}}, C_{\text{incorrect}})$ and $\Delta\text{LL}(Q, C_{\text{correct}}, C_{\text{irrelevant}})$ separately for each example and report the fraction for which the result is greater than zero. $^\dagger$ and $^\ddagger$ indicate statistical significance compared to Prompt w/ A and Prompt w/o A, respectively (p-value < 0.05, two-sided permutation test).}
% \label{tab:main_results_rq2}
% \end{table*}

\section{Discussion}

\subsection{RQ1}

\paragraph{Prompting Baselines}
Here, we look at how well calibrated the prompting-based approaches are by looking at a reliability diagram and measuring smooth expected calibration error (Smooth ECE)~\citep{blasiok2023smooth}.
In our reliability diagrams, we plot the confidence (x-axis) vs. accuracy (y-axis) and plot the line $y=x$ indicating a perfectly calibrated system.
\Cref{fig:smooth_ece_with_answer,fig:smooth_ece_without_answer,fig:smooth_ece_with_answer_mmlu,fig:smooth_ece_without_answer_mmlu} in our Appendix show that all prompting baselines for all LLMs for both TriviaQA and MMLU are systematically overconfident (\ie predictions with $x$\% confidence yield $<x$\% accuracy).
 %for both prompting baselines in \cref{fig:smooth_ece_with_answer,fig:smooth_ece_without_answer} (TriviaQA) and \cref{fig:smooth_ece_with_answer_mmlu,fig:smooth_ece_without_answer_mmlu} (MMLU) in the Appendix. A perfectly calibrated model would follow the diagonal line, where predicted confidence matches observed accuracy. However, the calibration curves for all models lie consistently above the diagonal, indicating systematic overconfidence over the tested models (e.g., predictions with 80\% confidence yield only around 70\% accuracy).

%To gain a more nuanced understanding of the calibration behavior of prompting-based approaches, we visualize the Smooth Expected Calibration Error (Smooth ECE) \citep{blasiok2023smooth} for both prompting baselines in \cref{fig:smooth_ece_with_answer,fig:smooth_ece_without_answer} (TriviaQA) and \cref{fig:smooth_ece_with_answer_mmlu,fig:smooth_ece_without_answer_mmlu} (MMLU) in the Appendix. A perfectly calibrated model would follow the diagonal line, where predicted confidence matches observed accuracy. However, the calibration curves for all models lie consistently above the diagonal, indicating systematic overconfidence over the tested models (e.g., predictions with 80\% confidence yield only around 70\% accuracy).

\paragraph{Logit Lens vs. Tuned Lens}

% \begin{table}[h!]
% \centering
% \begin{tabular}{lllll}
% \toprule
% Metrics & Estimator & LLaMA 3 8B & LLaMA 2 13B & LLaMA 2 7B \\
% \midrule
% \multirow{2}{*}{ACC} & Logit Lens & 0.925 & 0.918 & 0.925 \\
%  & Tuned Lens & 0.926 & 0.914 & 0.920 \\
% \midrule
% \multirow{2}{*}{AUC-ROC} & Logit Lens & 0.966 & 0.975 & 0.975 \\
%  & Tuned Lens & 0.964 & 0.973 & 0.976 \\
% \bottomrule
% \end{tabular}
% \caption{Correctness estimator}
% \label{tab:logit_lens_vs_tuned_lens}
% \end{table}

In addition to the comparison presented in \cref{tab:main_results_rq1}, we further train separate classifiers on logit lens and tuned lens features of each individual layer to assess performance across layers on TriviaQA (see \cref{fig:layerwise_auc_logit_vs_tuned} in Appendix). The performance of the two methods varies across layers without a consistent trend and neither consistently outperforms the other.
We observe that earlier layers are also highly predictive of generation correctness, achieving performance comparable to later layers. This strengthens our finding that information about output correctness is available early on in the models' activations, underscoring the potential of early auditing.

\paragraph{Logit/Tuned Lens: Feature Importance}

We analyze feature importance scores from random forest classifiers for TriviaQA in  \cref{fig:feature_importance_lens} in the Appendix. Entropy and cross-entropy from the final layer emerge as the most important features across all models. This is intuitive: output token logits and distributions directly capture the model’s confidence when generating the correct answer. Additionally, features from the last $\sim5$ layers generally exhibit higher importance.

\paragraph{Logit/Tuned Lens: AUC-ROC Curves of Features from External vs. Internal Layers}

To examine whether features from non-final layers (denoted as \textit{Internal}) are as predictive as those from the final layer alone (denoted as \textit{External}), we train separate classifiers using each feature set and compare their AUC-ROC curves, as shown in~\cref{fig:roc_curves_all_models} for TriviaQA and~\cref{fig:roc_curves_all_models_mmlu} for MMLU in the Appendix. Interestingly, across all models, classifiers trained on internal layers achieve higher AUC scores than those using only the final layer. This suggests that intermediate representations can carry more information than the final layer alone for predicting generation correctness.

\paragraph{Hidden States: Effect of Z-Score Normalization}
\label{sec:discussion_hidden_state}

To evaluate the impact of z-score normalization on classifier performance, we compare models trained on hidden state features with and without normalization and find virtually identical performance (see \cref{apn:hidden_states}). This suggests that the classifier is robust to the absolute magnitude of hidden state feature values across different dimensions and examples.

\subsection{RQ2}
\label{sec:discussion_rq2}

% \paragraph{PKS and ECS:}
% While we have shown that both $\Delta \Psi(Q, C_{\text{correct}}, C_{\text{incorrect}})$ and $\Delta \Psi(Q, C_{\text{correct}}, C_{\text{irrelevant}})$ demonstrate strong discriminative power, it remains unclear which component, PKS or ECS, contributes more to this separation. 

% \cref{fig:pks_ecs_histogram} in the Appendix presents the distributions of PKS and ECS scores across all TriviaQA examples. We observe that PKS values remain stable regardless of the context type (correct, incorrect, irrelevant or no context), indicating that parametric knowledge is largely unaffected by contextual input. In contrast, ECS scores show pronounced sensitivity to context variation: switching from correct to either incorrect or irrelevant context causes a noticeable drop in ECS. These findings support our interpretation that PKS reflects the model’s reliance on internal parametric knowledge and serves as a proxy for intrinsic confidence, while ECS captures the extent to which the model attends to and integrates external contexts. High ECS scores are associated with helpful contexts that the model learns to attend to, suggesting that internal representations indeed encode a meaningful signal for context utility.

\paragraph{Example Analysis}

We visualize ECS and PKS values across model layers and output tokens for selected TriviaQA examples in Figures~\ref{fig:ag_redeep_1}, \ref{fig:ag_redeep_2}, and \ref{fig:ag_redeep_3} in the Appendix. While PKS scores show only small variations across tokens and layers, later layers show noticeably higher values. Here, initial output tokens have higher PKS values than later ones. These patterns suggest that parametric knowledge accumulated in later layers has a stronger influence on the output distribution, particularly at the beginning of generation.%~\nishant{This still doesnt make sense; can you rewrite}

% We visualize ECS and PKS values across model layers and output tokens for selected examples in Figures~\ref{fig:ag_redeep_1}, \ref{fig:ag_redeep_2}, and \ref{fig:ag_redeep_3}. PKS scores remain relatively uniform across tokens and layers, with noticeably higher value in later layers. In these late layers, PKS values are higher for the initial output tokens than for the last ones, suggesting that parametric knowledge becomes more strongly activated in the later layers and primarily guides the model at the onset of token generation.

In contrast, ECS scores vary more across tokens, reflecting the model’s selective use of context depending on the importance and influence of the context on each output token. Uninformative tokens tend to exhibit lower ECS scores. Across layers, ECS shows relatively little variation, indicating that attention to external context is distributed broadly across the network rather than being isolated to specific layers. While features continue to evolve across layers, the degree to which tokens attend to one another’s features appears relatively stable.%~\nishant{theres something left to be said that something like even though the features evolve layer by layer, the amount each token attends to the features of a different token seems to not vary, but we dont directly test this. Might be a way to incorporate something like that in.}
\section{Conclusion}
Our experiments demonstrate that we can indeed predict the correctness of a model generation from model internals alone.
In fact, with just the activations of first output token, we can predict correctness with about 75\% accuracy, hinting that early auditing could be possible.
Prompting, on the other hand, is poorly calibrated and thus has little utility.
Additionally, using model internals based \textit{contextualized log-likelihood}, we can estimate the efficacy of external context along two axis: correctness and relevancy.
Taken together, our results suggest that deciphering model internals could provide valuable insight into making language models more trustworthy.

\section*{Reproducibility Statement}
We provide the dataset details, training details, model specifications, and prompts used in the main text or the Appendix to ensure our experiments are reproducible. We use mainstream and commonly-used software libraries, datasets, and models for our experiments. We publicly release our code. %We intend to publicly release our code upon paper publication.~\nishant{update this to include the code}

\section*{Ethics Statement}

Large language models carry significant potential for misuse, both intentional and accidental. Our work aims to advance understanding of the decision-making processes in LLMs and to identify hallucinations in model generations, thereby helping to mitigate some of the harms associated with the spread of inaccurate, model-generated content. Nevertheless, our methods should not be seen as a substitute for careful usage and verification; all model outputs must be independently checked for accuracy, particularly in domains where correctness is critical.

We emphasize that all datasets and models used in this study are available under permissive licenses for research purposes. We rely on instruction-tuned models, which have already undergone safety-related training. At the same time, we acknowledge that probing or intervening in model internals may alter or undermine this safety tuning. Understanding and exposing hidden mechanisms inside LLMs can yield valuable scientific insight, but it also carries risks: such methods could, in principle, be adapted to bypass alignment safeguards or weaken safety behaviors.

We therefore caution that techniques for manipulating internal representations should be applied with care and with an awareness of their broader implications. Our intention in presenting this work is to promote transparency, accountability, and safer deployment of LLMs, not to provide tools that could be used to compromise alignment or safety constraints.

% Large language models have potential for misuse, both intentional and accidental. This work takes a step toward understanding decision-making processes in LLMs and identifying hallucinations in model generations, potentially preventing some of the damage caused by the proliferation of inaccurate, model-generated content. However, our methods should not be considered a substitute for careful usage and verification; all model-generated content should be independently verified for accuracy, especially when the correctness of such information is critical.
% Additionally, the authors note that all datasets and models used in this study are under permissive licenses for research purposes. We use instruction-tuned models, which have already undergone safety-related training.~\nishant{need to write more here, especially about how messing with model internals could lead to breaking safety-tuning and alignment. Change 'the authors' to 'we' wherever you can. It reads like a legal document where you're just checking a box to have an ethics statement.}

% ~\nishant{any acknowledgments?}

% \section*{Acknowledgments}

\bibliography{custom,iclr2026_conference,anthology}

\begin{thebibliography}{66}
\providecommand{\natexlab}[1]{#1}
\providecommand{\url}[1]{\texttt{#1}}
\expandafter\ifx\csname urlstyle\endcsname\relax
  \providecommand{\doi}[1]{doi: #1}\else
  \providecommand{\doi}{doi: \begingroup \urlstyle{rm}\Url}\fi

\bibitem[Akbik et~al.(2019)Akbik, Bergmann, Blythe, Rasul, Schweter, and Vollgraf]{akbik2019flair}
Alan Akbik, Tanja Bergmann, Duncan Blythe, Kashif Rasul, Stefan Schweter, and Roland Vollgraf.
\newblock {FLAIR}: An easy-to-use framework for state-of-the-art {NLP}.
\newblock In \emph{{NAACL} 2019, 2019 Annual Conference of the North American Chapter of the Association for Computational Linguistics (Demonstrations)}, pp.\  54--59, 2019.

\bibitem[Asai et~al.(2023)Asai, Wu, Wang, Sil, and Hajishirzi]{asai2023self}
Akari Asai, Zeqiu Wu, Yizhong Wang, Avirup Sil, and Hannaneh Hajishirzi.
\newblock Self-rag: Learning to retrieve, generate, and critique through self-reflection.
\newblock In \emph{The Twelfth International Conference on Learning Representations}, 2023.

\bibitem[Azaria \& Mitchell(2023)Azaria and Mitchell]{azaria-mitchell-2023-internal}
Amos Azaria and Tom Mitchell.
\newblock The internal state of an {LLM} knows when it`s lying.
\newblock In Houda Bouamor, Juan Pino, and Kalika Bali (eds.), \emph{Findings of the Association for Computational Linguistics: EMNLP 2023}, pp.\  967--976, Singapore, December 2023. Association for Computational Linguistics.
\newblock \doi{10.18653/v1/2023.findings-emnlp.68}.
\newblock URL \url{https://aclanthology.org/2023.findings-emnlp.68/}.

\bibitem[Bang et~al.(2023)Bang, Cahyawijaya, Lee, Dai, Su, Wilie, Lovenia, Ji, Yu, Chung, Do, Xu, and Fung]{bang-etal-2023-multitask}
Yejin Bang, Samuel Cahyawijaya, Nayeon Lee, Wenliang Dai, Dan Su, Bryan Wilie, Holy Lovenia, Ziwei Ji, Tiezheng Yu, Willy Chung, Quyet~V. Do, Yan Xu, and Pascale Fung.
\newblock A multitask, multilingual, multimodal evaluation of {C}hat{GPT} on reasoning, hallucination, and interactivity.
\newblock In Jong~C. Park, Yuki Arase, Baotian Hu, Wei Lu, Derry Wijaya, Ayu Purwarianti, and Adila~Alfa Krisnadhi (eds.), \emph{Proceedings of the 13th International Joint Conference on Natural Language Processing and the 3rd Conference of the Asia-Pacific Chapter of the Association for Computational Linguistics (Volume 1: Long Papers)}, pp.\  675--718, Nusa Dua, Bali, November 2023. Association for Computational Linguistics.
\newblock \doi{10.18653/v1/2023.ijcnlp-main.45}.
\newblock URL \url{https://aclanthology.org/2023.ijcnlp-main.45/}.

\bibitem[Belrose et~al.(2023)Belrose, Furman, Smith, Halawi, Ostrovsky, McKinney, Biderman, and Steinhardt]{tunedlens}
Nora Belrose, Zach Furman, Logan Smith, Danny Halawi, Igor Ostrovsky, Lev McKinney, Stella Biderman, and Jacob Steinhardt.
\newblock Eliciting latent predictions from transformers with the tuned lens, 2023.
\newblock URL \url{https://arxiv.org/abs/2303.08112}.

\bibitem[B{\l}asiok \& Nakkiran(2023)B{\l}asiok and Nakkiran]{blasiok2023smooth}
Jaros{\l}aw B{\l}asiok and Preetum Nakkiran.
\newblock Smooth ece: Principled reliability diagrams via kernel smoothing.
\newblock \emph{arXiv preprint arXiv:2309.12236}, 2023.

\bibitem[Burns et~al.(2024)Burns, Ye, Klein, and Steinhardt]{burns2024discoveringlatentknowledgelanguage}
Collin Burns, Haotian Ye, Dan Klein, and Jacob Steinhardt.
\newblock Discovering latent knowledge in language models without supervision, 2024.
\newblock URL \url{https://arxiv.org/abs/2212.03827}.

\bibitem[Chen et~al.(2024)Chen, Liu, Chen, Gu, Wu, Tao, Fu, and Ye]{chen2024insidellmsinternalstates}
Chao Chen, Kai Liu, Ze~Chen, Yi~Gu, Yue Wu, Mingyuan Tao, Zhihang Fu, and Jieping Ye.
\newblock Inside: Llms' internal states retain the power of hallucination detection, 2024.
\newblock URL \url{https://arxiv.org/abs/2402.03744}.

\bibitem[Dai et~al.(2022)Dai, Dong, Hao, Sui, Chang, and Wei]{dai2022knowledgeneuronspretrainedtransformers}
Damai Dai, Li~Dong, Yaru Hao, Zhifang Sui, Baobao Chang, and Furu Wei.
\newblock Knowledge neurons in pretrained transformers, 2022.
\newblock URL \url{https://arxiv.org/abs/2104.08696}.

\bibitem[Elhage et~al.(2021)Elhage, Nanda, Olsson, Henighan, Joseph, Mann, Askell, Bai, Chen, Conerly, DasSarma, Drain, Ganguli, Hatfield-Dodds, Hernandez, Jones, Kernion, Lovitt, Ndousse, Amodei, Brown, Clark, Kaplan, McCandlish, and Olah]{circuits}
Nelson Elhage, Neel Nanda, Catherine Olsson, Tom Henighan, Nicholas Joseph, Ben Mann, Amanda Askell, Yuntao Bai, Anna Chen, Tom Conerly, Nova DasSarma, Dawn Drain, Deep Ganguli, Zac Hatfield-Dodds, Danny Hernandez, Andy Jones, Jackson Kernion, Liane Lovitt, Kamal Ndousse, Dario Amodei, Tom Brown, Jack Clark, Jared Kaplan, Sam McCandlish, and Chris Olah.
\newblock A mathematical framework for transformer circuits.
\newblock \emph{Transformer Circuits Thread}, 2021.
\newblock https://transformer-circuits.pub/2021/framework/index.html.

\bibitem[Ethayarajh(2019)]{ethayarajh-2019-contextual}
Kawin Ethayarajh.
\newblock How contextual are contextualized word representations? {C}omparing the geometry of {BERT}, {ELM}o, and {GPT}-2 embeddings.
\newblock In Kentaro Inui, Jing Jiang, Vincent Ng, and Xiaojun Wan (eds.), \emph{Proceedings of the 2019 Conference on Empirical Methods in Natural Language Processing and the 9th International Joint Conference on Natural Language Processing (EMNLP-IJCNLP)}, pp.\  55--65, Hong Kong, China, November 2019. Association for Computational Linguistics.
\newblock \doi{10.18653/v1/D19-1006}.
\newblock URL \url{https://aclanthology.org/D19-1006/}.

\bibitem[Geng et~al.(2024)Geng, Cai, Wang, Koeppl, Nakov, and Gurevych]{geng-etal-2024-survey}
Jiahui Geng, Fengyu Cai, Yuxia Wang, Heinz Koeppl, Preslav Nakov, and Iryna Gurevych.
\newblock A survey of confidence estimation and calibration in large language models.
\newblock In Kevin Duh, Helena Gomez, and Steven Bethard (eds.), \emph{Proceedings of the 2024 Conference of the North American Chapter of the Association for Computational Linguistics: Human Language Technologies (Volume 1: Long Papers)}, pp.\  6577--6595, Mexico City, Mexico, June 2024. Association for Computational Linguistics.
\newblock \doi{10.18653/v1/2024.naacl-long.366}.
\newblock URL \url{https://aclanthology.org/2024.naacl-long.366/}.

\bibitem[Geva et~al.(2022)Geva, Caciularu, Wang, and Goldberg]{geva2022transformer}
Mor Geva, Avi Caciularu, Kevin~Ro Wang, and Yoav Goldberg.
\newblock Transformer feed-forward layers build predictions by promoting concepts in the vocabulary space.
\newblock \emph{arXiv preprint arXiv:2203.14680}, 2022.

\bibitem[Geva et~al.(2023)Geva, Bastings, Filippova, and Globerson]{geva2023attnknockout}
Mor Geva, Jasmijn Bastings, Katja Filippova, and Amir Globerson.
\newblock Dissecting recall of factual associations in auto-regressive language models.
\newblock \emph{arXiv preprint arXiv:2304.14767}, 2023.

\bibitem[Grattafiori et~al.(2024)Grattafiori, Dubey, Jauhri, Pandey, Kadian, Al-Dahle, Letman, Mathur, Schelten, Vaughan, et~al.]{grattafiori2024llama}
Aaron Grattafiori, Abhimanyu Dubey, Abhinav Jauhri, Abhinav Pandey, Abhishek Kadian, Ahmad Al-Dahle, Aiesha Letman, Akhil Mathur, Alan Schelten, Alex Vaughan, et~al.
\newblock The llama 3 herd of models.
\newblock \emph{arXiv preprint arXiv:2407.21783}, 2024.

\bibitem[Groeneveld et~al.(2024)Groeneveld, Beltagy, Walsh, Bhagia, Kinney, Tafjord, Jha, Ivison, Magnusson, Wang, Arora, Atkinson, Authur, Chandu, Cohan, Dumas, Elazar, Gu, Hessel, Khot, Merrill, Morrison, Muennighoff, Naik, Nam, Peters, Pyatkin, Ravichander, Schwenk, Shah, Smith, Strubell, Subramani, Wortsman, Dasigi, Lambert, Richardson, Zettlemoyer, Dodge, Lo, Soldaini, Smith, and Hajishirzi]{groeneveld-etal-2024-olmo}
Dirk Groeneveld, Iz~Beltagy, Evan Walsh, Akshita Bhagia, Rodney Kinney, Oyvind Tafjord, Ananya Jha, Hamish Ivison, Ian Magnusson, Yizhong Wang, Shane Arora, David Atkinson, Russell Authur, Khyathi Chandu, Arman Cohan, Jennifer Dumas, Yanai Elazar, Yuling Gu, Jack Hessel, Tushar Khot, William Merrill, Jacob Morrison, Niklas Muennighoff, Aakanksha Naik, Crystal Nam, Matthew Peters, Valentina Pyatkin, Abhilasha Ravichander, Dustin Schwenk, Saurabh Shah, William Smith, Emma Strubell, Nishant Subramani, Mitchell Wortsman, Pradeep Dasigi, Nathan Lambert, Kyle Richardson, Luke Zettlemoyer, Jesse Dodge, Kyle Lo, Luca Soldaini, Noah Smith, and Hannaneh Hajishirzi.
\newblock {OLM}o: Accelerating the science of language models.
\newblock In Lun-Wei Ku, Andre Martins, and Vivek Srikumar (eds.), \emph{Proceedings of the 62nd Annual Meeting of the Association for Computational Linguistics (Volume 1: Long Papers)}, pp.\  15789--15809, Bangkok, Thailand, August 2024. Association for Computational Linguistics.
\newblock \doi{10.18653/v1/2024.acl-long.841}.
\newblock URL \url{https://aclanthology.org/2024.acl-long.841/}.

\bibitem[Gu et~al.(2024)Gu, Jiang, Shi, Tan, Zhai, Xu, Li, Shen, Ma, Liu, et~al.]{gu2024survey}
Jiawei Gu, Xuhui Jiang, Zhichao Shi, Hexiang Tan, Xuehao Zhai, Chengjin Xu, Wei Li, Yinghan Shen, Shengjie Ma, Honghao Liu, et~al.
\newblock A survey on llm-as-a-judge.
\newblock \emph{arXiv preprint arXiv:2411.15594}, 2024.

\bibitem[Guerreiro et~al.(2023)Guerreiro, Alves, Waldendorf, Haddow, Birch, Colombo, and Martins]{guerreiro-etal-2023-hallucinations}
Nuno~M. Guerreiro, Duarte~M. Alves, Jonas Waldendorf, Barry Haddow, Alexandra Birch, Pierre Colombo, and Andr{\'e} F.~T. Martins.
\newblock Hallucinations in large multilingual translation models.
\newblock \emph{Transactions of the Association for Computational Linguistics}, 11:\penalty0 1500--1517, 2023.
\newblock \doi{10.1162/tacl_a_00615}.
\newblock URL \url{https://aclanthology.org/2023.tacl-1.85/}.

\bibitem[Hendrycks et~al.(2021)Hendrycks, Burns, Basart, Zou, Mazeika, Song, and Steinhardt]{hendryckstest2021}
Dan Hendrycks, Collin Burns, Steven Basart, Andy Zou, Mantas Mazeika, Dawn Song, and Jacob Steinhardt.
\newblock Measuring massive multitask language understanding.
\newblock \emph{Proceedings of the International Conference on Learning Representations (ICLR)}, 2021.

\bibitem[Holtzman et~al.(2019)Holtzman, Buys, Du, Forbes, and Choi]{Holtzman2019TheCC}
Ari Holtzman, Jan Buys, Li~Du, Maxwell Forbes, and Yejin Choi.
\newblock The curious case of neural text degeneration.
\newblock \emph{ArXiv}, abs/1904.09751, 2019.
\newblock URL \url{https://api.semanticscholar.org/CorpusID:127986954}.

\bibitem[Huang et~al.(2025)Huang, Yu, Ma, Zhong, Feng, Wang, Chen, Peng, Feng, Qin, and Liu]{survey_hallucination}
Lei Huang, Weijiang Yu, Weitao Ma, Weihong Zhong, Zhangyin Feng, Haotian Wang, Qianglong Chen, Weihua Peng, Xiaocheng Feng, Bing Qin, and Ting Liu.
\newblock A survey on hallucination in large language models: Principles, taxonomy, challenges, and open questions.
\newblock \emph{ACM Trans. Inf. Syst.}, 43\penalty0 (2), January 2025.
\newblock ISSN 1046-8188.
\newblock \doi{10.1145/3703155}.
\newblock URL \url{https://doi.org/10.1145/3703155}.

\bibitem[Huang et~al.(2023)Huang, Song, Wang, Zhao, Chen, Juefei-Xu, and Ma]{huang2023look}
Yuheng Huang, Jiayang Song, Zhijie Wang, Shengming Zhao, Huaming Chen, Felix Juefei-Xu, and Lei Ma.
\newblock Look before you leap: An exploratory study of uncertainty measurement for large language models.
\newblock \emph{arXiv preprint arXiv:2307.10236}, 2023.

\bibitem[Jiang et~al.(2021)Jiang, Araki, Ding, and Neubig]{jiang-etal-2021-know}
Zhengbao Jiang, Jun Araki, Haibo Ding, and Graham Neubig.
\newblock How can we know when language models know? on the calibration of language models for question answering.
\newblock \emph{Transactions of the Association for Computational Linguistics}, 9:\penalty0 962--977, 2021.
\newblock \doi{10.1162/tacl_a_00407}.
\newblock URL \url{https://aclanthology.org/2021.tacl-1.57/}.

\bibitem[Joshi et~al.(2017)Joshi, Choi, Weld, and Zettlemoyer]{joshi2017triviaqa}
Mandar Joshi, Eunsol Choi, Daniel~S Weld, and Luke Zettlemoyer.
\newblock Triviaqa: A large scale distantly supervised challenge dataset for reading comprehension.
\newblock \emph{arXiv preprint arXiv:1705.03551}, 2017.

\bibitem[Kadavath et~al.(2022{\natexlab{a}})Kadavath, Conerly, Askell, Henighan, Drain, Perez, Schiefer, Hatfield-Dodds, DasSarma, Tran-Johnson, Johnston, El-Showk, Jones, Elhage, Hume, Chen, Bai, Bowman, Fort, Ganguli, Hernandez, Jacobson, Kernion, Kravec, Lovitt, Ndousse, Olsson, Ringer, Amodei, Brown, Clark, Joseph, Mann, McCandlish, Olah, and Kaplan]{kadavath_language_2022}
Saurav Kadavath, Tom Conerly, Amanda Askell, Tom Henighan, Dawn Drain, Ethan Perez, Nicholas Schiefer, Zac Hatfield-Dodds, Nova DasSarma, Eli Tran-Johnson, Scott Johnston, Sheer El-Showk, Andy Jones, Nelson Elhage, Tristan Hume, Anna Chen, Yuntao Bai, Sam Bowman, Stanislav Fort, Deep Ganguli, Danny Hernandez, Josh Jacobson, Jackson Kernion, Shauna Kravec, Liane Lovitt, Kamal Ndousse, Catherine Olsson, Sam Ringer, Dario Amodei, Tom Brown, Jack Clark, Nicholas Joseph, Ben Mann, Sam McCandlish, Chris Olah, and Jared Kaplan.
\newblock Language models(mostly) know what they know, November 2022{\natexlab{a}}.
\newblock URL \url{http://arxiv.org/abs/2207.05221}.
\newblock arXiv:2207.05221 [cs].

\bibitem[Kadavath et~al.(2022{\natexlab{b}})Kadavath, Conerly, Askell, Henighan, Drain, Perez, Schiefer, Hatfield-Dodds, DasSarma, Tran-Johnson, et~al.]{kadavath2022language}
Saurav Kadavath, Tom Conerly, Amanda Askell, Tom Henighan, Dawn Drain, Ethan Perez, Nicholas Schiefer, Zac Hatfield-Dodds, Nova DasSarma, Eli Tran-Johnson, et~al.
\newblock Language models (mostly) know what they know.
\newblock \emph{arXiv preprint arXiv:2207.05221}, 2022{\natexlab{b}}.

\bibitem[Katz et~al.(2024)Katz, Belinkov, Geva, and Wolf]{katz2024backward}
Shahar Katz, Yonatan Belinkov, Mor Geva, and Lior Wolf.
\newblock Backward lens: Projecting language model gradients into the vocabulary space.
\newblock \emph{arXiv preprint arXiv:2402.12865}, 2024.

\bibitem[Lewis et~al.(2020)Lewis, Perez, Piktus, Petroni, Karpukhin, Goyal, K{\"u}ttler, Lewis, Yih, Rockt{\"a}schel, et~al.]{lewis2020retrieval}
Patrick Lewis, Ethan Perez, Aleksandra Piktus, Fabio Petroni, Vladimir Karpukhin, Naman Goyal, Heinrich K{\"u}ttler, Mike Lewis, Wen-tau Yih, Tim Rockt{\"a}schel, et~al.
\newblock Retrieval-augmented generation for knowledge-intensive nlp tasks.
\newblock \emph{Advances in neural information processing systems}, 33:\penalty0 9459--9474, 2020.

\bibitem[Li et~al.(2025)Li, Chen, and Tong]{li2025taming}
Gaotang Li, Yuzhong Chen, and Hanghang Tong.
\newblock Taming knowledge conflicts in language models.
\newblock \emph{arXiv preprint arXiv:2503.10996}, 2025.

\bibitem[Li et~al.(2023)Li, Patel, Vi\'{e}gas, Pfister, and Wattenberg]{li2023inferencetime}
Kenneth Li, Oam Patel, Fernanda Vi\'{e}gas, Hanspeter Pfister, and Martin Wattenberg.
\newblock Inference-time intervention: eliciting truthful answers from a language model.
\newblock In \emph{Proceedings of the 37th International Conference on Neural Information Processing Systems}, NIPS '23, Red Hook, NY, USA, 2023. Curran Associates Inc.

\bibitem[Li \& Subramani(2025)Li and Subramani]{Li2025ModelIS}
Michael Li and Nishant Subramani.
\newblock Model internal sleuthing: Finding lexical identity and inflectional morphology in modern language models.
\newblock \emph{ArXiv}, abs/2506.02132, 2025.
\newblock URL \url{https://api.semanticscholar.org/CorpusID:279118488}.

\bibitem[Manakul et~al.(2023)Manakul, Liusie, and Gales]{manakul2023selfcheckgpt}
Potsawee Manakul, Adian Liusie, and Mark~JF Gales.
\newblock Selfcheckgpt: Zero-resource black-box hallucination detection for generative large language models.
\newblock \emph{arXiv preprint arXiv:2303.08896}, 2023.

\bibitem[Meng et~al.(2022)Meng, Bau, Andonian, and Belinkov]{meng2022rome}
Kevin Meng, David Bau, Alex Andonian, and Yonatan Belinkov.
\newblock Locating and editing factual associations in gpt.
\newblock \emph{Advances in neural information processing systems}, 35:\penalty0 17359--17372, 2022.

\bibitem[Merrill et~al.(2021)Merrill, Ramanujan, Goldberg, Schwartz, and Smith]{merrill-etal-2021-effects}
William Merrill, Vivek Ramanujan, Yoav Goldberg, Roy Schwartz, and Noah~A. Smith.
\newblock Effects of parameter norm growth during transformer training: Inductive bias from gradient descent.
\newblock In Marie-Francine Moens, Xuanjing Huang, Lucia Specia, and Scott Wen-tau Yih (eds.), \emph{Proceedings of the 2021 Conference on Empirical Methods in Natural Language Processing}, pp.\  1766--1781, Online and Punta Cana, Dominican Republic, November 2021. Association for Computational Linguistics.
\newblock \doi{10.18653/v1/2021.emnlp-main.133}.
\newblock URL \url{https://aclanthology.org/2021.emnlp-main.133/}.

\bibitem[Mielke et~al.(2022{\natexlab{a}})Mielke, Szlam, Dinan, and Boureau]{mielke-etal-2022-reducing}
Sabrina~J. Mielke, Arthur Szlam, Emily Dinan, and Y-Lan Boureau.
\newblock Reducing conversational agents' overconfidence through linguistic calibration.
\newblock \emph{Transactions of the Association for Computational Linguistics}, 10:\penalty0 857--872, 2022{\natexlab{a}}.
\newblock \doi{10.1162/tacl_a_00494}.
\newblock URL \url{https://aclanthology.org/2022.tacl-1.50/}.

\bibitem[Mielke et~al.(2022{\natexlab{b}})Mielke, Szlam, Dinan, and Boureau]{mielke2022reducing}
Sabrina~J Mielke, Arthur Szlam, Emily Dinan, and Y-Lan Boureau.
\newblock Reducing conversational agents’ overconfidence through linguistic calibration.
\newblock \emph{Transactions of the Association for Computational Linguistics}, 10:\penalty0 857--872, 2022{\natexlab{b}}.

\bibitem[Murray \& Chiang(2018)Murray and Chiang]{murray2018correcting}
Kenton Murray and David Chiang.
\newblock Correcting length bias in neural machine translation.
\newblock \emph{arXiv preprint arXiv:1808.10006}, 2018.

\bibitem[nostalgebraist(2020)]{logitlens}
nostalgebraist.
\newblock interpreting gpt: the logit lens, Aug 2020.
\newblock URL \url{https://www.lesswrong.com/posts/AcKRB8wDpdaN6v6ru/interpreting-gpt-the-logit-lens}.

\bibitem[OpenAI et~al.(2024)OpenAI, Hurst, Lerer, Goucher, Perelman, Ramesh, Clark, Ostrow, Welihinda, Hayes, Radford, Madry, Baker-Whitcomb, Beutel, Borzunov, Carney, Chow, Kirillov, Nichol, Paino, Renzin, Passos, Kirillov, Christakis, Conneau, Kamali, Jabri, Moyer, Tam, Crookes, Tootoochian, Tootoonchian, Kumar, Vallone, Karpathy, Braunstein, Cann, Codispoti, Galu, Kondrich, Tulloch, Mishchenko, Baek, Jiang, Pelisse, Woodford, Gosalia, Dhar, Pantuliano, Nayak, Oliver, Zoph, Ghorbani, Leimberger, Rossen, Sokolowsky, Wang, Zweig, Hoover, Samic, McGrew, Spero, Giertler, Cheng, Lightcap, Walkin, Quinn, Guarraci, Hsu, Kellogg, Eastman, Lugaresi, Wainwright, Bassin, Hudson, Chu, Nelson, Li, Shern, Conger, Barette, Voss, Ding, Lu, Zhang, Beaumont, Hallacy, Koch, Gibson, Kim, Choi, McLeavey, Hesse, Fischer, Winter, Czarnecki, Jarvis, Wei, Koumouzelis, Sherburn, Kappler, Levin, Levy, Carr, Farhi, Mely, Robinson, Sasaki, Jin, Valladares, Tsipras, Li, Nguyen, Findlay, Oiwoh, Wong, Asdar, Proehl, Yang, Antonow, Kramer,
  Peterson, Sigler, Wallace, Brevdo, Mays, Khorasani, Such, Raso, Zhang, von Lohmann, Sulit, Goh, Oden, Salmon, Starace, Brockman, Salman, Bao, Hu, Wong, Wang, Schmidt, Whitney, Jun, Kirchner, de~Oliveira~Pinto, Ren, Chang, Chung, Kivlichan, O'Connell, O'Connell, Osband, Silber, Sohl, Okuyucu, Lan, Kostrikov, Sutskever, Kanitscheider, Gulrajani, Coxon, Menick, Pachocki, Aung, Betker, Crooks, Lennon, Kiros, Leike, Park, Kwon, Phang, Teplitz, Wei, Wolfe, Chen, Harris, Varavva, Lee, Shieh, Lin, Yu, Weng, Tang, Yu, Jang, Candela, Beutler, Landers, Parish, Heidecke, Schulman, Lachman, McKay, Uesato, Ward, Kim, Huizinga, Sitkin, Kraaijeveld, Gross, Kaplan, Snyder, Achiam, Jiao, Lee, Zhuang, Harriman, Fricke, Hayashi, Singhal, Shi, Karthik, Wood, Rimbach, Hsu, Nguyen, Gu-Lemberg, Button, Liu, Howe, Muthukumar, Luther, Ahmad, Kai, Itow, Workman, Pathak, Chen, Jing, Guy, Fedus, Zhou, Mamitsuka, Weng, McCallum, Held, Ouyang, Feuvrier, Zhang, Kondraciuk, Kaiser, Hewitt, Metz, Doshi, Aflak, Simens, Boyd, Thompson,
  Dukhan, Chen, Gray, Hudnall, Zhang, Aljubeh, Litwin, Zeng, Johnson, Shetty, Gupta, Shah, Yatbaz, Yang, Zhong, Glaese, Chen, Janner, Lampe, Petrov, Wu, Wang, Fradin, Pokrass, Castro, de~Castro, Pavlov, Brundage, Wang, Khan, Murati, Bavarian, Lin, Yesildal, Soto, Gimelshein, Cone, Staudacher, Summers, LaFontaine, Chowdhury, Ryder, Stathas, Turley, Tezak, Felix, Kudige, Keskar, Deutsch, Bundick, Puckett, Nachum, Okelola, Boiko, Murk, Jaffe, Watkins, Godement, Campbell-Moore, Chao, McMillan, Belov, Su, Bak, Bakkum, Deng, Dolan, Hoeschele, Welinder, Tillet, Pronin, Tillet, Dhariwal, Yuan, Dias, Lim, Arora, Troll, Lin, Lopes, Puri, Miyara, Leike, Gaubert, Zamani, Wang, Donnelly, Honsby, Smith, Sahai, Ramchandani, Huet, Carmichael, Zellers, Chen, Chen, Nigmatullin, Cheu, Jain, Altman, Schoenholz, Toizer, Miserendino, Agarwal, Culver, Ethersmith, Gray, Grove, Metzger, Hermani, Jain, Zhao, Wu, Jomoto, Wu, Shuaiqi, Xia, Phene, Papay, Narayanan, Coffey, Lee, Hall, Balaji, Broda, Stramer, Xu, Gogineni, Christianson,
  Sanders, Patwardhan, Cunninghman, Degry, Dimson, Raoux, Shadwell, Zheng, Underwood, Markov, Sherbakov, Rubin, Stasi, Kaftan, Heywood, Peterson, Walters, Eloundou, Qi, Moeller, Monaco, Kuo, Fomenko, Chang, Zheng, Zhou, Manassra, Sheu, Zaremba, Patil, Qian, Kim, Cheng, Zhang, He, Zhang, Jin, Dai, and Malkov]{openai2024gpt4ocard}
OpenAI, Aaron Hurst, Adam Lerer, Adam~P. Goucher, Adam Perelman, Aditya Ramesh, Aidan Clark, AJ~Ostrow, Akila Welihinda, Alan Hayes, Alec Radford, Aleksander Madry, Alex Baker-Whitcomb, Alex Beutel, Alex Borzunov, Alex Carney, Alex Chow, Alex Kirillov, Alex Nichol, Alex Paino, Alex Renzin, Alex~Tachard Passos, Alexander Kirillov, Alexi Christakis, Alexis Conneau, Ali Kamali, Allan Jabri, Allison Moyer, Allison Tam, Amadou Crookes, Amin Tootoochian, Amin Tootoonchian, Ananya Kumar, Andrea Vallone, Andrej Karpathy, Andrew Braunstein, Andrew Cann, Andrew Codispoti, Andrew Galu, Andrew Kondrich, Andrew Tulloch, Andrey Mishchenko, Angela Baek, Angela Jiang, Antoine Pelisse, Antonia Woodford, Anuj Gosalia, Arka Dhar, Ashley Pantuliano, Avi Nayak, Avital Oliver, Barret Zoph, Behrooz Ghorbani, Ben Leimberger, Ben Rossen, Ben Sokolowsky, Ben Wang, Benjamin Zweig, Beth Hoover, Blake Samic, Bob McGrew, Bobby Spero, Bogo Giertler, Bowen Cheng, Brad Lightcap, Brandon Walkin, Brendan Quinn, Brian Guarraci, Brian Hsu,
  Bright Kellogg, Brydon Eastman, Camillo Lugaresi, Carroll Wainwright, Cary Bassin, Cary Hudson, Casey Chu, Chad Nelson, Chak Li, Chan~Jun Shern, Channing Conger, Charlotte Barette, Chelsea Voss, Chen Ding, Cheng Lu, Chong Zhang, Chris Beaumont, Chris Hallacy, Chris Koch, Christian Gibson, Christina Kim, Christine Choi, Christine McLeavey, Christopher Hesse, Claudia Fischer, Clemens Winter, Coley Czarnecki, Colin Jarvis, Colin Wei, Constantin Koumouzelis, Dane Sherburn, Daniel Kappler, Daniel Levin, Daniel Levy, David Carr, David Farhi, David Mely, David Robinson, David Sasaki, Denny Jin, Dev Valladares, Dimitris Tsipras, Doug Li, Duc~Phong Nguyen, Duncan Findlay, Edede Oiwoh, Edmund Wong, Ehsan Asdar, Elizabeth Proehl, Elizabeth Yang, Eric Antonow, Eric Kramer, Eric Peterson, Eric Sigler, Eric Wallace, Eugene Brevdo, Evan Mays, Farzad Khorasani, Felipe~Petroski Such, Filippo Raso, Francis Zhang, Fred von Lohmann, Freddie Sulit, Gabriel Goh, Gene Oden, Geoff Salmon, Giulio Starace, Greg Brockman, Hadi
  Salman, Haiming Bao, Haitang Hu, Hannah Wong, Haoyu Wang, Heather Schmidt, Heather Whitney, Heewoo Jun, Hendrik Kirchner, Henrique~Ponde de~Oliveira~Pinto, Hongyu Ren, Huiwen Chang, Hyung~Won Chung, Ian Kivlichan, Ian O'Connell, Ian O'Connell, Ian Osband, Ian Silber, Ian Sohl, Ibrahim Okuyucu, Ikai Lan, Ilya Kostrikov, Ilya Sutskever, Ingmar Kanitscheider, Ishaan Gulrajani, Jacob Coxon, Jacob Menick, Jakub Pachocki, James Aung, James Betker, James Crooks, James Lennon, Jamie Kiros, Jan Leike, Jane Park, Jason Kwon, Jason Phang, Jason Teplitz, Jason Wei, Jason Wolfe, Jay Chen, Jeff Harris, Jenia Varavva, Jessica~Gan Lee, Jessica Shieh, Ji~Lin, Jiahui Yu, Jiayi Weng, Jie Tang, Jieqi Yu, Joanne Jang, Joaquin~Quinonero Candela, Joe Beutler, Joe Landers, Joel Parish, Johannes Heidecke, John Schulman, Jonathan Lachman, Jonathan McKay, Jonathan Uesato, Jonathan Ward, Jong~Wook Kim, Joost Huizinga, Jordan Sitkin, Jos Kraaijeveld, Josh Gross, Josh Kaplan, Josh Snyder, Joshua Achiam, Joy Jiao, Joyce Lee, Juntang
  Zhuang, Justyn Harriman, Kai Fricke, Kai Hayashi, Karan Singhal, Katy Shi, Kavin Karthik, Kayla Wood, Kendra Rimbach, Kenny Hsu, Kenny Nguyen, Keren Gu-Lemberg, Kevin Button, Kevin Liu, Kiel Howe, Krithika Muthukumar, Kyle Luther, Lama Ahmad, Larry Kai, Lauren Itow, Lauren Workman, Leher Pathak, Leo Chen, Li~Jing, Lia Guy, Liam Fedus, Liang Zhou, Lien Mamitsuka, Lilian Weng, Lindsay McCallum, Lindsey Held, Long Ouyang, Louis Feuvrier, Lu~Zhang, Lukas Kondraciuk, Lukasz Kaiser, Luke Hewitt, Luke Metz, Lyric Doshi, Mada Aflak, Maddie Simens, Madelaine Boyd, Madeleine Thompson, Marat Dukhan, Mark Chen, Mark Gray, Mark Hudnall, Marvin Zhang, Marwan Aljubeh, Mateusz Litwin, Matthew Zeng, Max Johnson, Maya Shetty, Mayank Gupta, Meghan Shah, Mehmet Yatbaz, Meng~Jia Yang, Mengchao Zhong, Mia Glaese, Mianna Chen, Michael Janner, Michael Lampe, Michael Petrov, Michael Wu, Michele Wang, Michelle Fradin, Michelle Pokrass, Miguel Castro, Miguel Oom~Temudo de~Castro, Mikhail Pavlov, Miles Brundage, Miles Wang, Minal
  Khan, Mira Murati, Mo~Bavarian, Molly Lin, Murat Yesildal, Nacho Soto, Natalia Gimelshein, Natalie Cone, Natalie Staudacher, Natalie Summers, Natan LaFontaine, Neil Chowdhury, Nick Ryder, Nick Stathas, Nick Turley, Nik Tezak, Niko Felix, Nithanth Kudige, Nitish Keskar, Noah Deutsch, Noel Bundick, Nora Puckett, Ofir Nachum, Ola Okelola, Oleg Boiko, Oleg Murk, Oliver Jaffe, Olivia Watkins, Olivier Godement, Owen Campbell-Moore, Patrick Chao, Paul McMillan, Pavel Belov, Peng Su, Peter Bak, Peter Bakkum, Peter Deng, Peter Dolan, Peter Hoeschele, Peter Welinder, Phil Tillet, Philip Pronin, Philippe Tillet, Prafulla Dhariwal, Qiming Yuan, Rachel Dias, Rachel Lim, Rahul Arora, Rajan Troll, Randall Lin, Rapha~Gontijo Lopes, Raul Puri, Reah Miyara, Reimar Leike, Renaud Gaubert, Reza Zamani, Ricky Wang, Rob Donnelly, Rob Honsby, Rocky Smith, Rohan Sahai, Rohit Ramchandani, Romain Huet, Rory Carmichael, Rowan Zellers, Roy Chen, Ruby Chen, Ruslan Nigmatullin, Ryan Cheu, Saachi Jain, Sam Altman, Sam Schoenholz, Sam
  Toizer, Samuel Miserendino, Sandhini Agarwal, Sara Culver, Scott Ethersmith, Scott Gray, Sean Grove, Sean Metzger, Shamez Hermani, Shantanu Jain, Shengjia Zhao, Sherwin Wu, Shino Jomoto, Shirong Wu, Shuaiqi, Xia, Sonia Phene, Spencer Papay, Srinivas Narayanan, Steve Coffey, Steve Lee, Stewart Hall, Suchir Balaji, Tal Broda, Tal Stramer, Tao Xu, Tarun Gogineni, Taya Christianson, Ted Sanders, Tejal Patwardhan, Thomas Cunninghman, Thomas Degry, Thomas Dimson, Thomas Raoux, Thomas Shadwell, Tianhao Zheng, Todd Underwood, Todor Markov, Toki Sherbakov, Tom Rubin, Tom Stasi, Tomer Kaftan, Tristan Heywood, Troy Peterson, Tyce Walters, Tyna Eloundou, Valerie Qi, Veit Moeller, Vinnie Monaco, Vishal Kuo, Vlad Fomenko, Wayne Chang, Weiyi Zheng, Wenda Zhou, Wesam Manassra, Will Sheu, Wojciech Zaremba, Yash Patil, Yilei Qian, Yongjik Kim, Youlong Cheng, Yu~Zhang, Yuchen He, Yuchen Zhang, Yujia Jin, Yunxing Dai, and Yury Malkov.
\newblock Gpt-4o system card, 2024.
\newblock URL \url{https://arxiv.org/abs/2410.21276}.

\bibitem[Orgad et~al.(2024)Orgad, Toker, Gekhman, Reichart, Szpektor, Kotek, and Belinkov]{orgad2024llmsknowshowintrinsic}
Hadas Orgad, Michael Toker, Zorik Gekhman, Roi Reichart, Idan Szpektor, Hadas Kotek, and Yonatan Belinkov.
\newblock Llms know more than they show: On the intrinsic representation of llm hallucinations, 2024.
\newblock URL \url{https://arxiv.org/abs/2410.02707}.

\bibitem[Pal et~al.(2023)Pal, Sun, Yuan, Wallace, and Bau]{pal2023future}
Koyena Pal, Jiuding Sun, Andrew Yuan, Byron~C Wallace, and David Bau.
\newblock Future lens: Anticipating subsequent tokens from a single hidden state.
\newblock \emph{arXiv preprint arXiv:2311.04897}, 2023.

\bibitem[Ravichander et~al.(2025)Ravichander, Ghela, Wadden, and Choi]{ravichander2025halogenfantasticllmhallucinations}
Abhilasha Ravichander, Shrusti Ghela, David Wadden, and Yejin Choi.
\newblock Halogen: Fantastic llm hallucinations and where to find them, 2025.
\newblock URL \url{https://arxiv.org/abs/2501.08292}.

\bibitem[Reimers \& Gurevych(2019)Reimers and Gurevych]{reimers-2019-sentence-bert}
Nils Reimers and Iryna Gurevych.
\newblock Sentence-bert: Sentence embeddings using siamese bert-networks.
\newblock In \emph{Proceedings of the 2019 Conference on Empirical Methods in Natural Language Processing}. Association for Computational Linguistics, 11 2019.
\newblock URL \url{https://arxiv.org/abs/1908.10084}.

\bibitem[Roller et~al.(2020)Roller, Dinan, Goyal, Ju, Williamson, Liu, Xu, Ott, Shuster, Smith, et~al.]{roller2020recipes}
Stephen Roller, Emily Dinan, Naman Goyal, Da~Ju, Mary Williamson, Yinhan Liu, Jing Xu, Myle Ott, Kurt Shuster, Eric~M Smith, et~al.
\newblock Recipes for building an open-domain chatbot.
\newblock \emph{arXiv preprint arXiv:2004.13637}, 2020.

\bibitem[Scao et~al.(2023)Scao, Fan, Akiki, Pavlick, Ili{\'c}, Hesslow, Castagn{\'e}, Luccioni, Yvon, Gall{\'e}, Tow, Rush, Biderman, Webson, Ammanamanchi, Wang, Sagot, Muennighoff, del Moral, Ruwase, Bawden, Bekman, Mcmillan-Major, Beltagy, Nguyen, Saulnier, Tan, Ortiz~Suarez, Sanh, Lauren{\c c}on, Jernite, Launay, Mitchell, Raffel, Gokaslan, Simhi, Soroa, Aji, Alfassy, Rogers, Nitzav, Xu, Mou, Emezue, Klamm, Leong, van Strien, Adelani, Radev, Ponferrada, Levkovizh, Kim, Natan, de~Toni, Dupont, Kruszewski, Pistilli, Elsahar, Benyamina, Tran, Yu, Abdulmumin, Johnson, Gonzalez-Dios, de~la Rosa, Chim, Dodge, Zhu, Chang, Frohberg, Tobing, Bhattacharjee, Almubarak, Chen, Lo, von Werra, Weber, Phan, Allal, Tanguy, Dey, Mu{\~n}oz, Masoud, Grandury, {\v S}a{\v s}ko, Huang, Coavoux, Singh, Jiang, Vu, Jauhar, Ghaleb, Subramani, Kassner, Khamis, Nguyen, Espejel, de~Gibert, Villegas, Henderson, Colombo, Amuok, Lhoest, Harliman, Bommasani, L{\'o}pez, Ribeiro, Osei, Pyysalo, Nagel, Bose, Muhammad, Sharma, Longpre,
  Nikpoor, Silberberg, Pai, Zink, Torrent, Schick, Thrush, Danchev, Nikoulina, Laippala, Lepercq, Prabhu, Alyafeai, Talat, Raja, Heinzerling, Si, Salesky, Mielke, Lee, Sharma, Santilli, Chaffin, Stiegler, Datta, Szczechla, Chhablani, Wang, Pandey, Strobelt, Fries, Rozen, Gao, Sutawika, Bari, Al-Shaibani, Manica, Teehan, Albanie, Shen, Ben-David, Bach, Kim, Bers, Fevry, Neeraj, Thakker, Raunak, Tang, Yong, Sun, Brody, Uri, Tojarieh, Roberts, Chung, Tae, Phang, Press, Li, Narayanan, Bourfoune, Casper, Rasley, Ryabinin, Mishra, Zhang, Shoeybi, Peyrounette, Patry, Tazi, Sanseviero, von Platen, Cornette, Lavall{\'e}e, Lacroix, Rajbhandari, Gandhi, Smith, Requena, Patil, Dettmers, Baruwa, Singh, Cheveleva, Ligozat, Subramonian, N{\'e}v{\'e}ol, Lovering, Garrette, Tunuguntla, Reiter, Taktasheva, Voloshina, Bogdanov, Winata, Schoelkopf, Kalo, Novikova, Forde, Clive, Kasai, Kawamura, Hazan, Carpuat, Clinciu, Kim, Cheng, Serikov, Antverg, van~der Wal, Zhang, Zhang, Gehrmann, Pais, Shavrina, Scialom, Yun, Limisiewicz,
  Rieser, Protasov, Mikhailov, Pruksachatkun, Belinkov, Bamberger, Kasner, Rueda, Pestana, Feizpour, Khan, Faranak, Santos, Hevia, Unldreaj, Aghagol, Abdollahi, Tammour, Hajihosseini, Behroozi, Ajibade, Saxena, Ferrandis, Contractor, Lansky, David, Kiela, Nguyen, Tan, Baylor, Ozoani, Mirza, Ononiwu, Rezanejad, Jones, Bhattacharya, Solaiman, Sedenko, Nejadgholi, Passmore, Seltzer, Sanz, Dutra, Samagaio, Elbadri, Mieskes, Gerchick, Akinlolu, Mckenna, Qiu, Ghauri, Burynok, Abrar, Rajani, Elkott, Fahmy, Samuel, An, Kromann, Hao, Alizadeh, Shubber, Wang, Roy, Viguier, Le, Oyebade, Le, Yang, Nguyen, Kashyap, Palasciano, Callahan, Shukla, Miranda-Escalada, Singh, Beilharz, Wang, Brito, Zhou, Jain, Xu, Fourrier, Peri{\~n}{\'a}n, Molano, Yu, Manjavacas, Barth, Fuhrimann, Altay, Bayrak, Burns, Vrabec, Bello, Dash, Kang, Giorgi, Golde, Posada, Sivaraman, Bulchandani, Liu, Shinzato, de~Bykhovetz, Takeuchi, P{\`a}mies, Castillo, Nezhurina, S{\"a}nger, Samwald, Cullan, Weinberg, de~Wolf, Mihaljcic, Liu, Freidank, Kang,
  Seelam, Dahlberg, Broad, Muellner, Fung, Haller, Chandrasekhar, Eisenberg, Martin, Canalli, Su, Su, Cahyawijaya, Garda, Deshmukh, Mishra, Kiblawi, Ott, Sang-Aroonsiri, Kumar, Schweter, Bharati, Laud, Gigant, Kainuma, Kusa, Labrak, Bajaj, Venkatraman, Xu, Xu, Xu, Tan, Xie, Ye, Bras, Belkada, and Wolf]{bloom}
Teven~Le Scao, Angela Fan, Christopher Akiki, Ellie Pavlick, Suzana Ili{\'c}, Daniel Hesslow, Roman Castagn{\'e}, Alexandra~Sasha Luccioni, Fran{\c c}ois Yvon, Matthias Gall{\'e}, Jonathan Tow, Alexander~M. Rush, Stella Biderman, Albert Webson, Pawan~Sasanka Ammanamanchi, Thomas Wang, Beno{\^i}t Sagot, Niklas Muennighoff, Albert~Villanova del Moral, Olatunji Ruwase, Rachel Bawden, Stas Bekman, Angelina Mcmillan-Major, Iz~Beltagy, Huu Nguyen, Lucile Saulnier, Samson Tan, Pedro Ortiz~Suarez, Victor Sanh, Hugo Lauren{\c c}on, Yacine Jernite, Julien Launay, Margaret Mitchell, Colin Raffel, Aaron Gokaslan, Adi Simhi, Aitor Soroa, Alham~Fikri Aji, Amit Alfassy, Anna Rogers, Ariel~Kreisberg Nitzav, Canwen Xu, Chenghao Mou, Chris Emezue, Christopher Klamm, Colin Leong, Daniel van Strien, David~Ifeoluwa Adelani, Dragomir Radev, Eduardo~Gonz{\'a}lez Ponferrada, Efrat Levkovizh, Ethan Kim, Eyal~Bar Natan, Francesco de~Toni, G{\'e}rard Dupont, Germ{\'a}n Kruszewski, Giada Pistilli, Hady Elsahar, Hamza Benyamina, Hieu
  Tran, Ian Yu, Idris Abdulmumin, Isaac Johnson, Itziar Gonzalez-Dios, Javier de~la Rosa, Jenny Chim, Jesse Dodge, Jian Zhu, Jonathan Chang, J{\"o}rg Frohberg, Joseph Tobing, Joydeep Bhattacharjee, Khalid Almubarak, Kimbo Chen, Kyle Lo, Leandro von Werra, Leon Weber, Long Phan, Loubna~Ben Allal, Ludovic Tanguy, Manan Dey, Manuel~Romero Mu{\~n}oz, Maraim Masoud, Mar{\'i}a Grandury, Mario {\v S}a{\v s}ko, Max Huang, Maximin Coavoux, Mayank Singh, Mike Tian-Jian Jiang, Minh~Chien Vu, Mohammad~A. Jauhar, Mustafa Ghaleb, Nishant Subramani, Nora Kassner, Nurulaqilla Khamis, Olivier Nguyen, Omar Espejel, Ona de~Gibert, Paulo Villegas, Peter Henderson, Pierre Colombo, Priscilla Amuok, Quentin Lhoest, Rheza Harliman, Rishi Bommasani, Roberto~Luis L{\'o}pez, Rui Ribeiro, Salomey Osei, Sampo Pyysalo, Sebastian Nagel, Shamik Bose, Shamsuddeen~Hassan Muhammad, Shanya Sharma, Shayne Longpre, Somaieh Nikpoor, Stanislav Silberberg, Suhas Pai, Sydney Zink, Tiago~Timponi Torrent, Timo Schick, Tristan Thrush, Valentin Danchev,
  Vassilina Nikoulina, Veronika Laippala, Violette Lepercq, Vrinda Prabhu, Zaid Alyafeai, Zeerak Talat, Arun Raja, Benjamin Heinzerling, Chenglei Si, Elizabeth Salesky, Sabrina~J. Mielke, Wilson~Y. Lee, Abheesht Sharma, Andrea Santilli, Antoine Chaffin, Arnaud Stiegler, Debajyoti Datta, Eliza Szczechla, Gunjan Chhablani, Han Wang, Harshit Pandey, Hendrik Strobelt, Jason~Alan Fries, Jos Rozen, Leo Gao, Lintang Sutawika, M~Saiful Bari, Maged~S. Al-Shaibani, Matteo Manica, Ryan Teehan, Samuel Albanie, Sheng Shen, Srulik Ben-David, Stephen~H. Bach, Taewoon Kim, Tali Bers, Thibault Fevry, Trishala Neeraj, Urmish Thakker, Vikas Raunak, Xiangru Tang, Zheng-Xin Yong, Zhiqing Sun, Shaked Brody, Yallow Uri, Hadar Tojarieh, Adam Roberts, Hyung~Won Chung, Jaesung Tae, Jason Phang, Ofir Press, Conglong Li, Deepak Narayanan, Hatim Bourfoune, Jared Casper, Jeff Rasley, Max Ryabinin, Mayank Mishra, Minjia Zhang, Mohammad Shoeybi, Myriam Peyrounette, Nicolas Patry, Nouamane Tazi, Omar Sanseviero, Patrick von Platen, Pierre
  Cornette, Pierre~Fran{\c c}ois Lavall{\'e}e, R{\'e}mi Lacroix, Samyam Rajbhandari, Sanchit Gandhi, Shaden Smith, St{\'e}phane Requena, Suraj Patil, Tim Dettmers, Ahmed Baruwa, Amanpreet Singh, Anastasia Cheveleva, Anne-Laure Ligozat, Arjun Subramonian, Aur{\'e}lie N{\'e}v{\'e}ol, Charles Lovering, Dan Garrette, Deepak Tunuguntla, Ehud Reiter, Ekaterina Taktasheva, Ekaterina Voloshina, Eli Bogdanov, Genta~Indra Winata, Hailey Schoelkopf, Jan-Christoph Kalo, Jekaterina Novikova, Jessica~Zosa Forde, Jordan Clive, Jungo Kasai, Ken Kawamura, Liam Hazan, Marine Carpuat, Miruna Clinciu, Najoung Kim, Newton Cheng, Oleg Serikov, Omer Antverg, Oskar van~der Wal, Rui Zhang, Ruochen Zhang, Sebastian Gehrmann, Shani Pais, Tatiana Shavrina, Thomas Scialom, Tian Yun, Tomasz Limisiewicz, Verena Rieser, Vitaly Protasov, Vladislav Mikhailov, Yada Pruksachatkun, Yonatan Belinkov, Zachary Bamberger, Zden{\v e}k Kasner, Alice Rueda, Amanda Pestana, Amir Feizpour, Ammar Khan, Amy Faranak, Ana Santos, Anthony Hevia, Antigona
  Unldreaj, Arash Aghagol, Arezoo Abdollahi, Aycha Tammour, Azadeh Hajihosseini, Bahareh Behroozi, Benjamin Ajibade, Bharat Saxena, Carlos~Mu{\~n}oz Ferrandis, Danish Contractor, David Lansky, Davis David, Douwe Kiela, Duong~A. Nguyen, Edward Tan, Emi Baylor, Ezinwanne Ozoani, Fatima Mirza, Frankline Ononiwu, Habib Rezanejad, Hessie Jones, Indrani Bhattacharya, Irene Solaiman, Irina Sedenko, Isar Nejadgholi, Jesse Passmore, Josh Seltzer, Julio~Bonis Sanz, Livia Dutra, Mairon Samagaio, Maraim Elbadri, Margot Mieskes, Marissa Gerchick, Martha Akinlolu, Michael Mckenna, Mike Qiu, Muhammed Ghauri, Mykola Burynok, Nafis Abrar, Nazneen Rajani, Nour Elkott, Nour Fahmy, Olanrewaju Samuel, Ran An, Rasmus Kromann, Ryan Hao, Samira Alizadeh, Sarmad Shubber, Silas Wang, Sourav Roy, Sylvain Viguier, Thanh Le, Tobi Oyebade, Trieu Le, Yoyo Yang, Zach Nguyen, Abhinav~Ramesh Kashyap, Alfredo Palasciano, Alison Callahan, Anima Shukla, Antonio Miranda-Escalada, Ayush Singh, Benjamin Beilharz, Bo~Wang, Caio Brito, Chenxi Zhou,
  Chirag Jain, Chuxin Xu, Cl{\'e}mentine Fourrier, Daniel~Le{\'o}n Peri{\~n}{\'a}n, Daniel Molano, Dian Yu, Enrique Manjavacas, Fabio Barth, Florian Fuhrimann, Gabriel Altay, Giyaseddin Bayrak, Gully Burns, Helena~U. Vrabec, Imane Bello, Ishani Dash, Jihyun Kang, John Giorgi, Jonas Golde, Jose~David Posada, Karthik~Rangasai Sivaraman, Lokesh Bulchandani, Lu~Liu, Luisa Shinzato, Madeleine~Hahn de~Bykhovetz, Maiko Takeuchi, Marc P{\`a}mies, Maria~A Castillo, Marianna Nezhurina, Mario S{\"a}nger, Matthias Samwald, Michael Cullan, Michael Weinberg, Michiel de~Wolf, Mina Mihaljcic, Minna Liu, Moritz Freidank, Myungsun Kang, Natasha Seelam, Nathan Dahlberg, Nicholas~Michio Broad, Nikolaus Muellner, Pascale Fung, Patrick Haller, Ramya Chandrasekhar, Renata Eisenberg, Robert Martin, Rodrigo Canalli, Rosaline Su, Ruisi Su, Samuel Cahyawijaya, Samuele Garda, Shlok~S Deshmukh, Shubhanshu Mishra, Sid Kiblawi, Simon Ott, Sinee Sang-Aroonsiri, Srishti Kumar, Stefan Schweter, Sushil Bharati, Tanmay Laud, Th{\'e}o Gigant,
  Tomoya Kainuma, Wojciech Kusa, Yanis Labrak, Yash~Shailesh Bajaj, Yash Venkatraman, Yifan Xu, Yingxin Xu, Yu~Xu, Zhe Tan, Zhongli Xie, Zifan Ye, Mathilde Bras, Younes Belkada, and Thomas Wolf.
\newblock {BLOOM: A 176B-Parameter Open-Access Multilingual Language Model}.
\newblock working paper or preprint, November 2023.
\newblock URL \url{https://inria.hal.science/hal-03850124}.

\bibitem[Shannon(1948)]{shannon1948mathematical}
Claude~E Shannon.
\newblock A mathematical theory of communication.
\newblock \emph{The Bell system technical journal}, 27\penalty0 (3):\penalty0 379--423, 1948.

\bibitem[Shuster et~al.(2021)Shuster, Poff, Chen, Kiela, and Weston]{shuster2021retrieval}
Kurt Shuster, Spencer Poff, Moya Chen, Douwe Kiela, and Jason Weston.
\newblock Retrieval augmentation reduces hallucination in conversation.
\newblock \emph{arXiv preprint arXiv:2104.07567}, 2021.

\bibitem[Subramani \& Suresh(2020)Subramani and Suresh]{subramani2020discovering}
Nishant Subramani and Nivedita Suresh.
\newblock Discovering useful sentence representations from large pretrained language models.
\newblock \emph{arXiv preprint arXiv:2008.09049}, 2020.

\bibitem[Subramani et~al.(2022)Subramani, Suresh, and Peters]{subramani2022extracting}
Nishant Subramani, Nivedita Suresh, and Matthew~E Peters.
\newblock Extracting latent steering vectors from pretrained language models.
\newblock \emph{arXiv preprint arXiv:2205.05124}, 2022.

\bibitem[Subramani et~al.(2025)Subramani, Eisner, Svegliato, Van~Durme, Su, and Thomson]{mice_for_cats}
Nishant Subramani, Jason Eisner, Justin Svegliato, Benjamin Van~Durme, Yu~Su, and Sam Thomson.
\newblock {MICE} for {CAT}s: Model-internal confidence estimation for calibrating agents with tools.
\newblock In Luis Chiruzzo, Alan Ritter, and Lu~Wang (eds.), \emph{Proceedings of the 2025 Conference of the Nations of the Americas Chapter of the Association for Computational Linguistics: Human Language Technologies (Volume 1: Long Papers)}, pp.\  12362--12375, Albuquerque, New Mexico, April 2025. Association for Computational Linguistics.
\newblock ISBN 979-8-89176-189-6.
\newblock URL \url{https://aclanthology.org/2025.naacl-long.615/}.

\bibitem[Sun et~al.(2025)Sun, Zang, Zheng, Xu, Zhang, Yu, Song, and Li]{redeep}
ZhongXiang Sun, Xiaoxue Zang, Kai Zheng, Jun Xu, Xiao Zhang, Weijie Yu, Yang Song, and Han Li.
\newblock Rede{EP}: Detecting hallucination in retrieval-augmented generation via mechanistic interpretability.
\newblock In \emph{The Thirteenth International Conference on Learning Representations}, 2025.
\newblock URL \url{https://openreview.net/forum?id=ztzZDzgfrh}.

\bibitem[Team et~al.(2024)Team, Riviere, Pathak, Sessa, Hardin, Bhupatiraju, Hussenot, Mesnard, Shahriari, Ram{\'e}, et~al.]{team2024gemma}
Gemma Team, Morgane Riviere, Shreya Pathak, Pier~Giuseppe Sessa, Cassidy Hardin, Surya Bhupatiraju, L{\'e}onard Hussenot, Thomas Mesnard, Bobak Shahriari, Alexandre Ram{\'e}, et~al.
\newblock Gemma 2: Improving open language models at a practical size.
\newblock \emph{arXiv preprint arXiv:2408.00118}, 2024.

\bibitem[Tenney et~al.(2019)Tenney, Das, and Pavlick]{tenney2019bert}
Ian Tenney, Dipanjan Das, and Ellie Pavlick.
\newblock Bert rediscovers the classical nlp pipeline.
\newblock \emph{arXiv preprint arXiv:1905.05950}, 2019.

\bibitem[Tian et~al.(2023)Tian, Mitchell, Zhou, Sharma, Rafailov, Yao, Finn, and Manning]{tian2023just}
Katherine Tian, Eric Mitchell, Allan Zhou, Archit Sharma, Rafael Rafailov, Huaxiu Yao, Chelsea Finn, and Christopher~D Manning.
\newblock Just ask for calibration: Strategies for eliciting calibrated confidence scores from language models fine-tuned with human feedback.
\newblock \emph{arXiv preprint arXiv:2305.14975}, 2023.

\bibitem[Touvron et~al.(2023)Touvron, Martin, Stone, Albert, Almahairi, Babaei, Bashlykov, Batra, Bhargava, Bhosale, et~al.]{touvron2023llama}
Hugo Touvron, Louis Martin, Kevin Stone, Peter Albert, Amjad Almahairi, Yasmine Babaei, Nikolay Bashlykov, Soumya Batra, Prajjwal Bhargava, Shruti Bhosale, et~al.
\newblock Llama 2: Open foundation and fine-tuned chat models.
\newblock \emph{arXiv preprint arXiv:2307.09288}, 2023.

\bibitem[Turner et~al.(2023)Turner, Thiergart, Leech, Udell, Vazquez, Mini, and MacDiarmid]{turner2023steering}
Alexander~Matt Turner, Lisa Thiergart, Gavin Leech, David Udell, Juan~J Vazquez, Ulisse Mini, and Monte MacDiarmid.
\newblock Steering language models with activation engineering.
\newblock \emph{arXiv preprint arXiv:2308.10248}, 2023.

\bibitem[Vazhentsev et~al.(2023)Vazhentsev, Tsvigun, Vashurin, Petrakov, Vasilev, Panov, Panchenko, and Shelmanov]{vazhentsev2023efficient}
Artem Vazhentsev, Akim Tsvigun, Roman Vashurin, Sergey Petrakov, Daniil Vasilev, Maxim Panov, Alexander Panchenko, and Artem Shelmanov.
\newblock Efficient out-of-domain detection for sequence to sequence models.
\newblock In \emph{Findings of the Association for Computational Linguistics: ACL 2023}, pp.\  1430--1454, 2023.

\bibitem[Wadhwa et~al.(2024)Wadhwa, Seetharaman, Aggarwal, Ghosh, Basu, Srinivasan, Zhao, Chaudhari, and Aghazadeh]{rags_riches}
Hitesh Wadhwa, Rahul Seetharaman, Somyaa Aggarwal, Reshmi Ghosh, Samyadeep Basu, Soundararajan Srinivasan, Wenlong Zhao, Shreyas Chaudhari, and Ehsan Aghazadeh.
\newblock From rags to rich parameters: Probing how language models utilize external knowledge over parametric information for factual queries.
\newblock \emph{arXiv preprint arXiv:2406.12824}, 2024.

\bibitem[Wu et~al.(2024)Wu, Wu, and Zou]{Wu2024ClashEvalQT}
Kevin Wu, Eric Wu, and James Zou.
\newblock Clasheval: Quantifying the tug-of-war between an llm's internal prior and external evidence.
\newblock In \emph{Neural Information Processing Systems}, 2024.
\newblock URL \url{https://api.semanticscholar.org/CorpusID:269157310}.

\bibitem[Yang et~al.(2024)Yang, Yang, Zhang, Hui, Zheng, Yu, Li, Liu, Huang, Wei, et~al.]{yang2024qwen2}
An~Yang, Baosong Yang, Beichen Zhang, Binyuan Hui, Bo~Zheng, Bowen Yu, Chengyuan Li, Dayiheng Liu, Fei Huang, Haoran Wei, et~al.
\newblock Qwen2. 5 technical report.
\newblock \emph{arXiv preprint arXiv:2412.15115}, 2024.

\bibitem[Yin et~al.(2023{\natexlab{a}})Yin, Sun, Guo, Wu, Qiu, and Huang]{yin-etal-2023-large}
Zhangyue Yin, Qiushi Sun, Qipeng Guo, Jiawen Wu, Xipeng Qiu, and Xuanjing Huang.
\newblock Do large language models know what they don`t know?
\newblock In Anna Rogers, Jordan Boyd-Graber, and Naoaki Okazaki (eds.), \emph{Findings of the Association for Computational Linguistics: ACL 2023}, pp.\  8653--8665, Toronto, Canada, July 2023{\natexlab{a}}. Association for Computational Linguistics.
\newblock \doi{10.18653/v1/2023.findings-acl.551}.
\newblock URL \url{https://aclanthology.org/2023.findings-acl.551/}.

\bibitem[Yin et~al.(2023{\natexlab{b}})Yin, Sun, Guo, Wu, Qiu, and Huang]{yin2023large}
Zhangyue Yin, Qiushi Sun, Qipeng Guo, Jiawen Wu, Xipeng Qiu, and Xuanjing Huang.
\newblock Do large language models know what they don't know?
\newblock \emph{arXiv preprint arXiv:2305.18153}, 2023{\natexlab{b}}.

\bibitem[Yuksekgonul et~al.(2023)Yuksekgonul, Chandrasekaran, Jones, Gunasekar, Naik, Palangi, Kamar, and Nushi]{yuksekgonul2023attention}
Mert Yuksekgonul, Varun Chandrasekaran, Erik Jones, Suriya Gunasekar, Ranjita Naik, Hamid Palangi, Ece Kamar, and Besmira Nushi.
\newblock Attention satisfies: A constraint-satisfaction lens on factual errors of language models.
\newblock \emph{arXiv preprint arXiv:2309.15098}, 2023.

\bibitem[Zhang et~al.(2023)Zhang, Li, Das, Malin, and Kumar]{zhang-etal-2023-sac3}
Jiaxin Zhang, Zhuohang Li, Kamalika Das, Bradley Malin, and Sricharan Kumar.
\newblock {SAC}$^3$: Reliable hallucination detection in black-box language models via semantic-aware cross-check consistency.
\newblock In Houda Bouamor, Juan Pino, and Kalika Bali (eds.), \emph{Findings of the Association for Computational Linguistics: EMNLP 2023}, pp.\  15445--15458, Singapore, December 2023. Association for Computational Linguistics.
\newblock \doi{10.18653/v1/2023.findings-emnlp.1032}.
\newblock URL \url{https://aclanthology.org/2023.findings-emnlp.1032/}.

\bibitem[Zhao et~al.(2024)Zhao, Xu, Gupta, Asthana, Zheng, and Gould]{zhao2024first}
Qinyu Zhao, Ming Xu, Kartik Gupta, Akshay Asthana, Liang Zheng, and Stephen Gould.
\newblock The first to know: How token distributions reveal hidden knowledge in large vision-language models?
\newblock In \emph{European Conference on Computer Vision}, pp.\  127--142. Springer, 2024.

\bibitem[Zheng et~al.(2023)Zheng, Chiang, Sheng, Zhuang, Wu, Zhuang, Lin, Li, Li, Xing, et~al.]{zheng2023judging}
Lianmin Zheng, Wei-Lin Chiang, Ying Sheng, Siyuan Zhuang, Zhanghao Wu, Yonghao Zhuang, Zi~Lin, Zhuohan Li, Dacheng Li, Eric Xing, et~al.
\newblock Judging llm-as-a-judge with mt-bench and chatbot arena.
\newblock \emph{Advances in Neural Information Processing Systems}, 36:\penalty0 46595--46623, 2023.

\end{thebibliography}
\bibliographystyle{iclr2026_conference}

\appendix
% \section{Limitations}
% This study has the following limitations. Several of our analyses are performed either in aggregate across token positions or only at the first output token. It may be valuable to study variations across token positions, especially in settings with longer model outputs. Although the TriviaQA dataset covers diverse domains, we restrict our analysis to short-answer factual question-answering for computational feasibility. It is valuable to study the generalizability of our results to different tasks and domains; we leave this for future work. Second, we only experiment with instruction-tuned models. Investigating how pre-trained models differ from our analysis may reveal how post-training processes affect models' confidence estimates.

\appendix

\section{Limitations}
This study has the following limitations. Several of our analyses are performed either in aggregate across token positions or only at the first output token. It may be valuable to study variations across token positions, especially in settings with longer model outputs. Although the TriviaQA and MMLU datasets cover diverse domains, we restrict our analysis to short-answer factual question-answering for computational feasibility. It is valuable to study the generalizability of our results to different tasks and domains; we leave this for future work. Second, we only experiment with instruction-tuned models. Investigating how pre-trained models differ from our analysis may reveal how post-training processes affect models' confidence estimates.

\section{Methodology Details for RQ1}

\subsection{Prompting}
\label{appn:prompting_details}

Below are the prompts used for RQ1 to elicit confidence estimates from the model. We explore two prompting formats: \textit{Prompting without Answer} and \textit{Prompting with Answer}.

\paragraph{Prompting without Answer}

In this setting, the model is asked to estimate its confidence in being able to answer the question correctly, without actually providing an answer. The prompt is as follows:

\begin{lstlisting}
For the question below, output your confidence in your ability to generate the correct answer as an integer between 0 and 100, where 0 means complete uncertainty and 100 means complete certainty. Provide only the confidence score, without answering the question or including any explanations or additional text.

Question:
```
{Question}
```
\end{lstlisting}

\paragraph{Prompting with Answer}

This format uses a multi-turn chat interaction. First, the model generates an answer. Then, in a second turn, it is asked to provide a confidence score for the generated answer.

\textit{Turn 1:}

\begin{lstlisting}
Answer the following question with a single word or phrase. Do not provide explanations or additional context:

{Question}
\end{lstlisting}

\textit{Turn 2:}

\begin{lstlisting}
Please output your confidence in the correctness of the answer as an integer between 0 and 100, where 0 means complete uncertainty and 100 means complete certainty. Output only the confidence score with no explanations or additional text.
\end{lstlisting}

\subsection{Logit Lens and Tuned Lens}
\label{appn:logit_lens_tuned_lens}

\paragraph{Logit Lens:} \textit{Logit Lens} projects the intermediate hidden states directly into the vocabulary space via the unembedding matrix or final linear layer of the model~\citep{logitlens}. 
Concretely, given a hidden state $\mathbf{h}_\ell \in \mathbb{R}^d$ at layer $\ell$, and the unembedding matrix $\mathbf{W}_U \in \mathbb{R}^{V \times d}$ (where $V$ is the vocabulary size and $d$ is the hidden size), the layer-wise logits under the logit lens are computed as: $\mathbf{z}_\ell^{\text{(LL)}} = \mathbf{W}_U \mathbf{h}_\ell$. 
For decoding, these logits are passed through a softmax as normal to produce a probability distribution over the vocabulary: $\mathbf{p}_\ell^{\text{(LL)}} = \text{softmax}(\mathbf{z}_\ell^{\text{(LL)}}).$

\paragraph{Tuned Lens:} \textit{Tuned Lens} first learns an affine transformation or \textit{translator} for each layer and then projects the transformed hidden state directly to the vocabulary space via the unembedding matrix~\citep{tunedlens}.
Since pretrained language models are typically not trained to project from intermediate hidden states to the unembedding matrix, there is often a mismatch.
\textit{Tuned Lens} remediates this by post-hoc training these \textit{translators} at each layer, improving intermediate hidden states alignment in both magnitude and direction to the unembedding matrix.
For each layer $\ell$, the learned affine map $(\mathbf{A}_\ell, \mathbf{b}_\ell)$ transforms the hidden state $\mathbf{h}_\ell$ before multiplying with the unembedding matrix: $\mathbf{z}_\ell^{\text{(TL)}} = \mathbf{W}_U (\mathbf{A}_\ell \mathbf{h}_\ell + \mathbf{b}_\ell).$ 
The resulting logits can be passed through a softmax producing a probability distribution over the vocabulary: $\mathbf{p}_\ell^{\text{(TL)}} = \text{softmax}(\mathbf{z}_\ell^{\text{(TL)}}).$ The \textit{translators} are trained to minimize the KL divergence between $\mathbf{p}_\ell^{\text{(TL)}}$ and the model's final output distribution on general-purpose pretraining data.

\subsection{Input Feature Details for Logit Lens and Tuned Lens}
\label{appn:input_features}

\paragraph{Input Feature 1: Entropy} 
Entropy measures the uncertainty in a probability distribution.
For us, entropy estimates how peaky (or uniform) the distribution over the vocabulary is at each token position at each layer. 
Given a distribution over the vocabulary using either \textit{Logit Lens} (LL) or \textit{Tuned Lens} (TL), $\mathbf{p}_\ell^{\text{(LL|TL)}} = \{p_\ell^{\text{(LL|TL)}}(i)\}_{i=1}^V$ at layer $\ell$, the entropy $H_\ell$ is defined as:
\begin{equation}
H_\ell = - \sum_{i=1}^{V} p_\ell^{\text{(LL|TL)}}(i) \log p_\ell^{\text{(LL|TL)}}(i)
\end{equation}

Note: these estimates are computed independently for each layer and sequence position.%, capturing how peaked or uniform the model’s beliefs are.

\paragraph{Input Feature 2: Rank of Output Token}
A probability of 0.01 in certain token positions may reflect the highest rank token if the preceding context is ambiguous (e.g. The <next\_word>), whereas could reflect a much lower rank token if the context is less ambiguous (e.g. Harriet Tub<next\_word>)
The rank of the output next token at intermediate layers loosely correlates with the probability, but can offer a more direct signal towards what token a decoding algorithm may likely generate.
At layer $\ell$, we compute either the \textit{Logit Lens} (LL) or \textit{Tuned Lens} (TL) distribution $\mathbf{p}_\ell^{\text{(LL|TL)}}$. Let $\phi_\ell(t) := \log p_\ell^{\text{(LL|TL)}}(t)$ denote the log-probability of token $t$ under this distribution. We sort the vocabulary tokens by descending log-probability:

\begin{align}
\phi_\ell\big(\sigma(1)\big) \ge \phi_\ell\big(\sigma(2)\big) \ge \cdots \ge \phi_\ell\big(\sigma(V)\big)
% \log p_\ell^{\text{(LL|TL)}}\big(\sigma(1)\big) & \ge \log p_\ell^{\text{(LL|TL)}}\big(\sigma(2)\big) \\
% & \ge \cdots \ge \log p_\ell^{\text{(LL|TL)}}\big(\sigma(V)\big)
\end{align}

Then, the \textit{rank} $r_\ell(y)$ of the target token $y$ is the index $k$ such that $\sigma(k) = y$. A rank of $1$ indicates the model considers $y$ the most likely token at that layer. This is computed per layer and position.

%The rank of the target token (i.e., the correct next token from the final layer) at intermediate layers reflects how highly the model scores it. 
%At layer $\ell$, we compute the logit lens distribution $\mathbf{p}_\ell^{\text{(LL)}}$ and sort the vocabulary tokens by descending log-probability:
%\[
%\log p_\ell^{\text{(LL)}}\big(\sigma(1)\big) \ge \log p_\ell^{\text{(LL)}}\big(\sigma(2)\big) \ge \cdots \ge \log p_\ell^{\text{(LL)}}\big(\sigma(V)\big)
%\]
%Then, the \textit{rank} $r_\ell(y)$ of the target token $y$ is the index $k$ such that $\sigma(k) = y$. A rank of $1$ indicates the model considers $y$ the most likely token at that layer. This is computed per layer and position.

\paragraph{Input Feature 3: Top-$p$ Presence}
Nucleus-sampling is one of the most common decoding algorithms for language models~\citep{Holtzman2019TheCC}, where rather than considering the entire distribution over the vocabulary, one considers only the highest probability tokens until a top-$p$ cumulative probability.
This eliminates the chances of randomly sampling a very low probability token in the long-tail.
Motivated by this, we compute a binary indicator of whether the target token appears within the top-$p$ nucleus set at each layer. 
Let $\mathbf{p}_\ell^{\text{(LL|TL)}}$ be sorted in descending order as $p_{(1)} \ge p_{(2)} \ge \cdots$. The top-$p$ nucleus set $V_p$ is the smallest set such that:
\begin{equation}
\sum_{i=1}^{k} p_{(i)} \ge p
\end{equation}
The \textit{top-$p$ presence} indicator $I_{p,\ell}(y)$ is defined as:
\begin{equation}
I_{p,\ell}(y) = 
\begin{cases}
1 & \text{if } y \in V_p \\
0 & \text{otherwise}
\end{cases}
\end{equation}

\paragraph{Input Feature 4: Cross-Entropy} 
Cross-entropy quantifies how well the intermediate predictions match the output next token. 
At each layer $\ell$, it is computed using the log probability assigned to the target token $y$ under the logit lens or tuned lens distribution:
\begin{equation}
\text{CE}_\ell(y) = -\log p_\ell^{\text{(LL|TL)}}(y)
\end{equation}
This value is lower when the model assigns higher confidence to the output token. It is evaluated per layer and token position.

\section{Methodology Details for RQ2}

\subsection{Prompting}
\label{appn:prompting_rq2}

Below are the prompts used for RQ2 to elicit confidence estimates from the model. We explore two prompting formats: \textit{Prompting without Answer} and \textit{Prompting with Answer}.

\paragraph{Prompting without Answer}

In this setting, the model is provided with context and is asked to estimate its confidence in answering the question correctly, without actually generating an answer. The prompt used is as follows:

\begin{lstlisting}
For the question below, output your confidence in your ability to generate the correct answer as an integer between 0 and 100, where 0 means complete uncertainty and 100 means complete certainty. Provide only the confidence score, without answering the question or including any explanations or additional text.

Context:
```
{context}
```

Question:
```
{question}
```
\end{lstlisting}

\paragraph{Prompting with Answer}

This format uses a multi-turn chat interaction. In the first turn, the model is provided with context and asked to generate an answer to the question. In the second turn, it is asked to estimate its confidence in the answer it just provided.

\textit{Turn 1:}

\begin{lstlisting}
Using the following context, answer the question that follows with a single word or phrase. Do not provide explanations or additional context:

> Context:
{Context}

> Question:
{Question}
\end{lstlisting}

\textit{Turn 2:}

\begin{lstlisting}
Please output your confidence in the correctness of the answer as an integer between 0 and 100, where 0 means complete uncertainty and 100 means complete certainty. Output only the confidence score with no explanations or additional text.
\end{lstlisting}

% \subsection{ECS}
% \label{appn:ecs}

\subsection{Reasoning for Context Comparison}
\label{appn:prompt_rq2_explanation}

\paragraph{Correct vs. Incorrect Context} A correct and relevant context ($C_{\text{correct}}$) directly supports the correct answer with accurate factual information. This should increase the model's confidence, especially when its parametric knowledge alone is insufficient. In contrast, an incorrect but relevant context ($C_{\text{incorrect}}$) may appear plausible but contains factual errors. While it may influence the model's output, it should not increase confidence as much—especially if the model can detect inconsistencies or is sensitive to contradiction. Thus, we expect higher confidence in the presence of $C_{\text{correct}}$ than $C_{\text{incorrect}}$.

\paragraph{Correct vs. Irrelevant Context} An irrelevant context ($C_{\text{irrelevant}}$) is topically unrelated to the question and provides no useful information for answering it. Ideally, the model should recognize its lack of utility and ignore it, leading to little or no increase in confidence compared to the no-context baseline. In contrast, $C_{\text{correct}}$ provides directly useful information, so the model should become more confident in its answer. Therefore, the confidence gain should be higher with $C_{\text{correct}}$ than with $C_{\text{irrelevant}}$.

\section{Experiment Setup Details}

\subsection{Dataset Details}
\label{appn:additional_dataset_details}

\begin{table*}[ht]
\centering
\resizebox{\textwidth}{!}{
\begin{tabular}{p{10em} p{30em}}
\toprule
Question & Which island in Kent is the second largest of England's isles? \\
Correct Answers & "Shurland Hall", "Isle of Sheppy", "Shurland House", "Isle of Sheppey" \\
\midrule
Question & Rita Coolidge sang the title song for which Bond film? \\
Correct Answers & "Kamal kahn", "List of Bond girls in Octopussy", "Magda (James Bond)", "List of James Bond allies in Octopussy", "Vijay (James Bond)", "Bond 13", "Octopussy (character)", "Penelope Smallbone", "Octopussy", "General Orlov", "Kamal Khan", "Octopussy (film)", "List of James Bond villains in Octopussy", "Jim Fanning (James Bond)" \\
\midrule
% Question & Who was the Vanity Fair photographer responsible for the cover featuring a naked, pregnant Demi Moore? \\
% Correct Answers & "Annie Leibovitz", "ANNIE LEIBOVITZ", "Annie Liebowitz", "Annie Leibowitz", "Anna-Lou Leibovitz", "Annie Liebovitz" \\
% \midrule
% Question & Discovery, Gillyflower and Rome Beauty are types of which fruit? \\
% Correct Answers & "Apple Blossom", "Appleblossom", "Green Apples", "Malus domesticus", "Appleblossoms", "Culture of apple", "Malus domestica", "Apple blossom", "Apple/Nutritional information", "Nutritional information about the apple", "Apple production", "Apple-blossoms", "Apple blossoms", "Apple peel", "An apple a day", "Apple (fruit)", "Apple trees", "Malus pumila", "Malus communis", "Pyrus malus", "Apple tree", "Apple Popularity", "Apples", "Dried apple", "Apple (Fruit)", "Green Apple", "Apple-tree", "Green apples", "Apple", "Apples and teachers", "Aplle", "Apple-blossom", "Apple (tree)" \\
% \midrule
Question & What was invented by Jonas Hanway in the late 1750s? \\
Correct Answers & "Umbrella", "Umbrela", "History of the umbrella", "Unbrella", "Beach umbrella", "Parasols", "Windproof umbrella", "Beach parasol", "Parasol", "History of the Umbrella", "Umbrellas" \\
\bottomrule
\end{tabular}
}
\caption{Examples from the TriviaQA dataset.}
\label{tab:trivia_qa_example_full}
% \vspace{-1em}
\end{table*}

\begin{table*}[ht]
\centering
\resizebox{\textwidth}{!}{
\begin{tabular}{p{10em} p{30em}}
\toprule
Question & Which of the following bones develop by endochondral ossification? \\
Choices & A: "The ribs", B: "The ribs and sternum", C: "The ribs, sternum and clavicle", D: "The ribs, sternum, clavicle and vertebrae" \\
Correct Answer & B \\

\midrule

Question & Which of the following gives the total spin quantum number of the electrons in the ground state of neutral nitrogen (Z = 7)? \\
Choices & A: "1/2", B: "1", C: "3/2", D: "5/2" \\
Correct Answer & C \\

\midrule

Question & The Hawthorn Studies are most associated with which writer? \\
Choices & A: "Mary Parker Follett", B: "Elton Mayo", C: "Lillian Gilbreth", D: "Frederick Taylor" \\
Correct Answer & B \\
\bottomrule
\end{tabular}
}
\caption{Examples from the MMLU dataset.}
\label{tab:mmlu_example_full}
% \vspace{-1em}
\end{table*}

%This improves robustness to false negatives arising due to phrasing variations or failures in following the output format.

% \begin{table}[ht]
% \centering
% \resizebox{\textwidth}{!}{
% \begin{tabular}{p{10em} p{30em}}
% \toprule
% Question & Which island in Kent is the second largest of England's isles? \\
% Correct Answers & "Shurland Hall", "Isle of Sheppy", "Shurland House", "Isle of Sheppey" \\
% \bottomrule
% \end{tabular}
% }
% \caption{An example from the TriviaQA dataset.}
% \label{tab:trivia_qa_example}
% % \vspace{-1em}
% \end{table}

\paragraph{Quality Filtering:}
TriviaQA includes multiple retrieved Wikipedia or web documents as evidence for each example, which we repurpose to serve as (correct) context for RQ2 experiments.
While the documents are generally of high quality, manual inspection reveals a small percentage of examples where the documents do not contain enough information to answer the question.
As a result, we exclude any example where no retrieved document contains one of the ground-truth answers and at least 60\% of entities extracted from the question.\footnote{We use the FLAIR Sequence Tagger for entity extraction~\citep{akbik2019flair}.} This yields a final set of 11,683 examples, which we use for all our experiments. No quality filtering is necessary for MMLU. \cref{tab:trivia_qa_example_full} and \cref{tab:mmlu_example_full} present example questions along with their corresponding ground-truth answers from TriviaQA and MMLU.

\paragraph{External Context Construction:}
\label{appn:external_context}

Concatenating multiple long documents and analyzing these is challenging for interpretability methods that normally operate at the single-token or very few token level.
To address this, we select a single evidence document from TriviaQA that meets the quality filtering criteria described above. If the selected document exceeds 500 tokens, we use GPT-4o to summarize it down to 500 tokens, using the LLaMA-3-8B tokenizer for consistency. 
This summary is termed the ``correct'' context for the experiments for RQ2.
To generate ``incorrect'' contexts, we take the ``correct'' context for a given question $Q$ and prompt GPT-4o to replace all references to the ground-truth answer with plausible but incorrect alternatives.
Crucially, we keep the rest of the text unchanged, which makes the surrounding context relevant to $Q$.
We also obtain ``irrelevant'' contexts by sampling ``correct'' contexts from a different question such that the context has no lexical overlap with the ground-truth answer.
Additionally, we require that the SentenceBERT~\citep{reimers-2019-sentence-bert} embeddings using \texttt{all-MiniLM-L6-v2} of the candidate context have a cosine similarity $<0.3$ with the ``correct'' context for that question $Q$. This context construction process is grounded in the original ``correc'' context, and the LLM is used solely for straightforward tasks such as summarization and term replacement.

\paragraph{Prompting Formats:}

Below are the prompt formats used to generate answers from models on TriviaQA questions, either with or without supporting context.

\textit{Without context:}

\begin{lstlisting}
Answer the following question with a single word or phrase. Do not provide explanations or additional context:

{Question}
\end{lstlisting}

\textit{With context:}

\begin{lstlisting}
Using the following context, answer the question that follows with a single word or phrase. Do not provide explanations or additional context:

> Context:
{Context}

> Question:
{Question}
\end{lstlisting}

We use the following prompt format to generate answers from models for MMLU questions.

\begin{lstlisting}
The following is a multiple-choice question. Please choose the most suitable one among A, B, C and D as the answer to this question. Only output the choice identifier (A, B, C, or D) and nothing else. Do not provide explanations or additional context.

{Question}
\end{lstlisting}

\paragraph{Evaluation:} While the TriviaQA evaluation framework relies on exact match after text normalization between the candidate answer and one of the ground-truth answers, this rule-based approach can result in significant false negatives when the answer is correct but not phrased in exactly the same wording as the ground-truth answers. We use GPT-4o to assess correctness when exact match is not found due to its performance in factual QA evaluation \citep{zheng2023judging, gu2024survey}.
Specifically, if the generated answer does not exactly match any of the ground-truth answers, we prompt GPT-4o with the question $Q$, the ground-truth answers $A*$, and the generated answer $A$ and ask it to judge correctness. 
Unattempted questions are marked incorrect.
This improves robustness, especially because LLMs can phrase answers in slightly different but equivalent ways and may not exactly follow a given output format, especially when minimally post-trained. Since MMLU questions are multiple-choice, we ask the model to only output the choice identifier and rely on regex-based answer extraction and verification.

\subsection{RQ1 Setup Details}
\label{appn:methodology_rq1}

\paragraph{Logit-Lens and Tuned-Lens} We include all top-$p$ values from the set \{0.5, 0.9, 0.95, 0.99\} as input features when training our classifier using Logit-Lens and Tuned-Lens representations. We extract the statistics described in~\cref{appn:input_features} (entropy, rank of the correct token, top-$p$ presence, and cross-entropy loss) from the decoded logits for each model layer.
We then train our classifier $f$ using these features to predict whether a generated answer is correct or not.

\paragraph{Hidden States} We train a separate classifier on the hidden states extracted from each transformer layer during the generation of the first answer token. 
We focus on the first token because, due to the autoregressive nature of transformer models, the output at this initial step is conditioned solely on the input prompt and is unaffected by previously generated tokens~\citep{zhao2024first}. 
For each example, per-dimension mean and standard deviation values used for calculating z-scores are obtained by collecting the first-output-token hidden states from each layer for questions from an unused split of the TriviaQA and MMLU datasets. This avoids any sharing of information across examples.
% \footnote{We use the Wikipedia-Test split of TriviaQA, which contains 7701 examples. This split does not contain ground-truth answers and is not used elsewhere in our experiments. It is different from the split we use for testing our classifiers.
 % }

\paragraph{Classifier:} In practice, one could use any classifier here. 
~\citet{mice_for_cats} show that random forests are best able to recalibrate tool-using agents when using model internals as features in comparison to other simple classifiers such as logistic regression.
In our experiments, we use a random forest due to its efficacy, simplicity, and interpretability.
We use grid search and five-fold cross-validation over a small subset (1024 examples) of our training set to set hyperparameters for the random forest classifier. 

During hyperparameter tuning, a grid search was conducted over the following parameters for the Random Forest classifier: the number of estimators (\texttt{n\_estimators} $\in$ \{100, 200, 300\}), maximum tree depth (\texttt{max\_depth} $\in$ \{None, 10, 20, 30\}), minimum number of samples required to split an internal node (\texttt{min\_samples\_split} $\in$ \{2, 5\}), minimum number of samples required at a leaf node (\texttt{min\_samples\_leaf} $\in$ \{1, 2\}), the number of features considered for splitting at each node (\texttt{max\_features} $\in$ \{``sqrt'', ``log2'', None\}), and class weighting strategies (\texttt{class\_weight} $\in$ \{``balanced'', ``balanced\_subsample'', None\}). The hyperparameters were set to the following for random forest classifiers trained on Logit Lens, Tuned Lens, and PKS features: \texttt{n\_estimators} $=300$, \texttt{max\_depth} $=$ None, \texttt{max\_features} $=$ ``log2'', \texttt{min\_samples\_leaf} $=1$, \texttt{min\_samples\_split} $=2$, \texttt{class\_weight} $=$ ``balanced''. The hyperparameters were set to the following for random forest classifiers trained on hidden state features: \texttt{n\_estimators} $=300$, \texttt{max\_depth} $=10$, \texttt{max\_features} $=$ ``sqrt'', \texttt{min\_samples\_leaf} $=1$, \texttt{min\_samples\_split} $=2$, \texttt{class\_weight} $=$ ``balanced\_subsample''. 

\section{Additional Results for RQ1}

\subsection{Prompting}

\begin{figure}[ht]
\centering

% Row 1
\begin{minipage}[t]{0.47\linewidth}
\centering
\includegraphics[width=\linewidth]{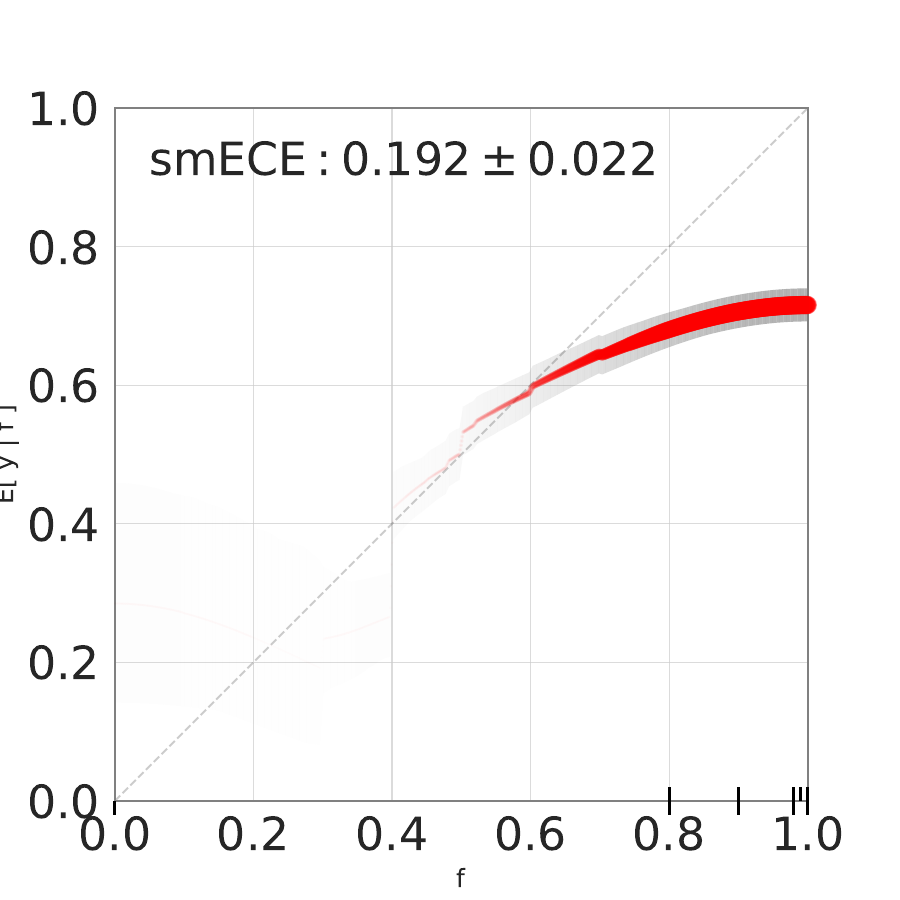}
\subcaption{LLaMA 3 8B}
\label{fig:smooth_ece_llama3_8b_with_answer}
\end{minipage}
\hfill
\begin{minipage}[t]{0.47\linewidth}
\centering
\includegraphics[width=\linewidth]{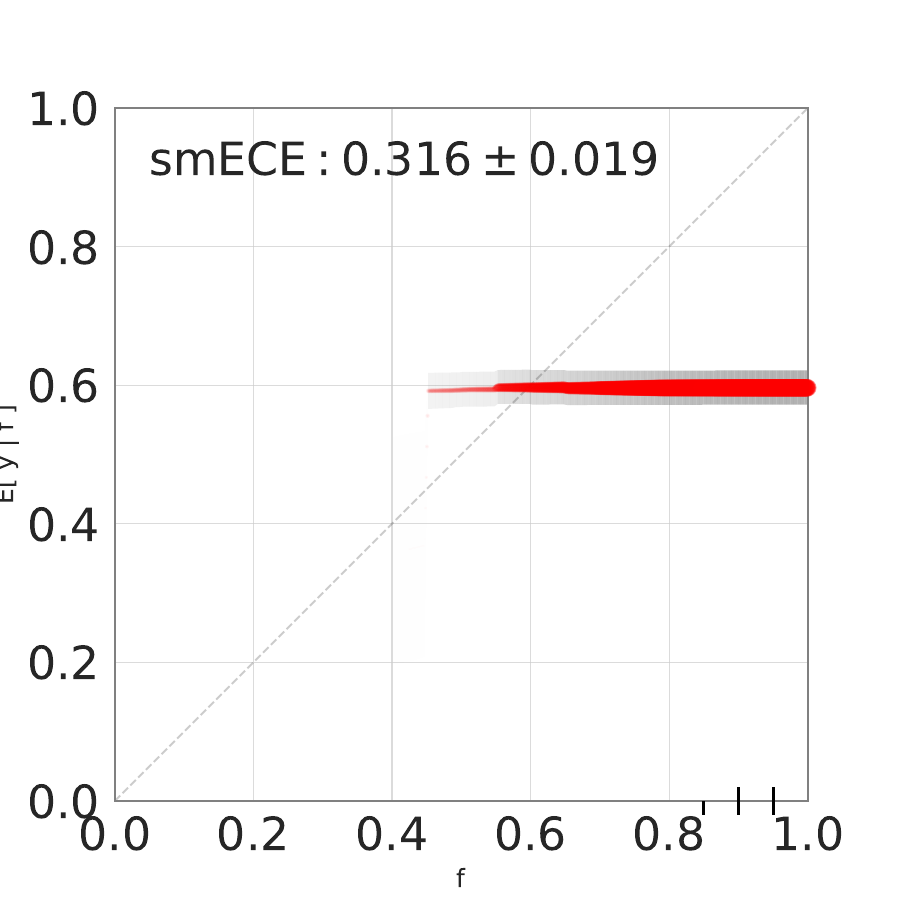}
\subcaption{LLaMA 2 13B}
\label{fig:smooth_ece_llama2_13b_with_answer}
\end{minipage}

\vspace{0.25cm}

% Row 2
\begin{minipage}[t]{0.47\linewidth}
\centering
\includegraphics[width=\linewidth]{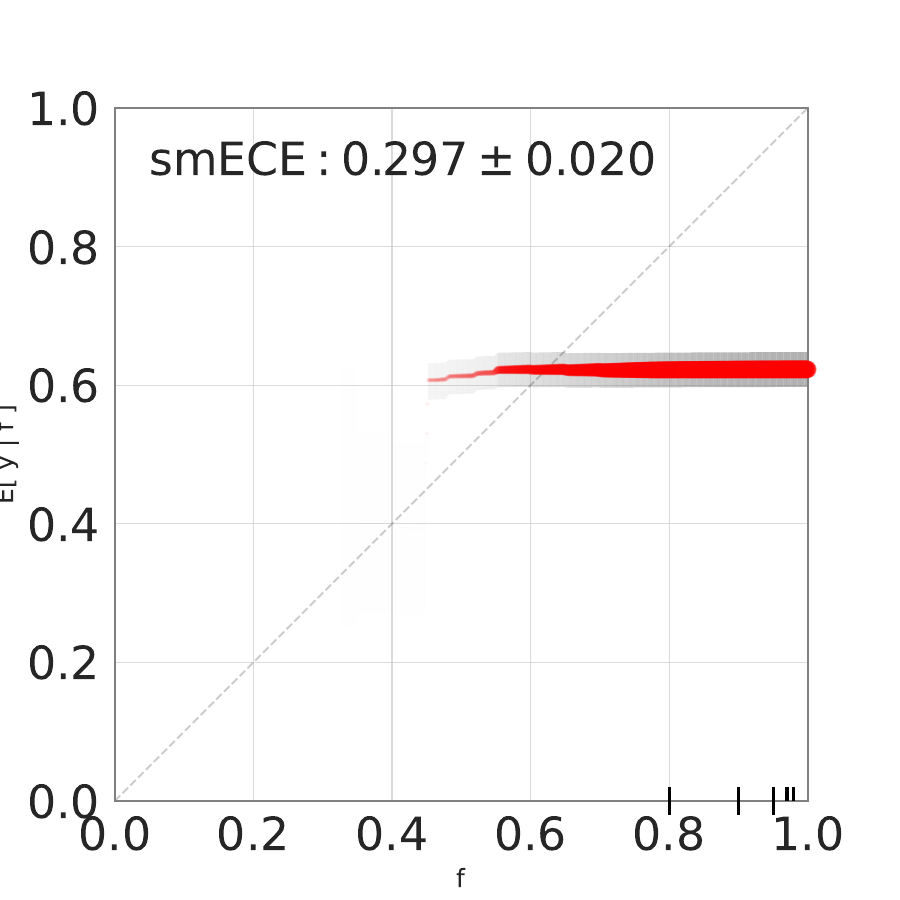}
\subcaption{LLaMA 2 7B}
\label{fig:smooth_ece_llama2_7b_with_answer}
\end{minipage}
\hfill
\begin{minipage}[t]{0.47\linewidth}
\centering
\includegraphics[width=\linewidth]{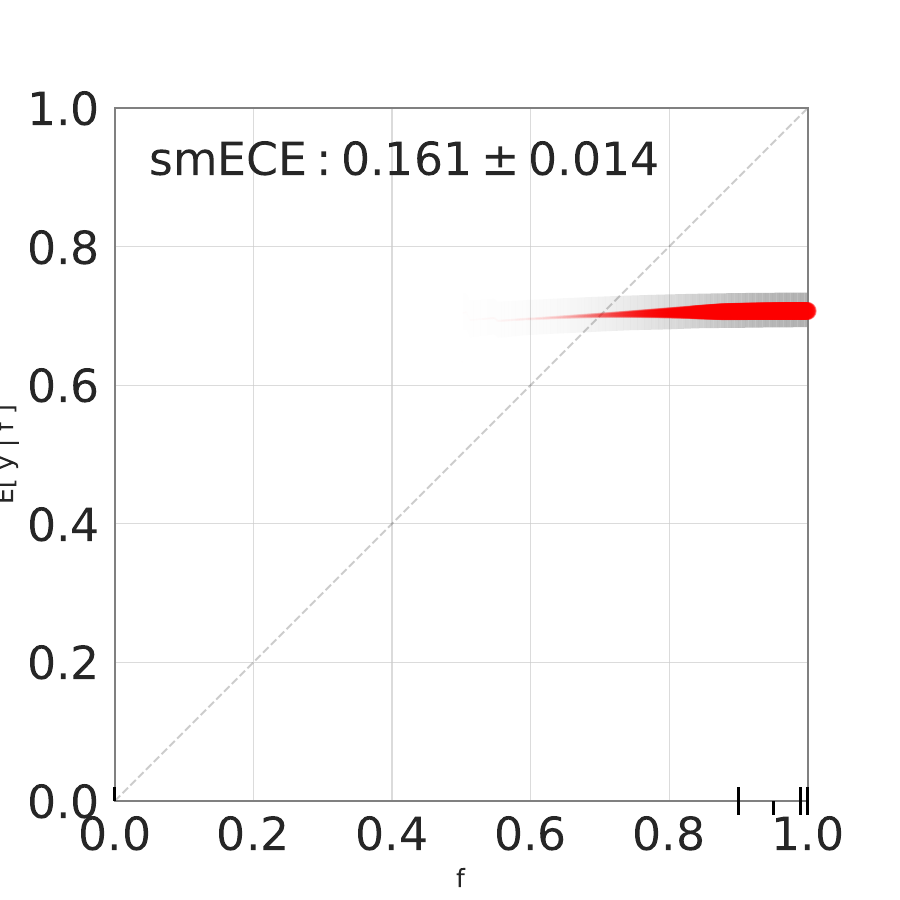}
\subcaption{Gemma2 9B}
\label{fig:smooth_ece_gemma2_9b_with_answer}
\end{minipage}

\vspace{0.25cm}

% Row 3
\begin{minipage}[t]{0.47\linewidth}
\centering
\includegraphics[width=\linewidth]{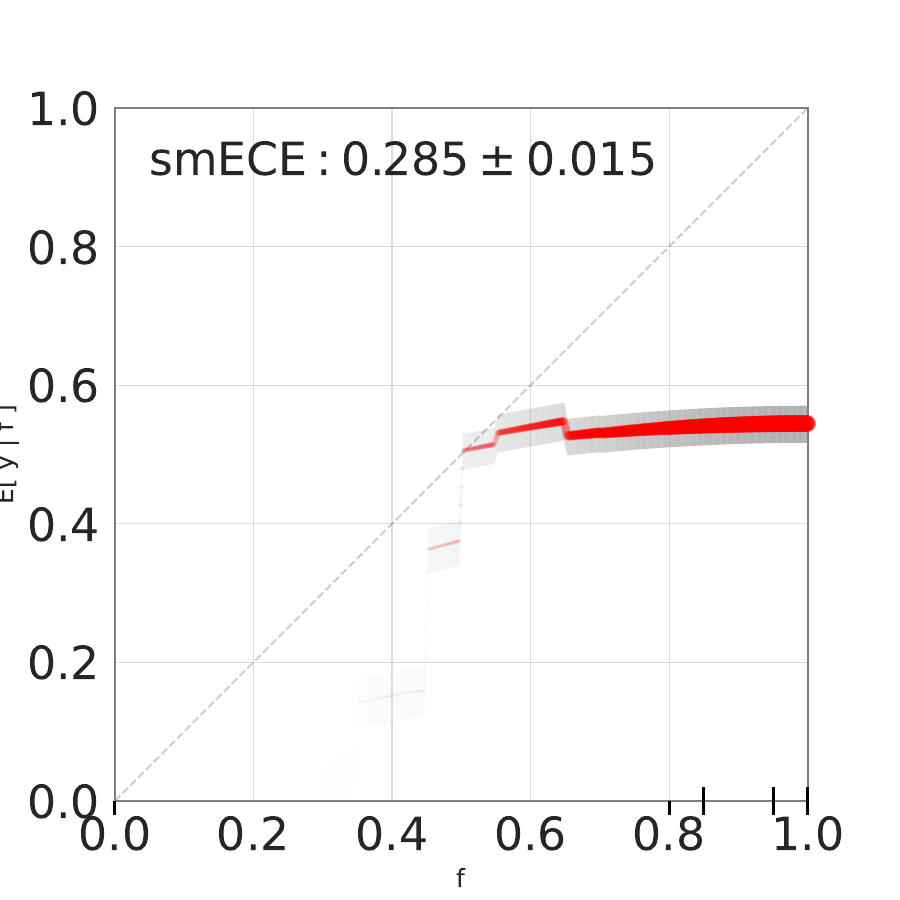}
\subcaption{Qwen 2.5 7B}
\label{fig:smooth_ece_qwen25_7b_with_answer}
\end{minipage}
\hfill
\begin{minipage}[t]{0.47\linewidth}
\centering
\includegraphics[width=\linewidth]{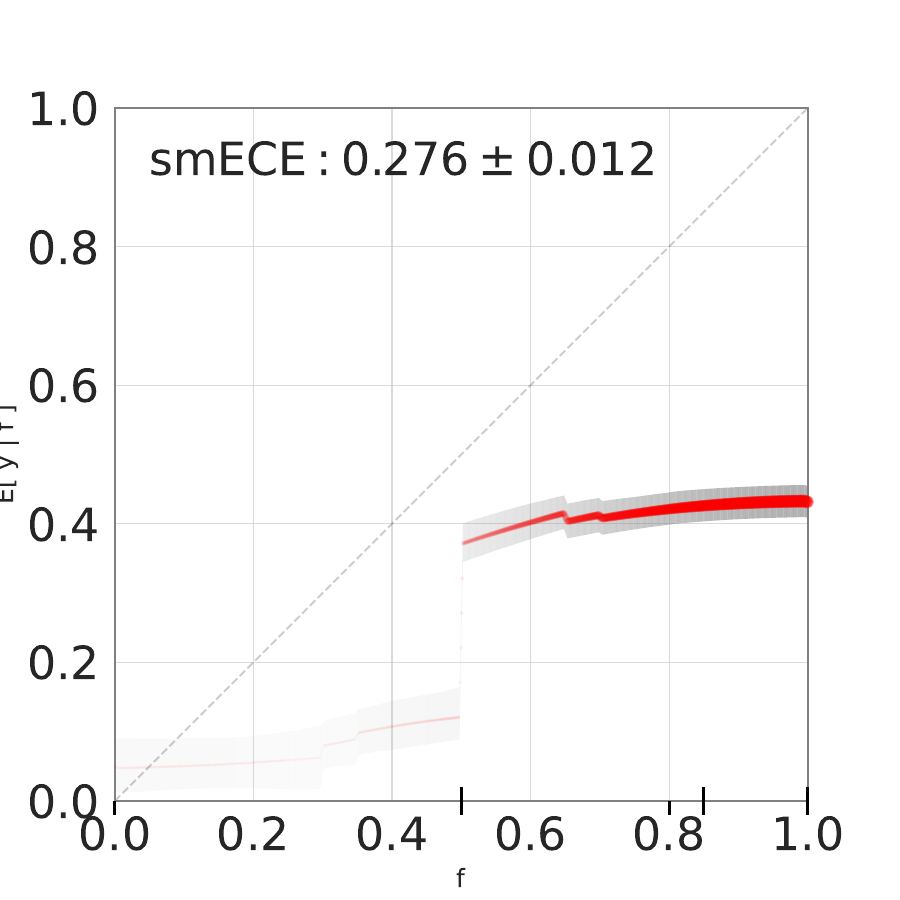}
\subcaption{Qwen 2.5 3B}
\label{fig:smooth_ece_qwen25_3b_with_answer}
\end{minipage}

\caption{Smooth ECE scores for prompting with answer on TriviaQA.}
\label{fig:smooth_ece_with_answer}
\end{figure}

\begin{figure}[ht]
\centering

% Row 1
\begin{minipage}[t]{0.47\linewidth}
\centering
\includegraphics[width=\linewidth]{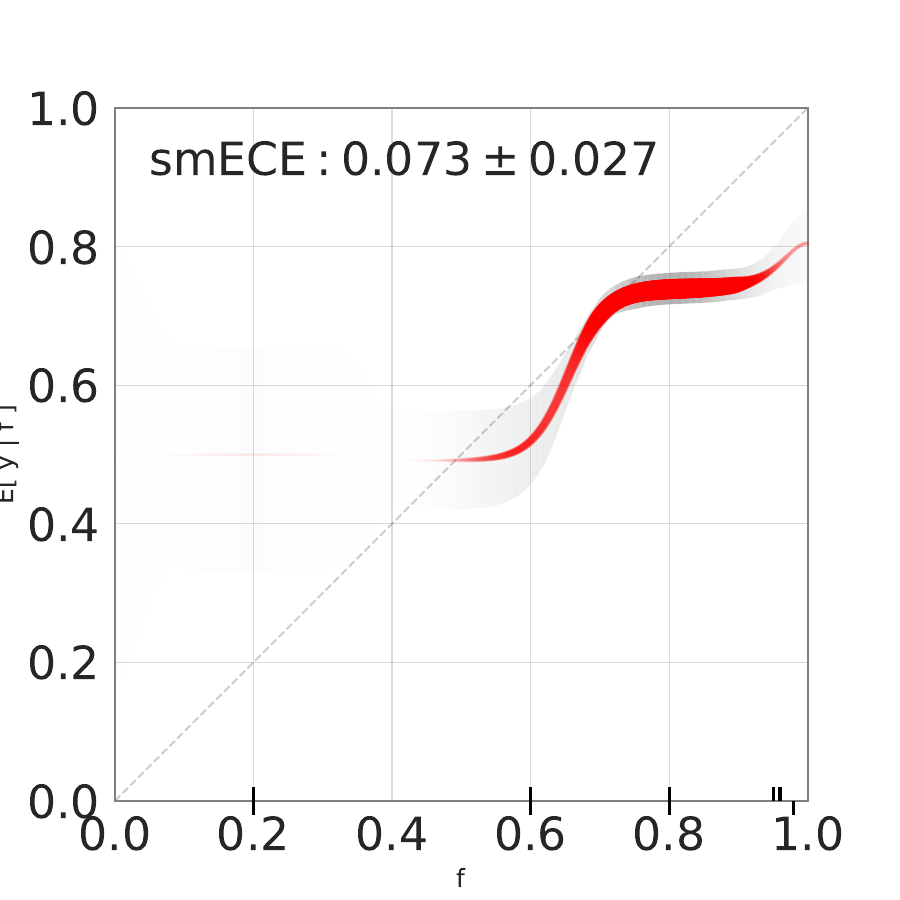}
\subcaption{LLaMA 3 8B}
\label{fig:smooth_ece_llama3_8b_without_answer}
\end{minipage}
\hfill
\begin{minipage}[t]{0.47\linewidth}
\centering
\includegraphics[width=\linewidth]{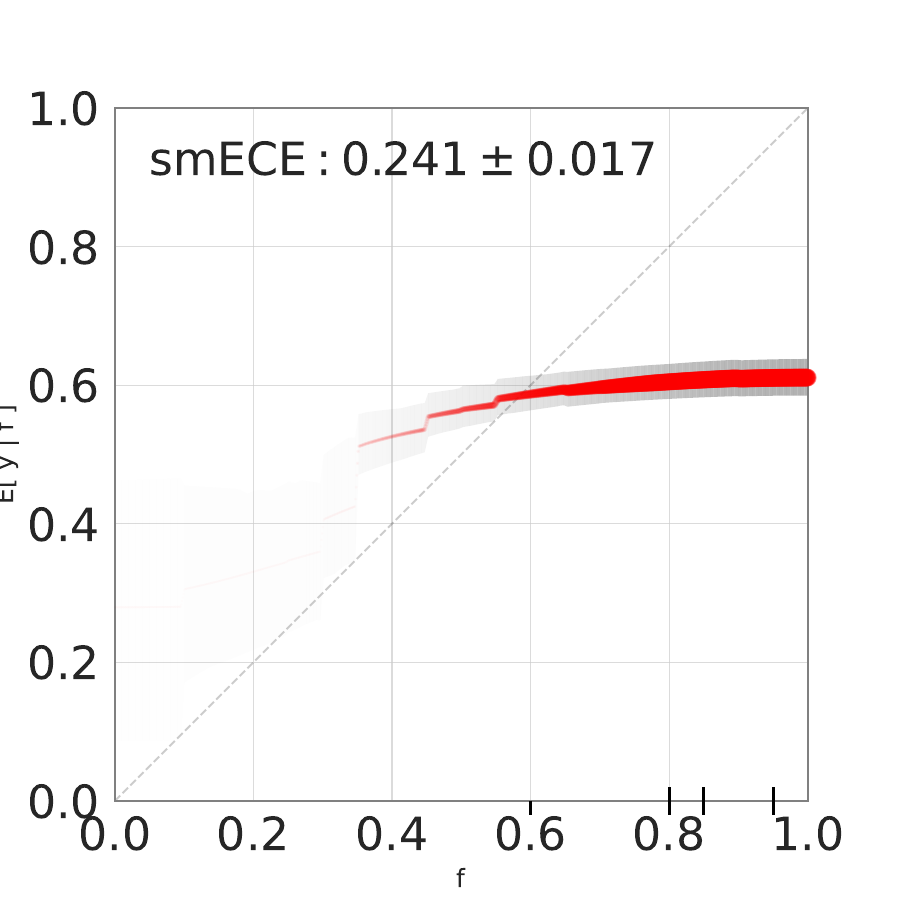}
\subcaption{LLaMA 2 13B}
\label{fig:smooth_ece_llama2_13b_without_answer}
\end{minipage}

\vspace{0.25cm}

% Row 2
\begin{minipage}[t]{0.47\linewidth}
\centering
\includegraphics[width=\linewidth]{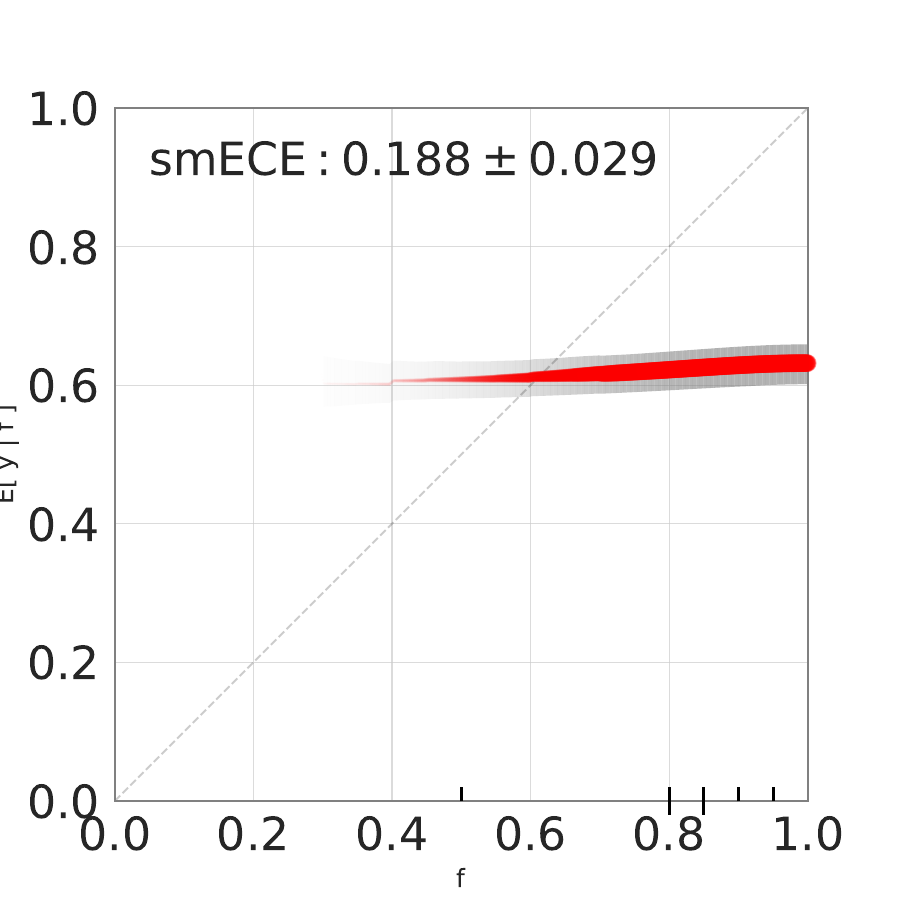}
\subcaption{LLaMA 2 7B}
\label{fig:smooth_ece_llama2_7b_without_answer}
\end{minipage}
\hfill
\begin{minipage}[t]{0.47\linewidth}
\centering
\includegraphics[width=\linewidth]{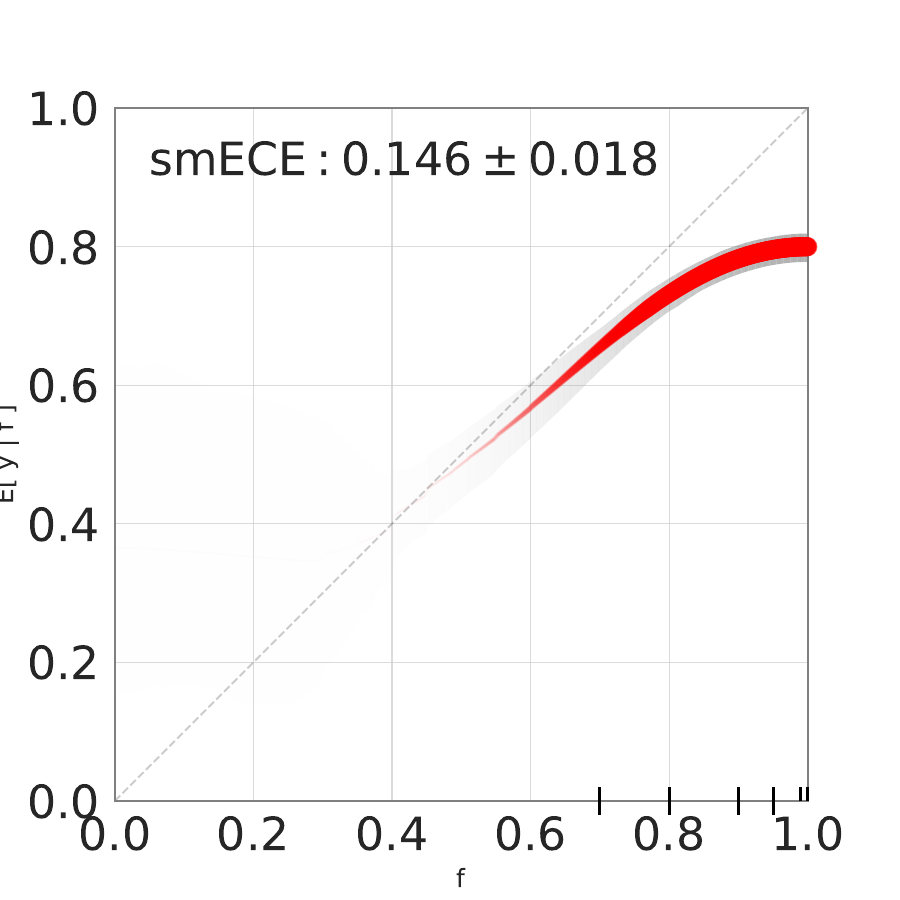}
\subcaption{Gemma2 9B}
\label{fig:smooth_ece_gemma2_9b_without_answer}
\end{minipage}

\vspace{0.25cm}

% Row 3
\begin{minipage}[t]{0.47\linewidth}
\centering
\includegraphics[width=\linewidth]{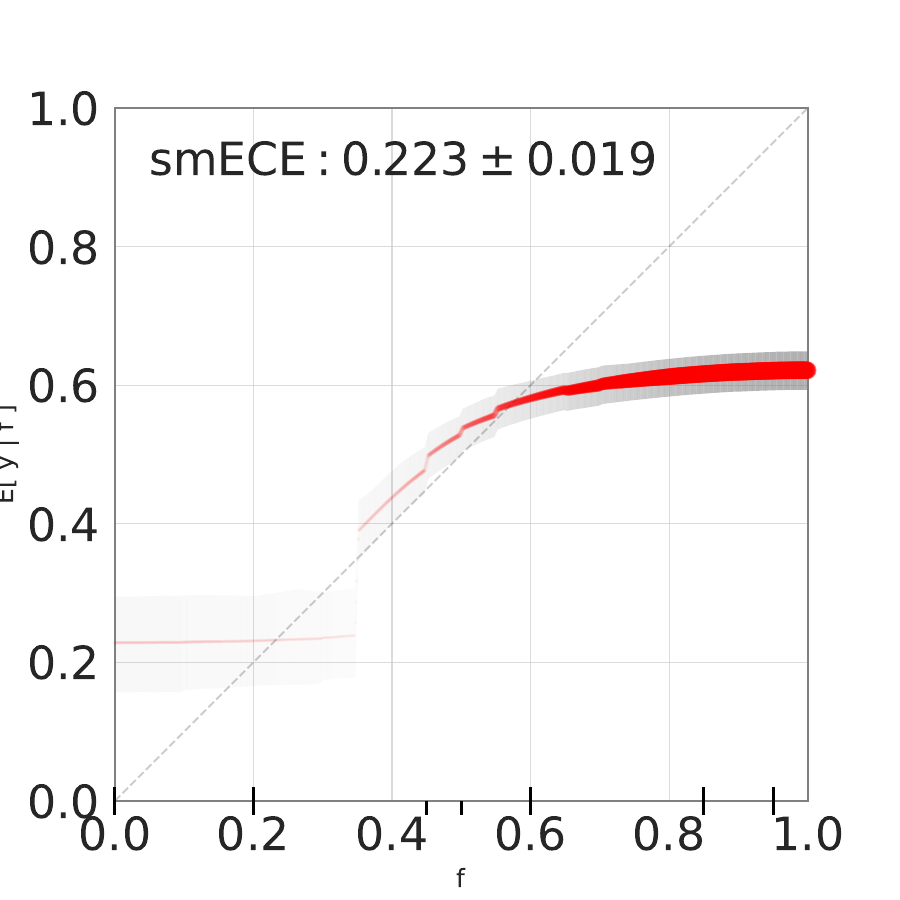}
\subcaption{Qwen 2.5 7B}
\label{fig:smooth_ece_qwen25_7b_without_answer}
\end{minipage}
\hfill
\begin{minipage}[t]{0.47\linewidth}
\centering
\includegraphics[width=\linewidth]{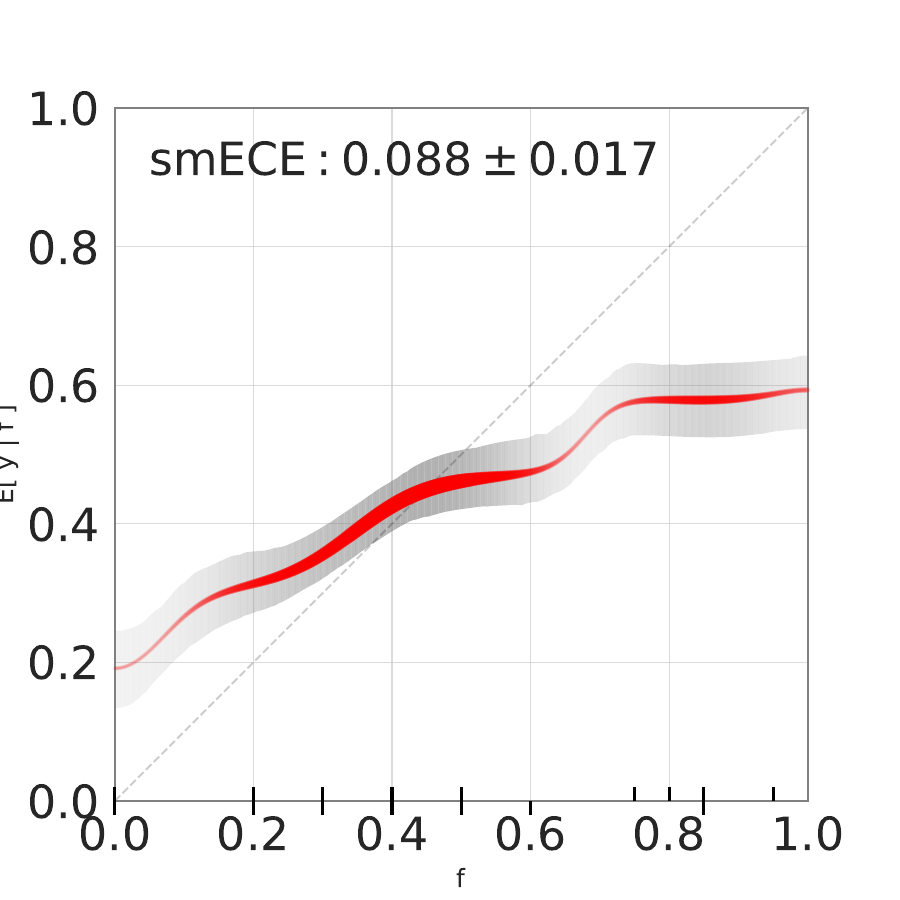}
\subcaption{Qwen 2.5 3B}
\label{fig:smooth_ece_qwen25_3b_without_answer}
\end{minipage}

\caption{Smooth ECE scores for prompting without answer on TriviaQA.}
\label{fig:smooth_ece_without_answer}

\end{figure}

\begin{figure}[ht]
\centering

% Row 1
\begin{minipage}[t]{0.47\linewidth}
\centering
\includegraphics[width=\linewidth]{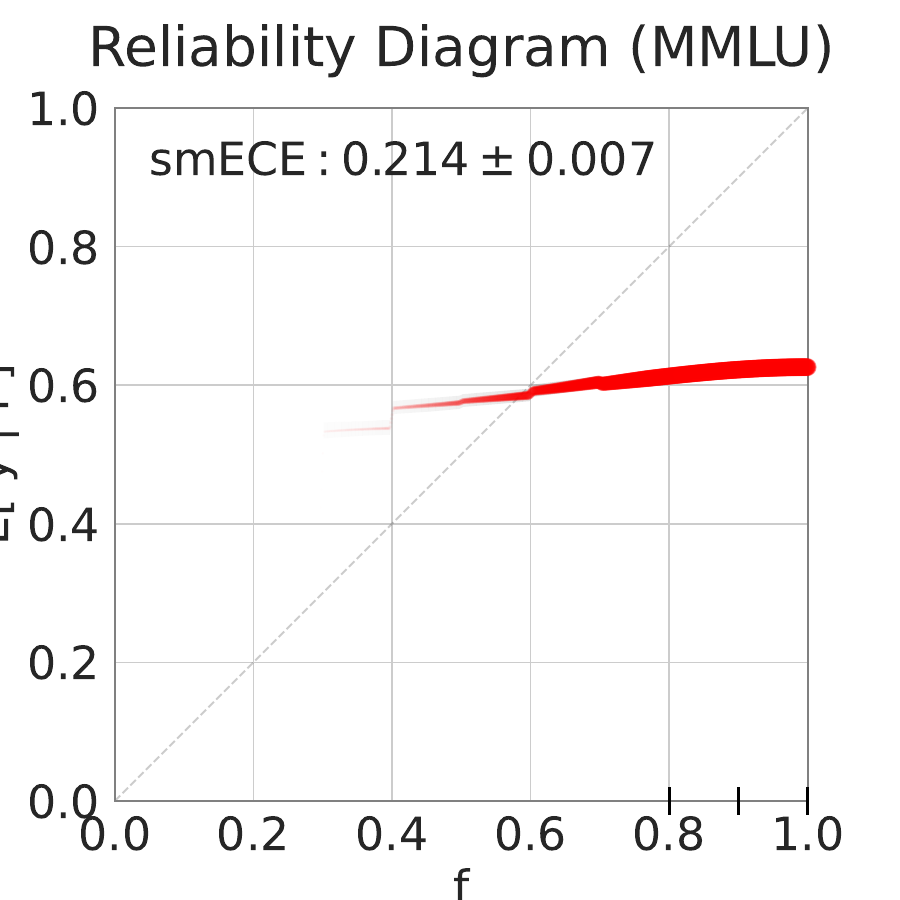}
\subcaption{LLaMA 3 8B}
\label{fig:smooth_ece_llama3_8b_with_answer_mmlu}
\end{minipage}
\hfill
\begin{minipage}[t]{0.47\linewidth}
\centering
\includegraphics[width=\linewidth]{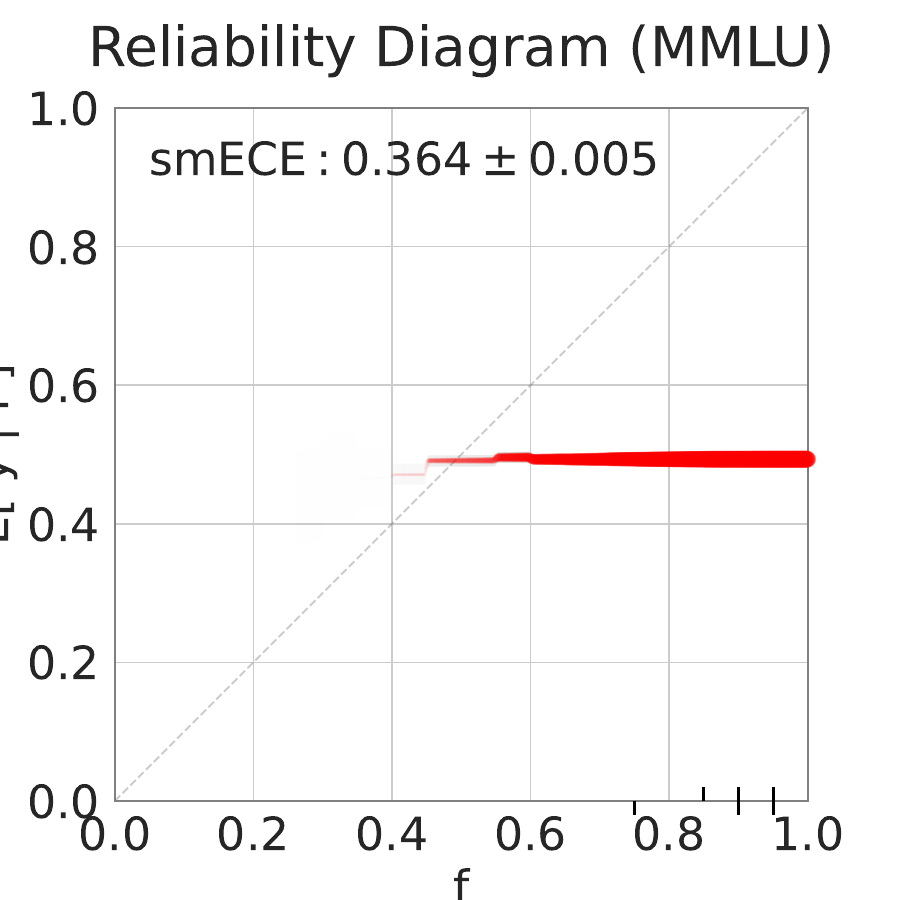}
\subcaption{LLaMA 2 13B}
\label{fig:smooth_ece_llama2_13b_with_answer_mmlu}
\end{minipage}

\vspace{0.25cm}

% Row 2
\begin{minipage}[t]{0.47\linewidth}
\centering
\includegraphics[width=\linewidth]{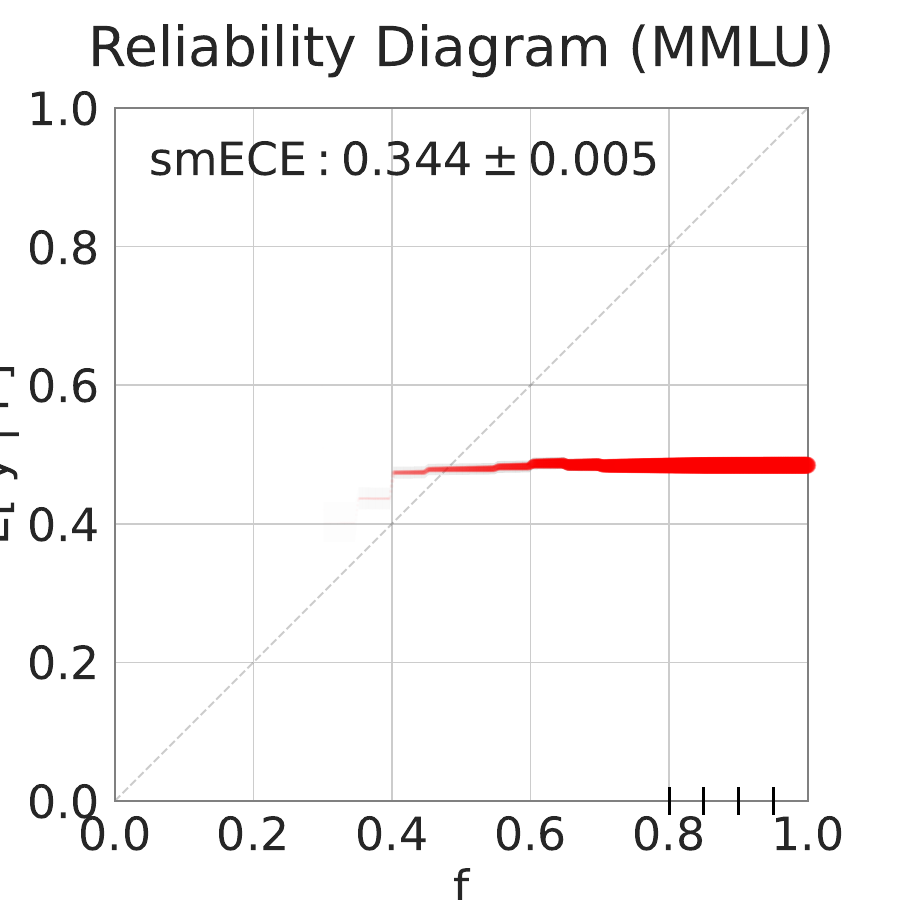}
\subcaption{LLaMA 2 7B}
\label{fig:smooth_ece_llama2_7b_with_answer_mmlu}
\end{minipage}
\hfill
\begin{minipage}[t]{0.47\linewidth}
\centering
\includegraphics[width=\linewidth]{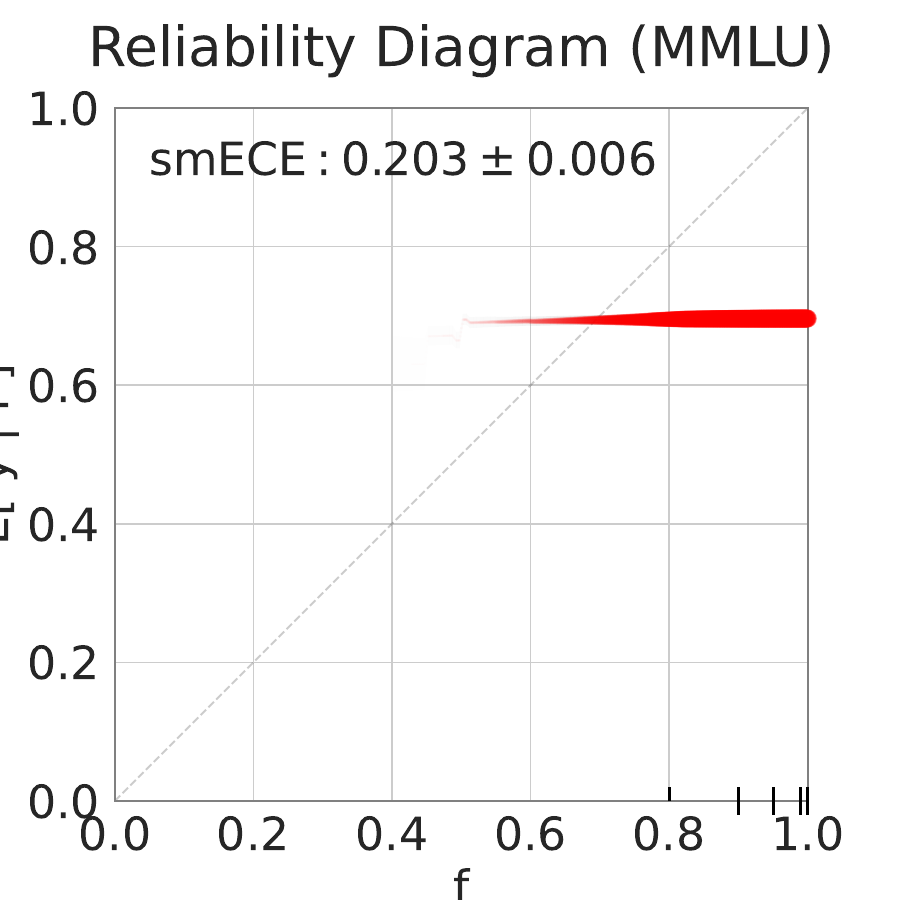}
\subcaption{Gemma2 9B}
\label{fig:smooth_ece_gemma2_9b_with_answer_mmlu}
\end{minipage}

\vspace{0.25cm}

% Row 3
\begin{minipage}[t]{0.47\linewidth}
\centering
\includegraphics[width=\linewidth]{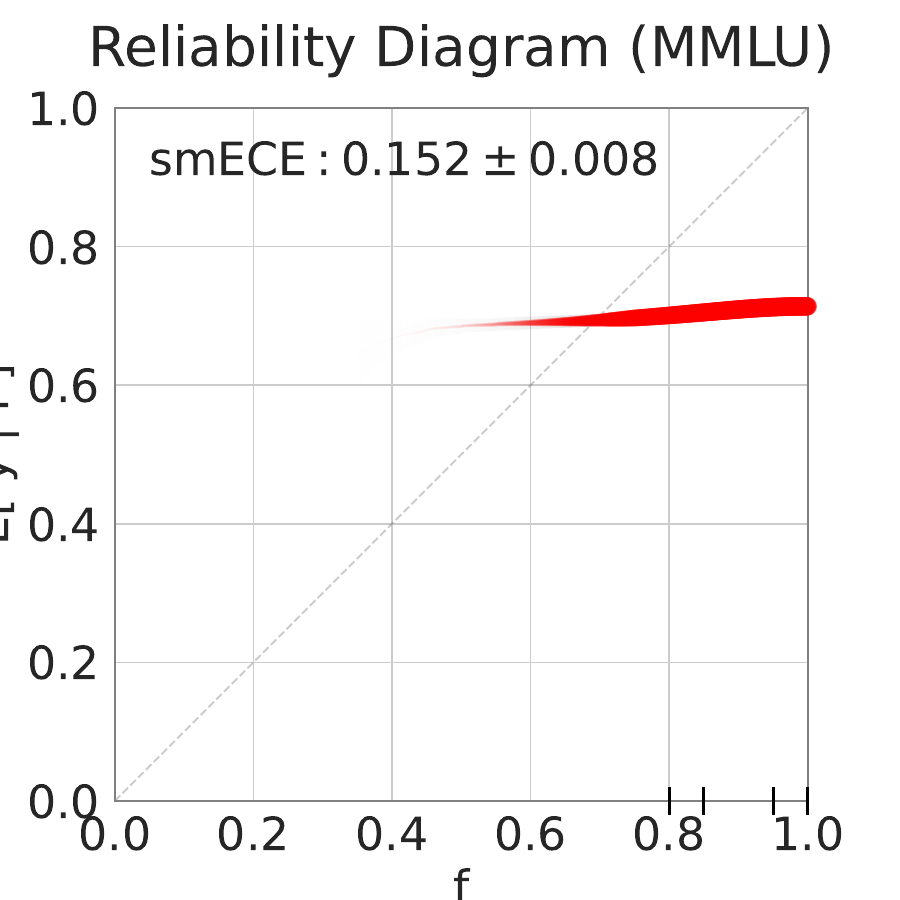}
\subcaption{Qwen 2.5 7B}
\label{fig:smooth_ece_qwen25_7b_with_answer_mmlu}
\end{minipage}
\hfill
\begin{minipage}[t]{0.47\linewidth}
\centering
\includegraphics[width=\linewidth]{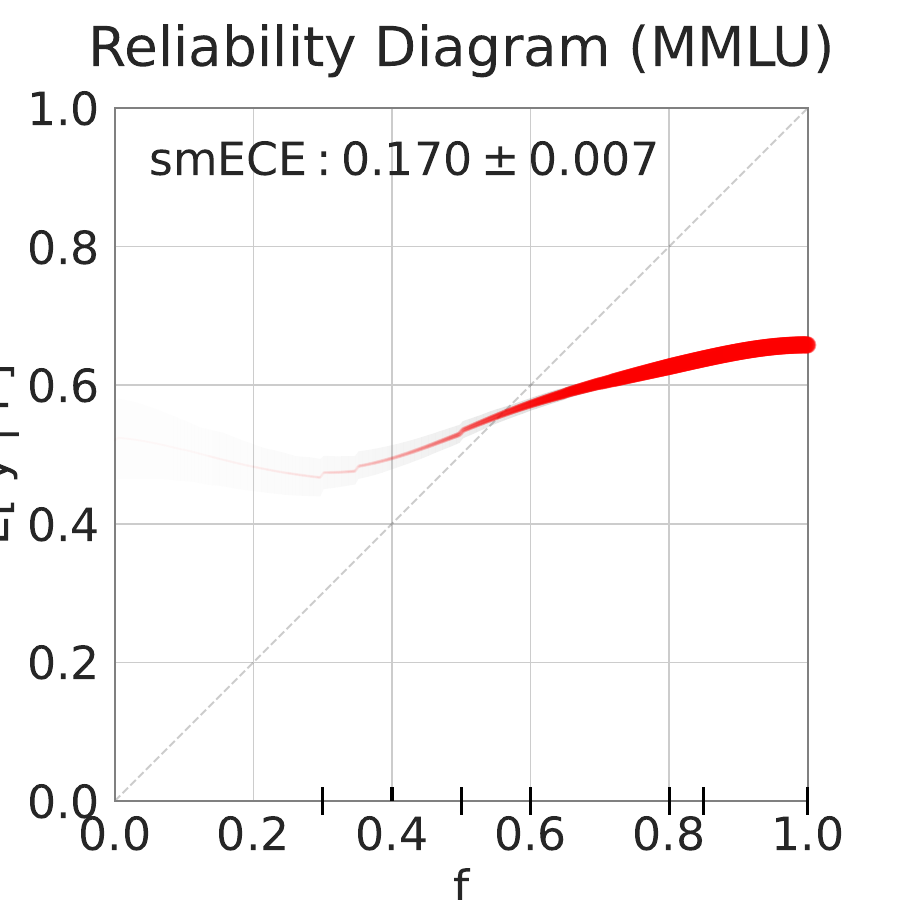}
\subcaption{Qwen 2.5 3B}
\label{fig:smooth_ece_qwen25_3b_with_answer_mmlu}
\end{minipage}

\caption{Smooth ECE scores for prompting with answer on MMLU.}
\label{fig:smooth_ece_with_answer_mmlu}

\end{figure}

\begin{figure}[ht]
\centering

% Row 1
\begin{minipage}[t]{0.47\linewidth}
\centering
\includegraphics[width=\linewidth]{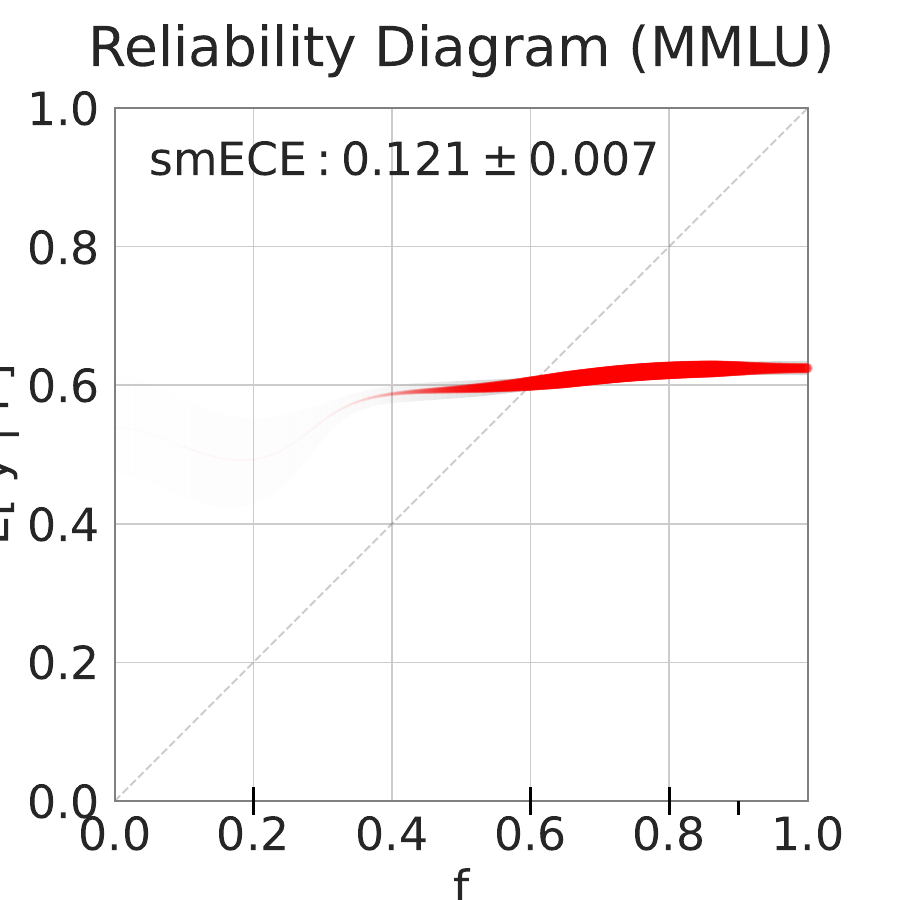}
\subcaption{LLaMA 3 8B}
\label{fig:smooth_ece_llama3_8b_without_answer_mmlu}
\end{minipage}
\hfill
\begin{minipage}[t]{0.47\linewidth}
\centering
\includegraphics[width=\linewidth]{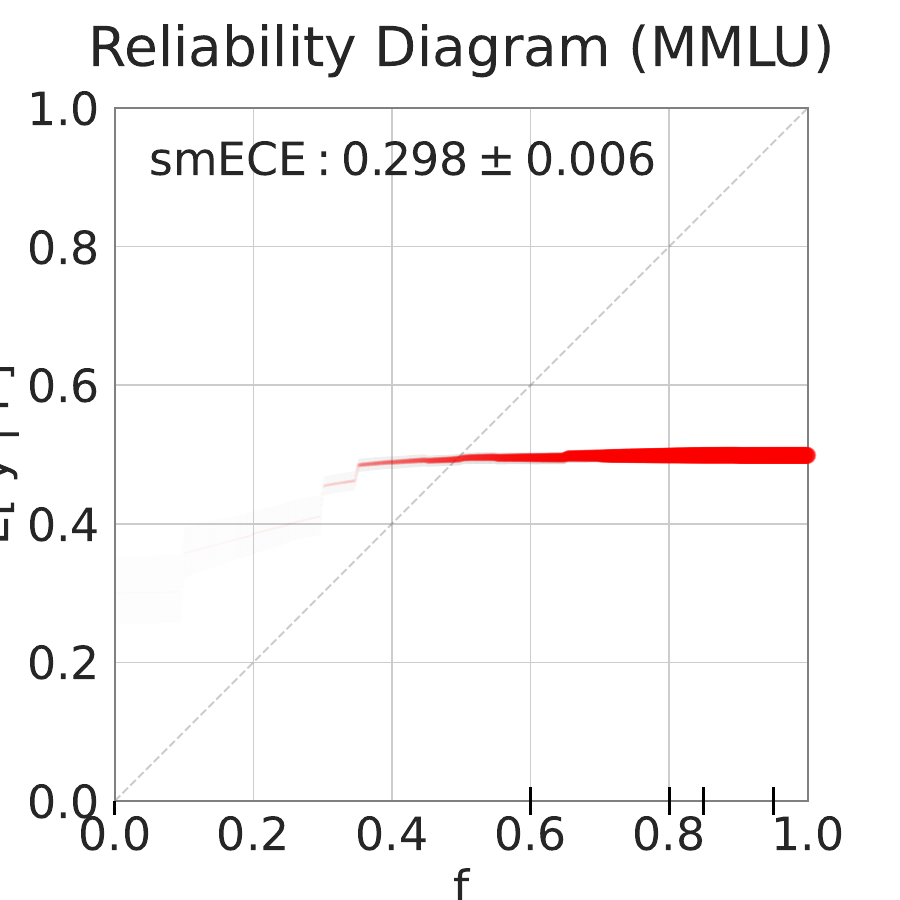}
\subcaption{LLaMA 2 13B}
\label{fig:smooth_ece_llama2_13b_without_answer_mmlu}
\end{minipage}

\vspace{0.25cm}

% Row 2
\begin{minipage}[t]{0.47\linewidth}
\centering
\includegraphics[width=\linewidth]{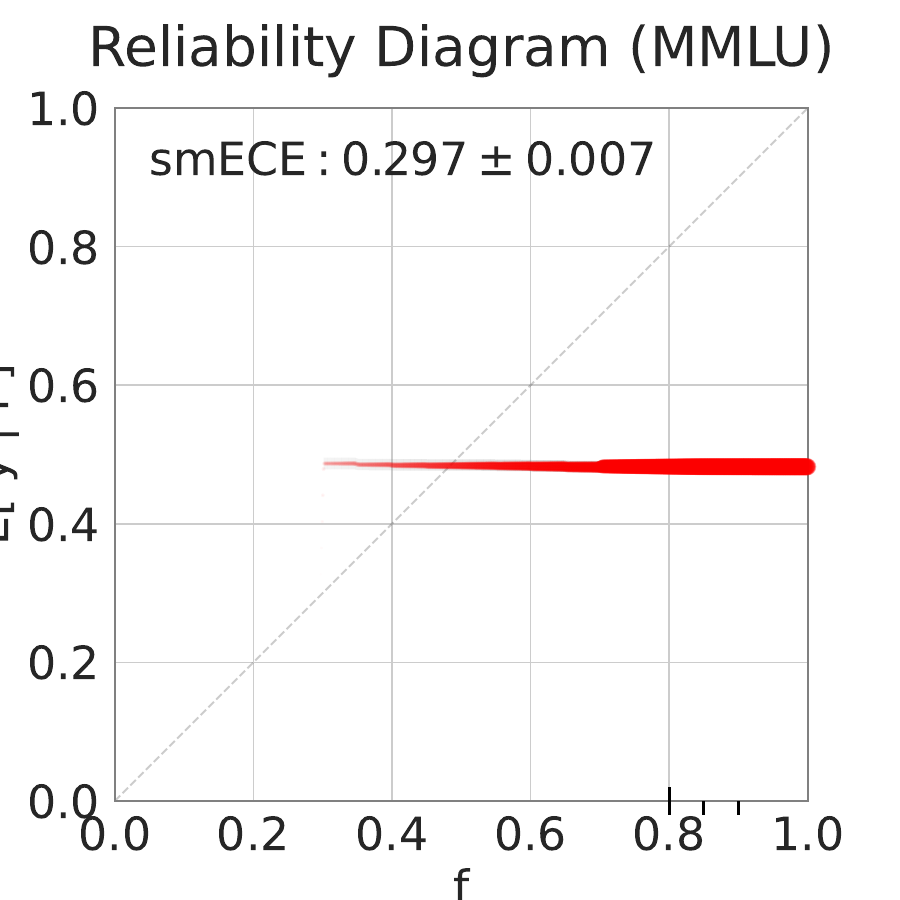}
\subcaption{LLaMA 2 7B}
\label{fig:smooth_ece_llama2_7b_without_answer_mmlu}
\end{minipage}
\hfill
\begin{minipage}[t]{0.47\linewidth}
\centering
\includegraphics[width=\linewidth]{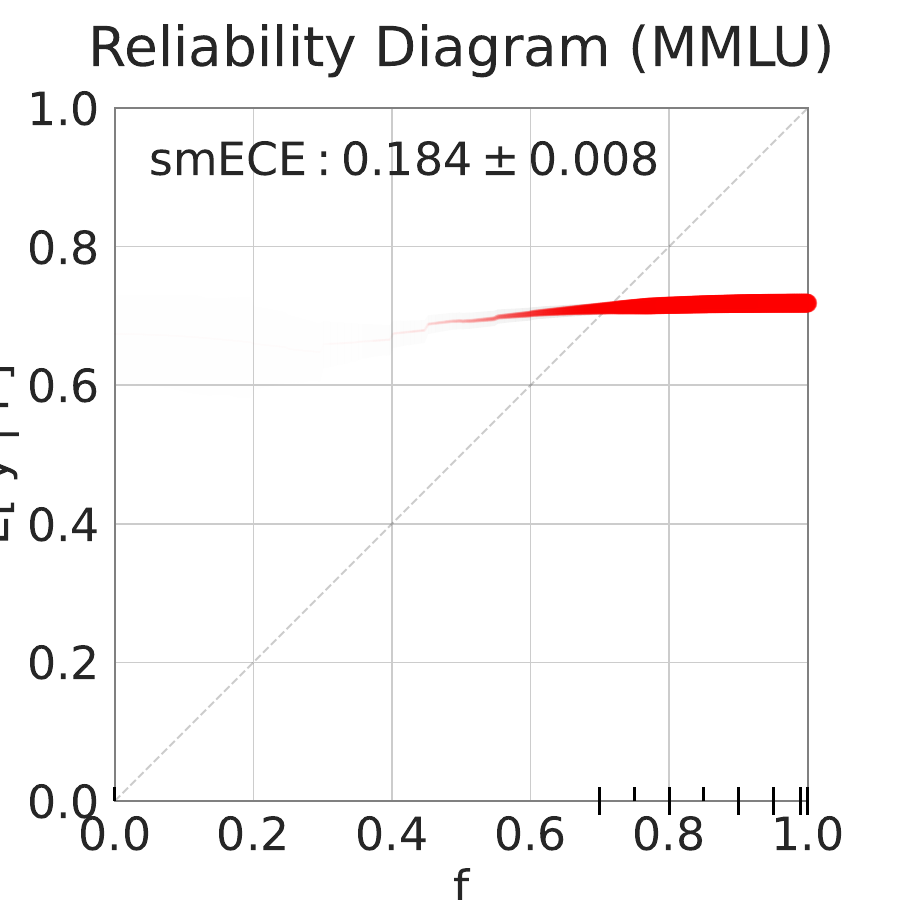}
\subcaption{Gemma2 9B}
\label{fig:smooth_ece_gemma2_9b_without_answer_mmlu}
\end{minipage}

\vspace{0.25cm}

% Row 3
\begin{minipage}[t]{0.47\linewidth}
\centering
\includegraphics[width=\linewidth]{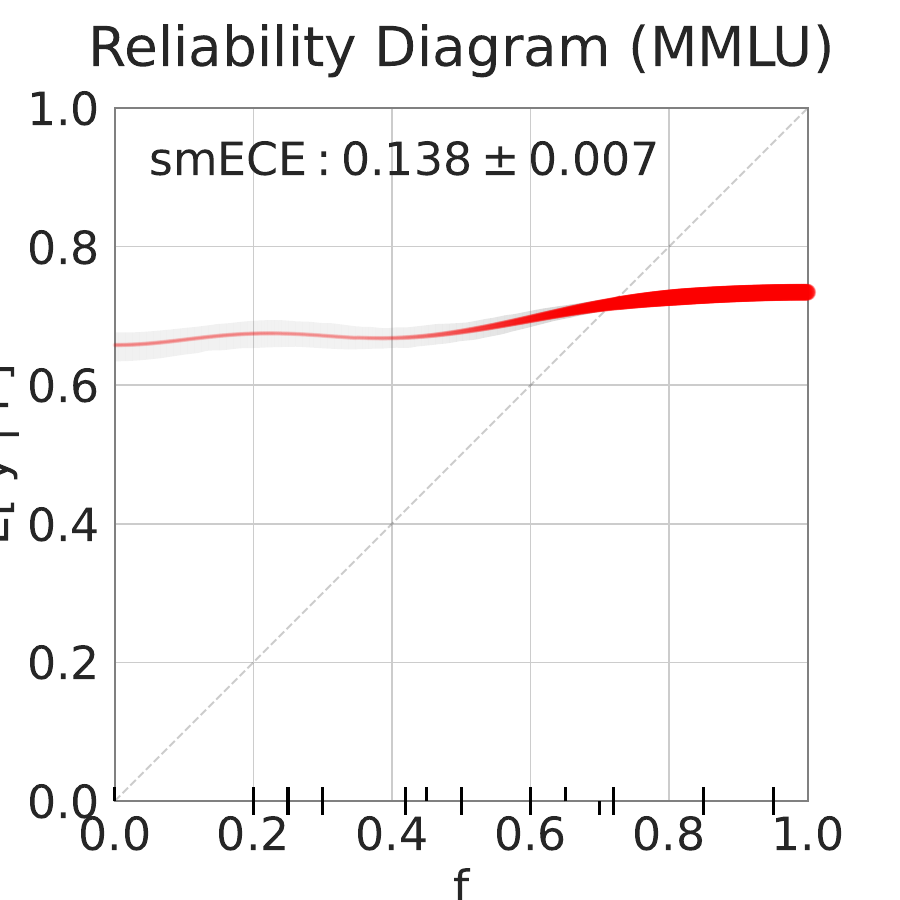}
\subcaption{Qwen 2.5 7B}
\label{fig:smooth_ece_qwen25_7b_without_answer_mmlu}
\end{minipage}
\hfill
\begin{minipage}[t]{0.47\linewidth}
\centering
\includegraphics[width=\linewidth]{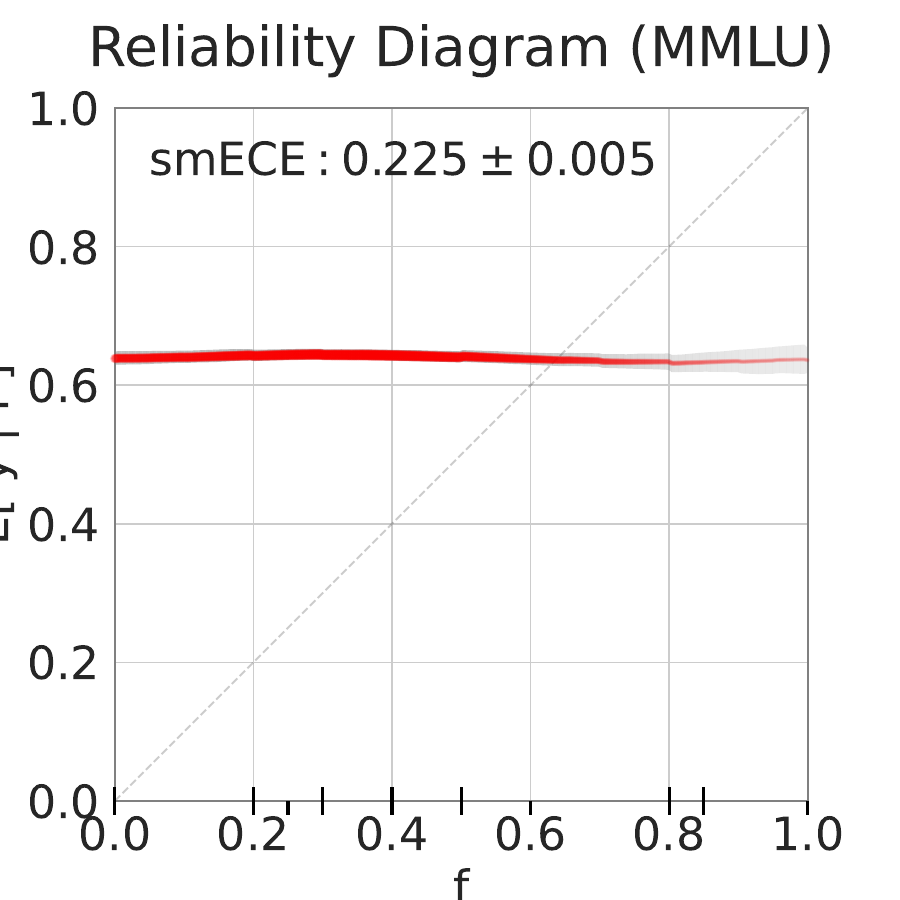}
\subcaption{Qwen 2.5 3B}
\label{fig:smooth_ece_qwen25_3b_without_answer_mmlu}
\end{minipage}

\caption{Smooth ECE scores for prompting without answer on MMLU.}
\label{fig:smooth_ece_without_answer_mmlu}

\end{figure}

For prompting, we evaluate all integer threshold values from 0 to 100 and select the one that yields the highest accuracy. Accordingly, the performance reported for prompting methods reflects the best-performing threshold. \Cref{fig:smooth_ece_with_answer} and \cref{fig:smooth_ece_with_answer_mmlu} present the Smooth ECE scores of various models using prompting with answers on TriviaQA and MMLU respectively, while \cref{fig:smooth_ece_without_answer} and \cref{fig:smooth_ece_without_answer_mmlu} show the scores for prompting without answers.

% Smooth ECE figures are shown here:
% \jivitesh{@jiarui could we refer to the figure directly in the main paper and remove reference text/headers like this from the Appendix, because latex will place the figure wherever it wants.}\jiarui{yes haven't reached here yet}

\subsection{Logit Lens}

\paragraph{Logit Lens VS. Tuned Lens}

\begin{figure}[ht]
\centering
% First row
\includegraphics[width=\linewidth]{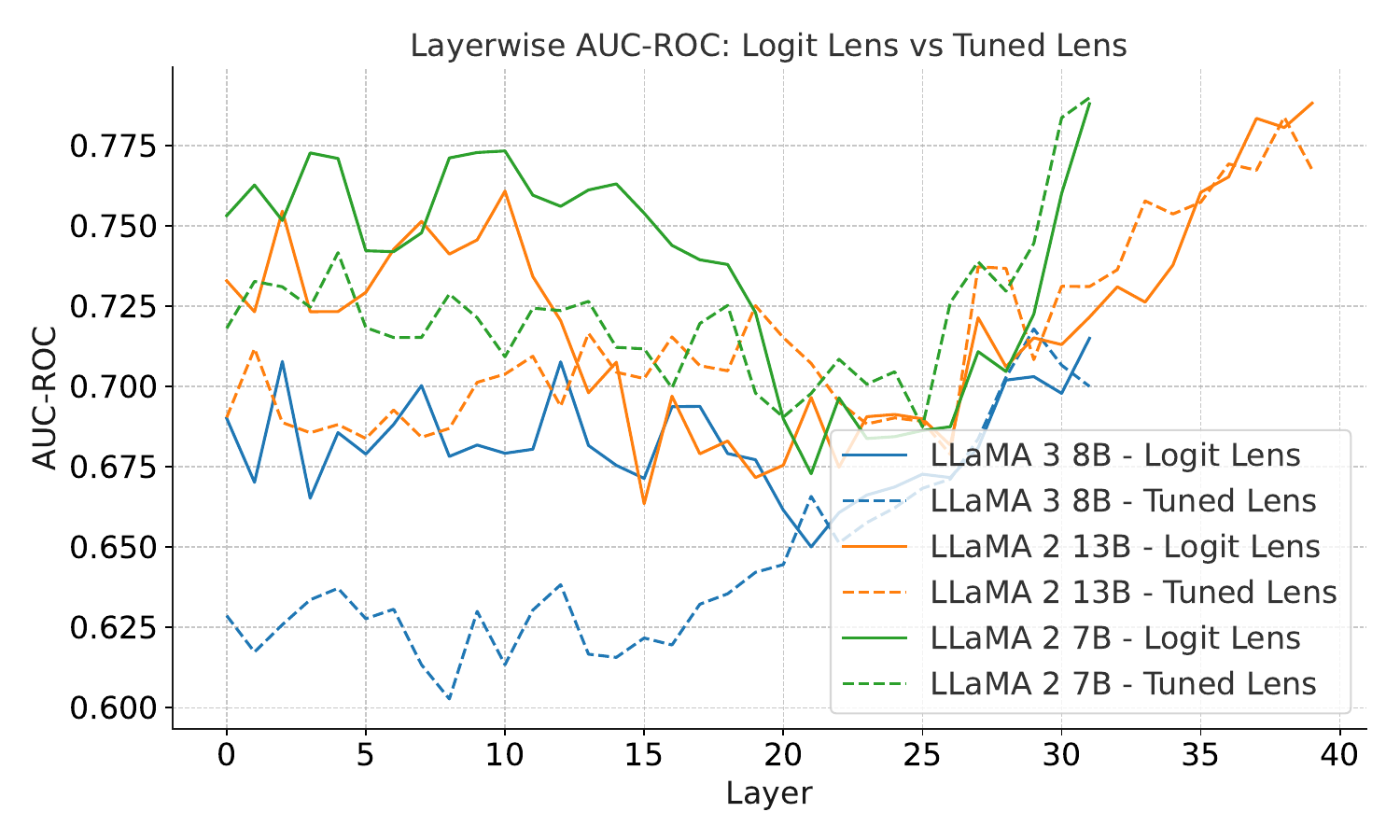}
\caption{Layerwise AUC-ROC for Logit Lens (solid lines) and Tuned Lens (dashed lines).}
\label{fig:layerwise_auc_logit_vs_tuned}
\end{figure}

\Cref{fig:layerwise_auc_logit_vs_tuned} shows the layerwise AUC-ROC performance of the Logit Lens and Tuned Lens on LLaMA 3 8B, LLaMA 2 13B, and LLaMA 2 7B. We do not observe a consistent trend in the relative performance that would clearly indicate which method is superior. However, both the Logit Lens and Tuned Lens exhibit predictive capability from very early layers.

\paragraph{Feature Importance}

\begin{figure*}[ht]
\centering

\includegraphics[width=\linewidth]{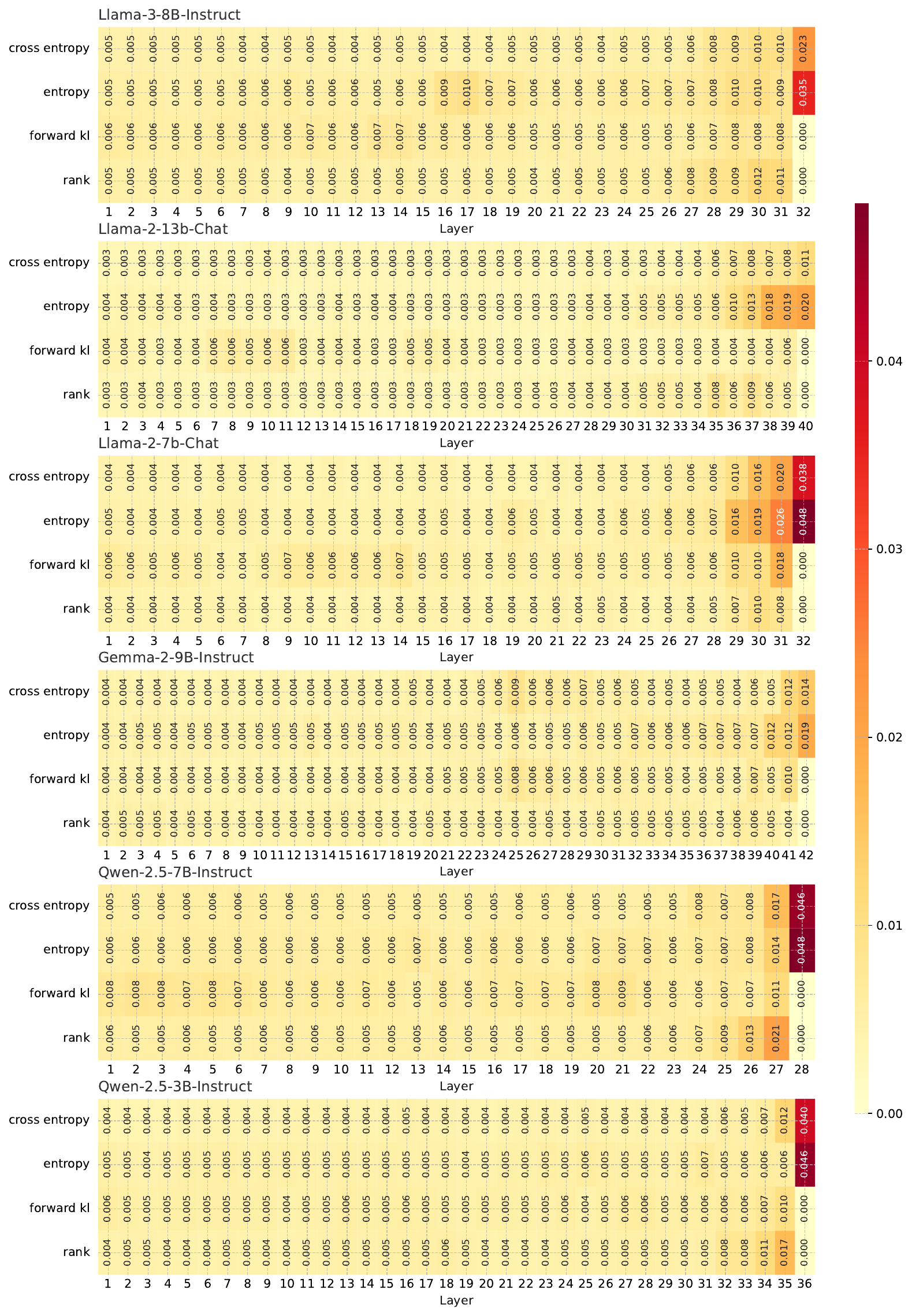}

\caption{Impurity-based random forest feature importance scores for Logit Lens features from each layer across six models on TriviaQA. Top-$p$ features contribute minimally and are therefore excluded.}
\label{fig:feature_importance_lens}
\end{figure*}

We present the Logit Lens feature importance heatmaps for all six models in \cref{fig:feature_importance_lens} based on the trained classifiers. Top-$k$ features are omitted due to their consistently low importance across all layers and models.

\paragraph{ROC Curve: External VS. Internal}

\begin{figure*}[ht]
\centering

% First row
\begin{minipage}[t]{0.49\linewidth}
    \centering
    \includegraphics[width=\linewidth]{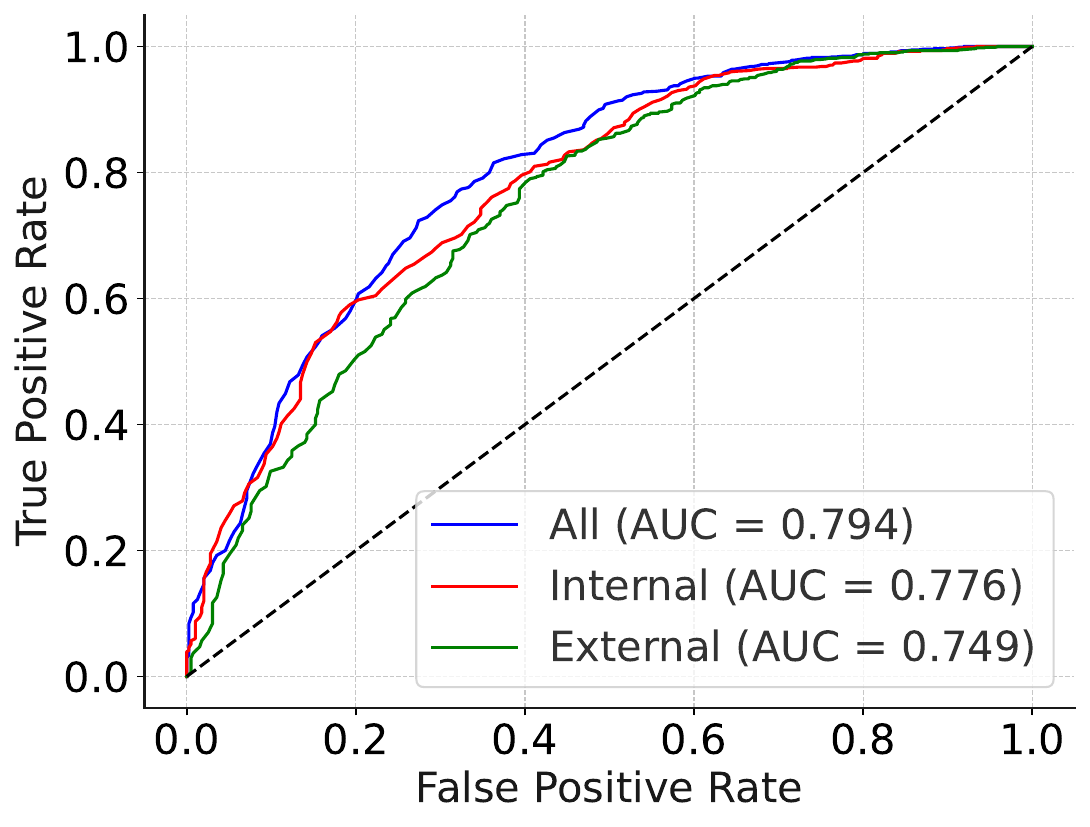}
    \subcaption{LLaMA 3 8B}
    \label{fig:roc_curve_llama3_8b}
\end{minipage}
\hfill
\begin{minipage}[t]{0.49\linewidth}
    \centering
    \includegraphics[width=\linewidth]{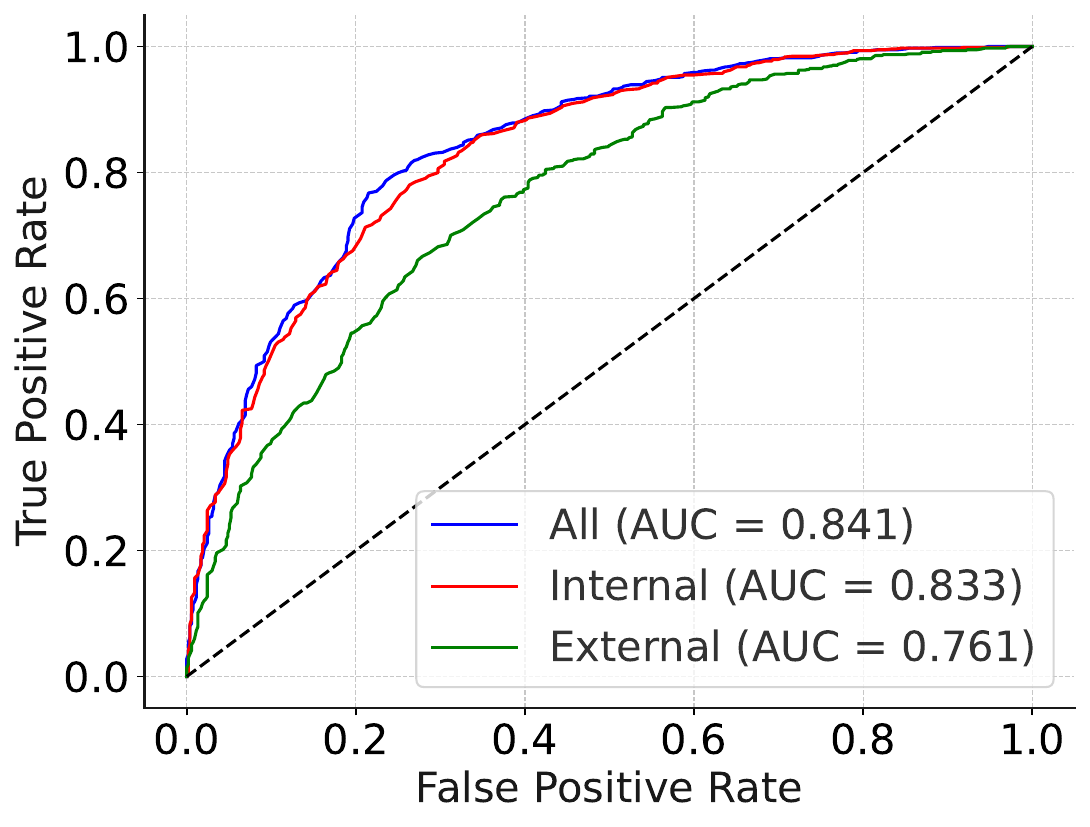}
    \subcaption{LLaMA 2 13B}
    \label{fig:roc_curve_llama2_13b}
\end{minipage}
\vspace{1em}

\begin{minipage}[t]{0.49\linewidth}
    \centering
    \includegraphics[width=\linewidth]{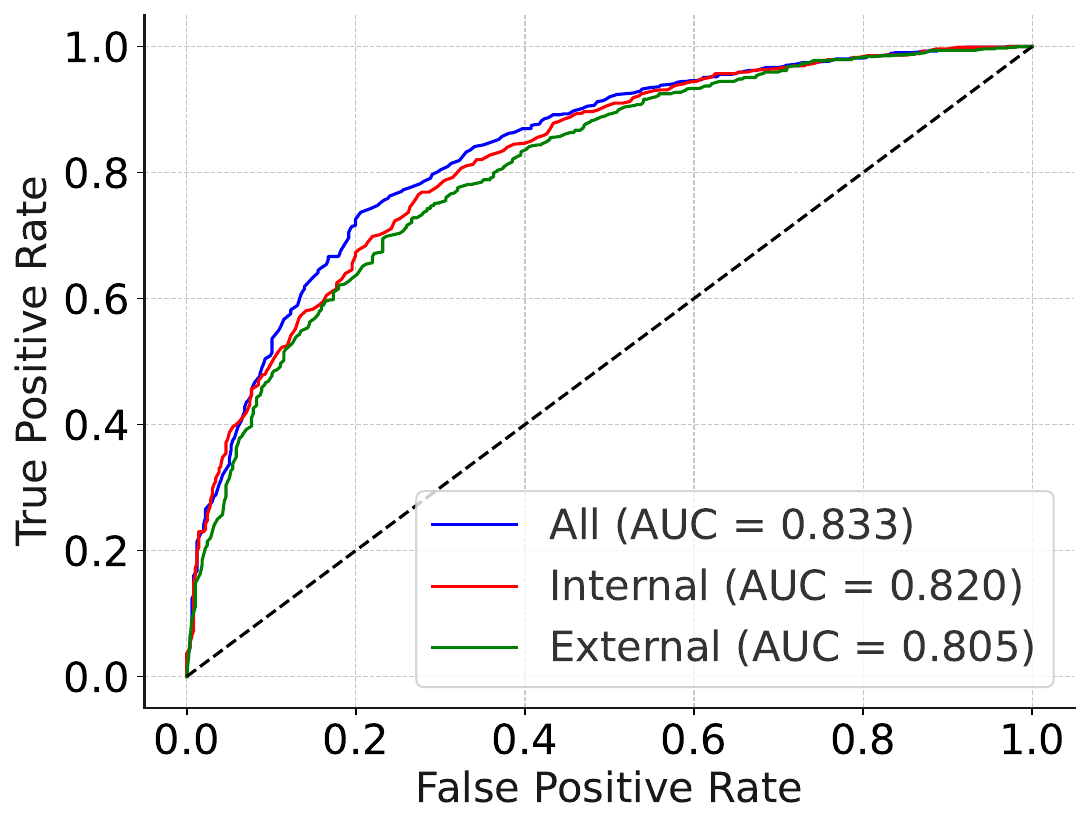}
    \subcaption{LLaMA 2 7B}
    \label{fig:roc_curve_llama2_7b}
\end{minipage}
\hfill
% Second row
\begin{minipage}[t]{0.49\linewidth}
    \centering
    \includegraphics[width=\linewidth]{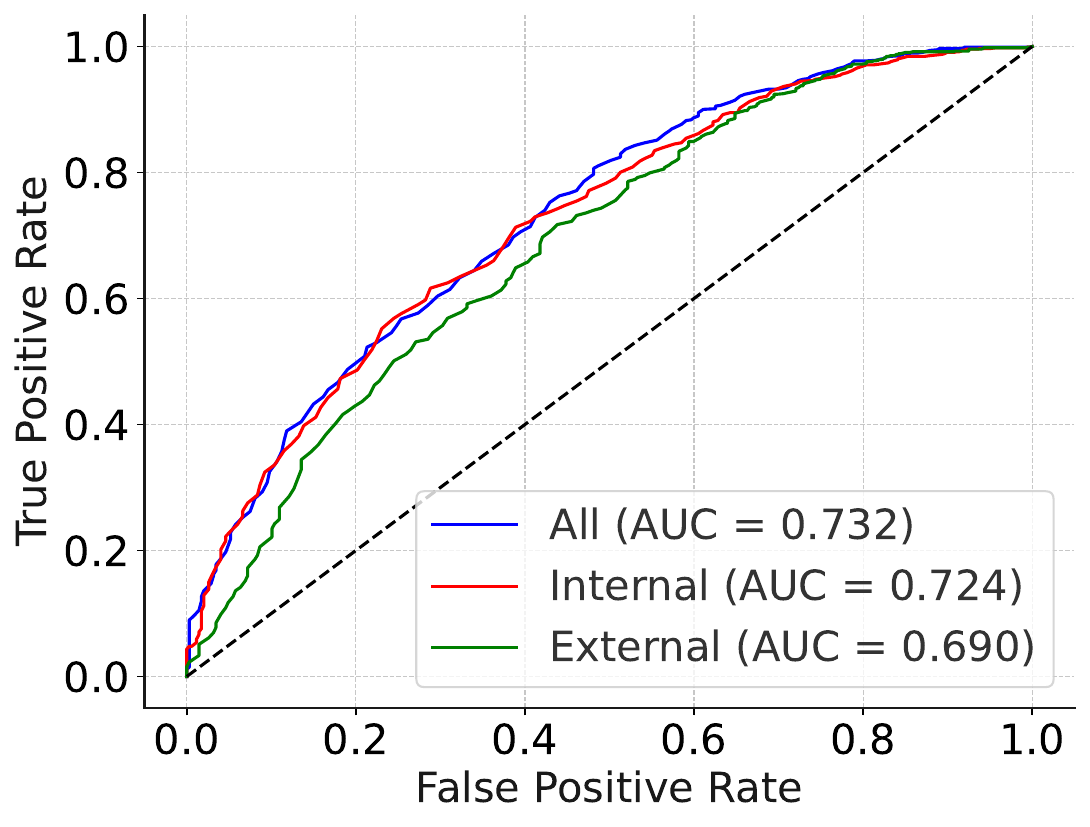}
    \subcaption{Gemma 2 9B}
    \label{fig:roc_curve_gemma2_9b}
\end{minipage}

\vspace{1em}
\begin{minipage}[t]{0.49\linewidth}
    \centering
    \includegraphics[width=\linewidth]{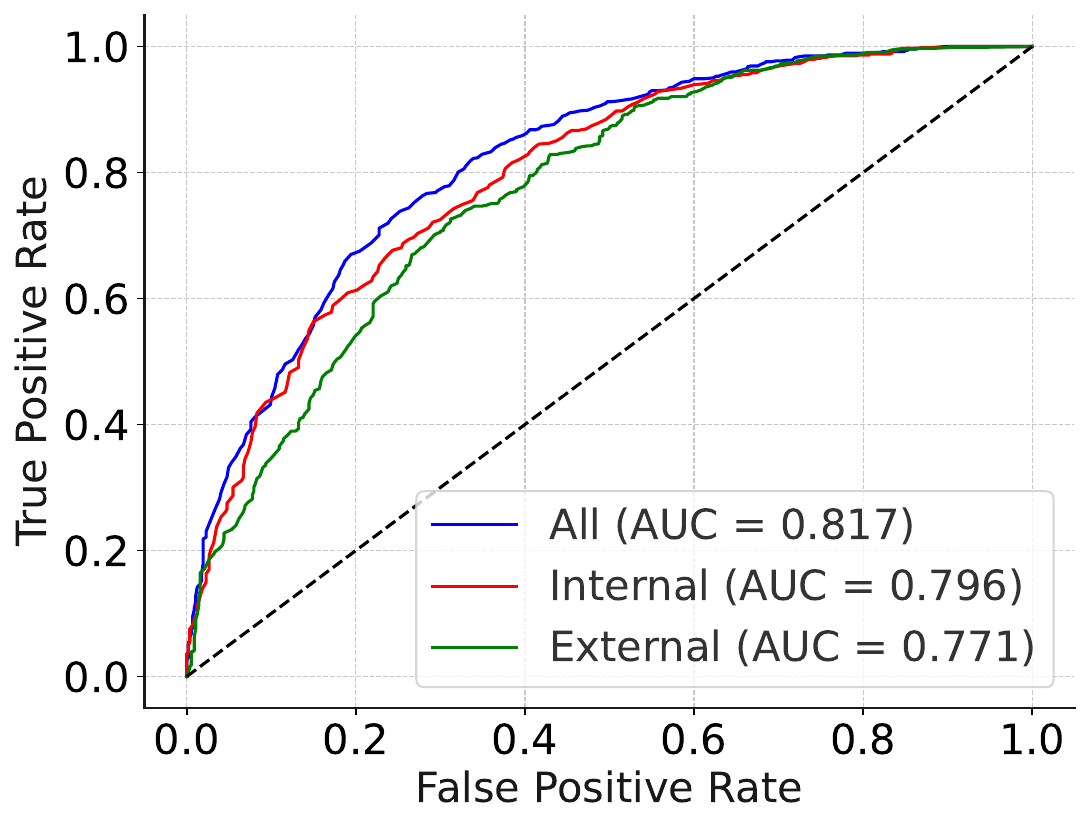}
    \subcaption{Qwen 2.5 7B}
    \label{fig:roc_curve_qwen25_7b}
\end{minipage}
\hfill
\begin{minipage}[t]{0.49\linewidth}
    \centering
    \includegraphics[width=\linewidth]{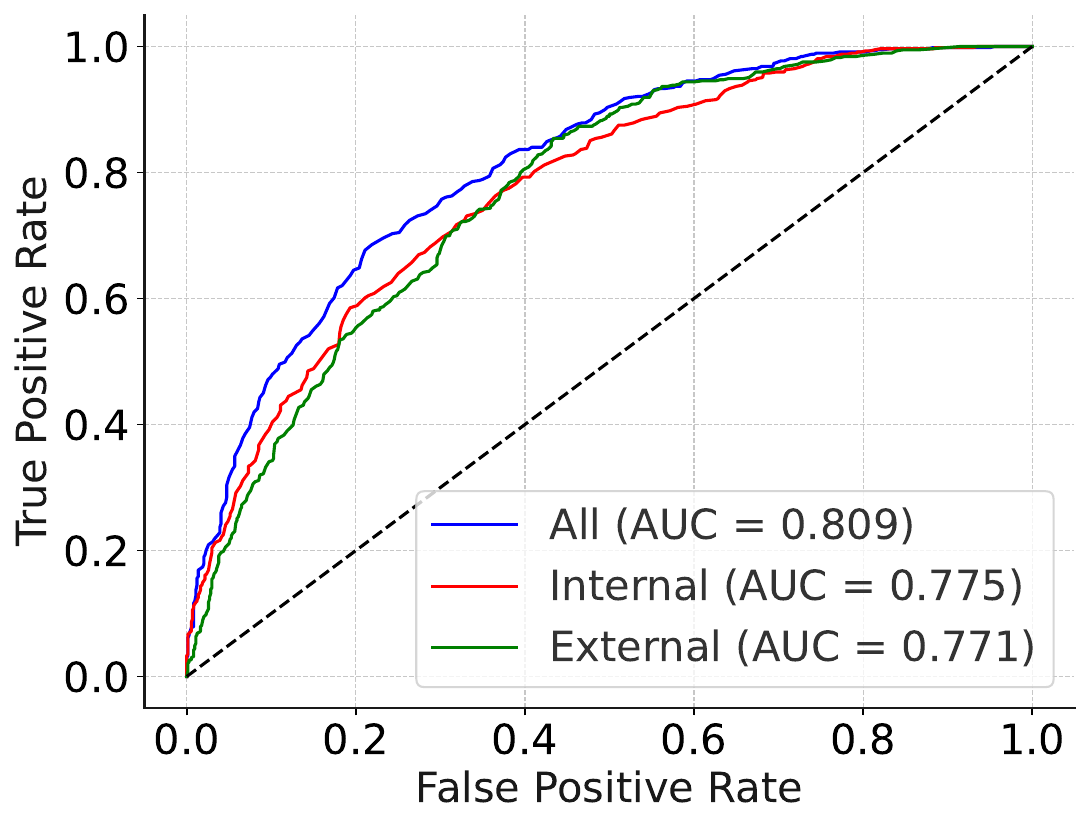}
    \subcaption{Qwen 2.5 3B}
    \label{fig:roc_curve_qwen25_3b}
\end{minipage}

\vspace{1em}

\caption{ROC curves comparing classifiers trained on last-layer features (external) versus internal-layer features, across six models on TriviaQA.}
\label{fig:roc_curves_all_models}
\end{figure*}

\Cref{fig:roc_curves_all_models} shows the ROC curve comparisons between classifiers trained solely on last-layer features (“external”) and those trained on internal-layer features (“internal”) across all six models. The internal classifier achieves performance comparable to using features from all layers, while consistently outperforming the external classifier across all tested models.

\begin{figure*}[ht]
\centering

% First row
\begin{minipage}[t]{0.49\linewidth}
    \centering
    \includegraphics[width=\linewidth]{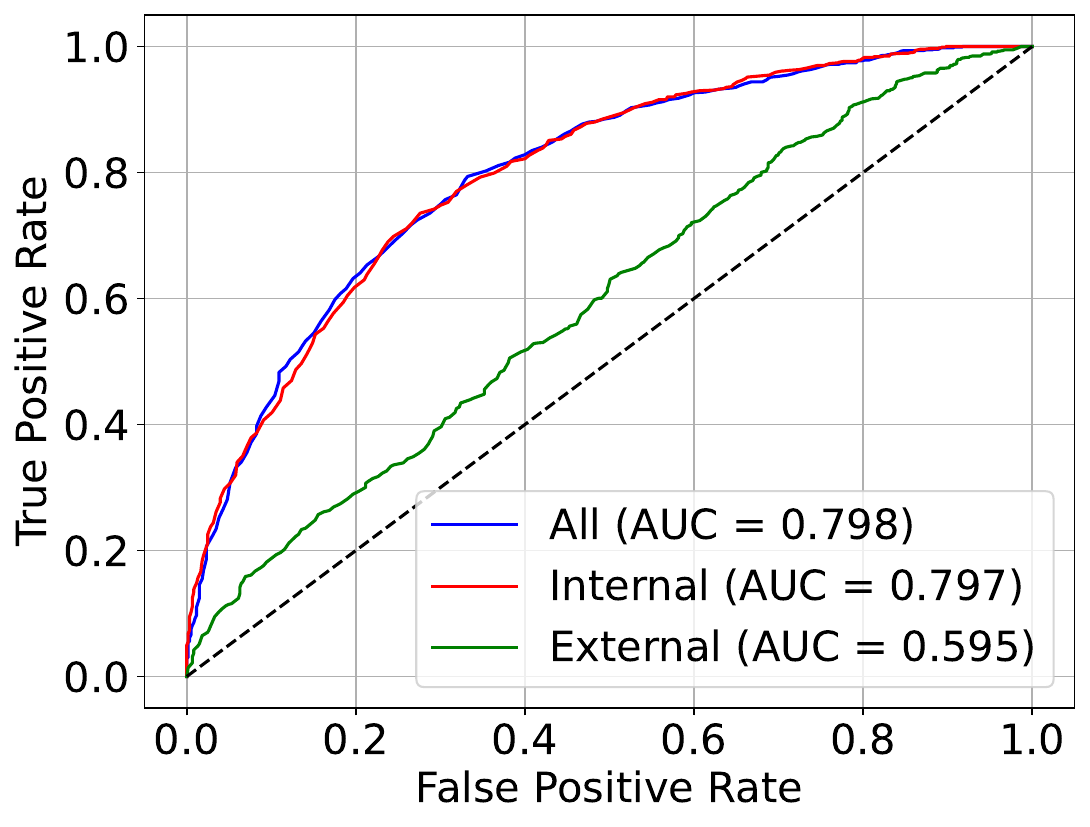}
    \subcaption{LLaMA 3 8B}
    \label{fig:roc_curve_llama3_8b_mmlu}
\end{minipage}
\hfill
\begin{minipage}[t]{0.49\linewidth}
    \centering
    \includegraphics[width=\linewidth]{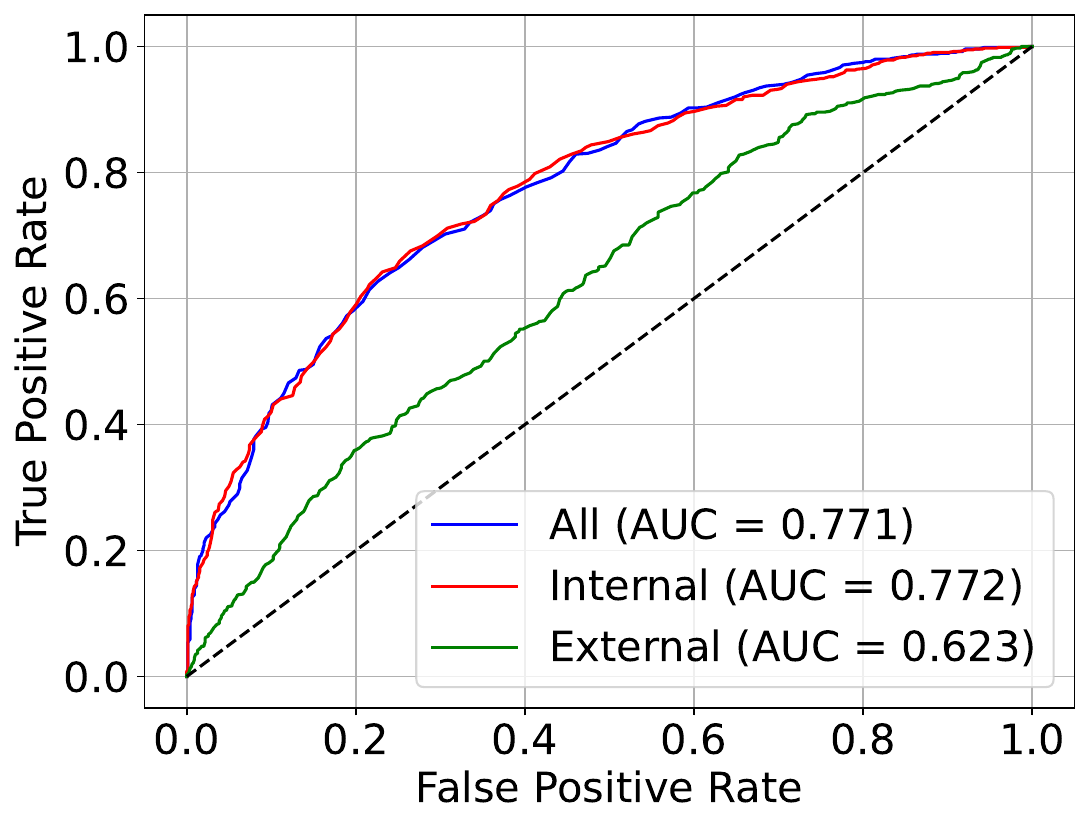}
    \subcaption{LLaMA 2 13B}
    \label{fig:roc_curve_llama2_13b_mmlu}
\end{minipage}
\vspace{1em}

\begin{minipage}[t]{0.49\linewidth}
    \centering
    \includegraphics[width=\linewidth]{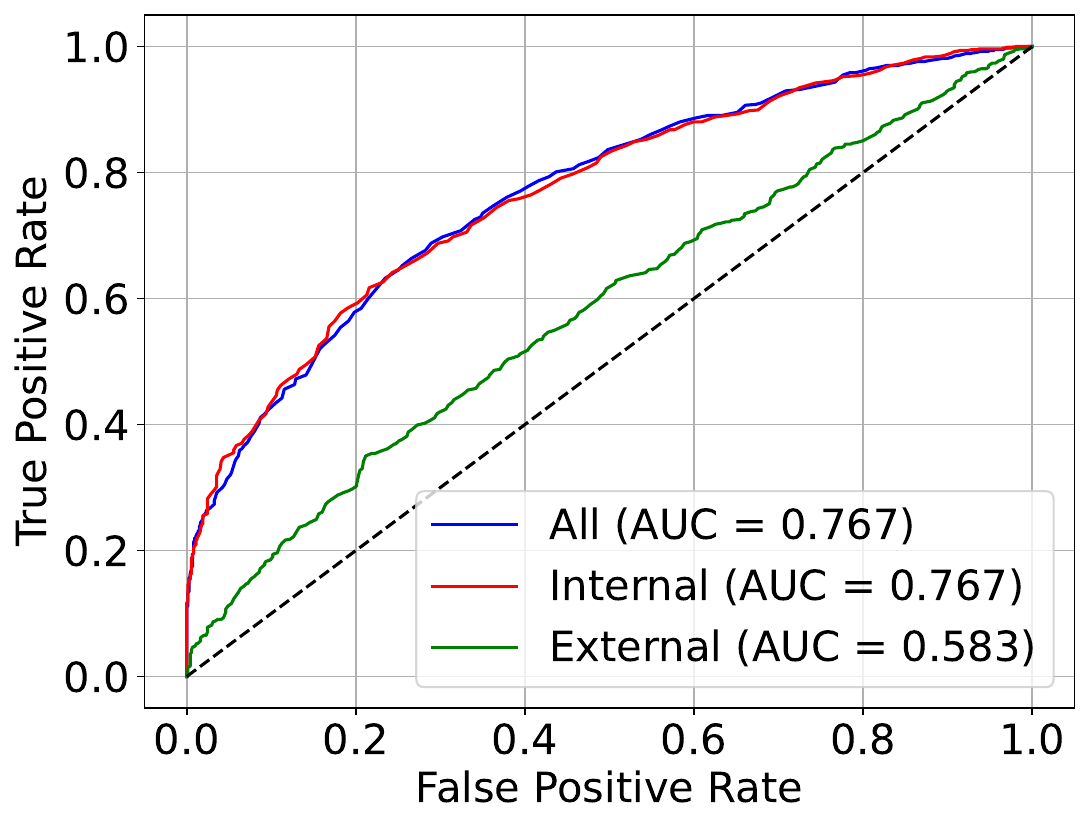}
    \subcaption{LLaMA 2 7B}
    \label{fig:roc_curve_llama2_7b_mmlu}
\end{minipage}
\hfill
% Second row
\begin{minipage}[t]{0.49\linewidth}
    \centering
    \includegraphics[width=\linewidth]{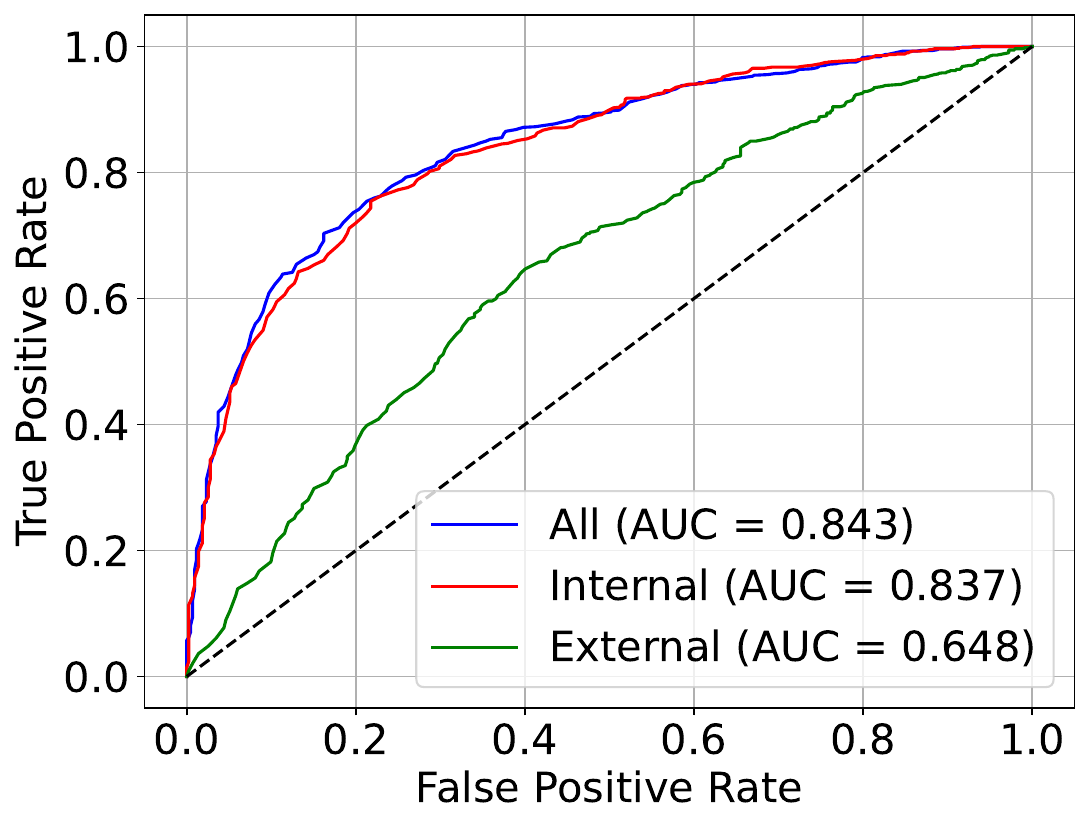}
    \subcaption{Gemma 2 9B}
    \label{fig:roc_curve_gemma2_9b_mmlu}
\end{minipage}

\vspace{1em}
\begin{minipage}[t]{0.49\linewidth}
    \centering
    \includegraphics[width=\linewidth]{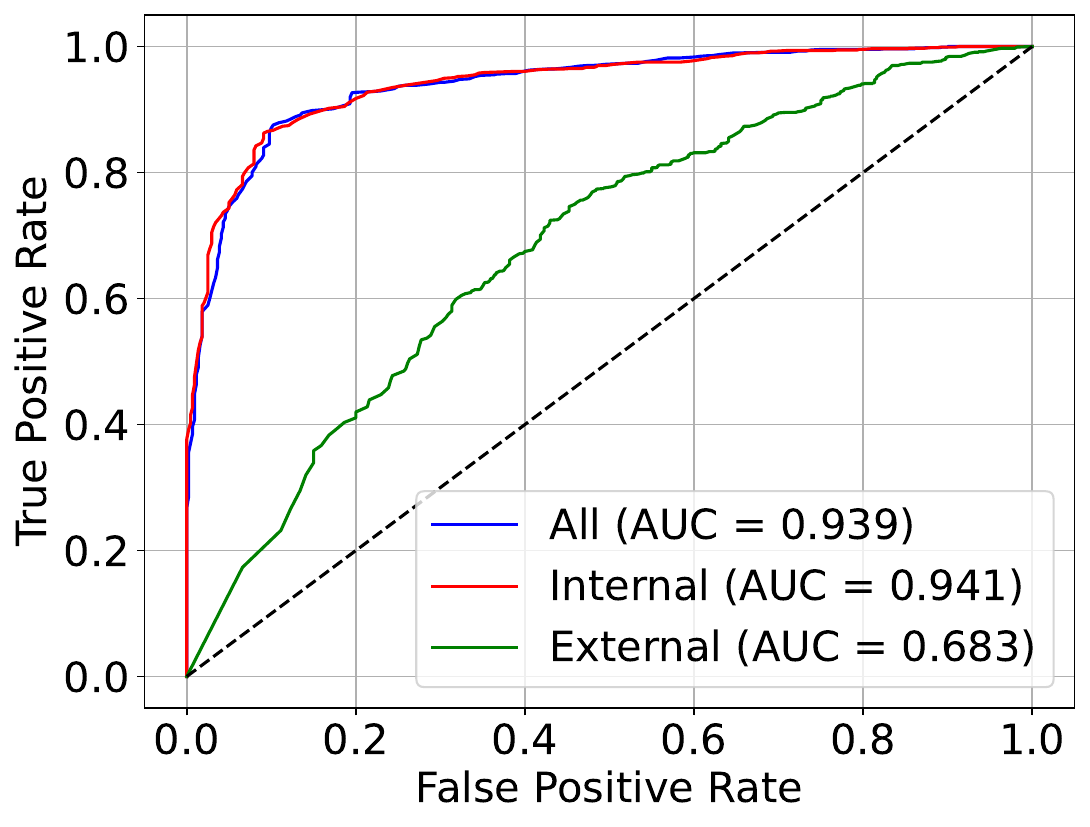}
    \subcaption{Qwen 2.5 7B}
    \label{fig:roc_curve_qwen25_7b_mmlu}
\end{minipage}
\hfill
\begin{minipage}[t]{0.49\linewidth}
    \centering
    \includegraphics[width=\linewidth]{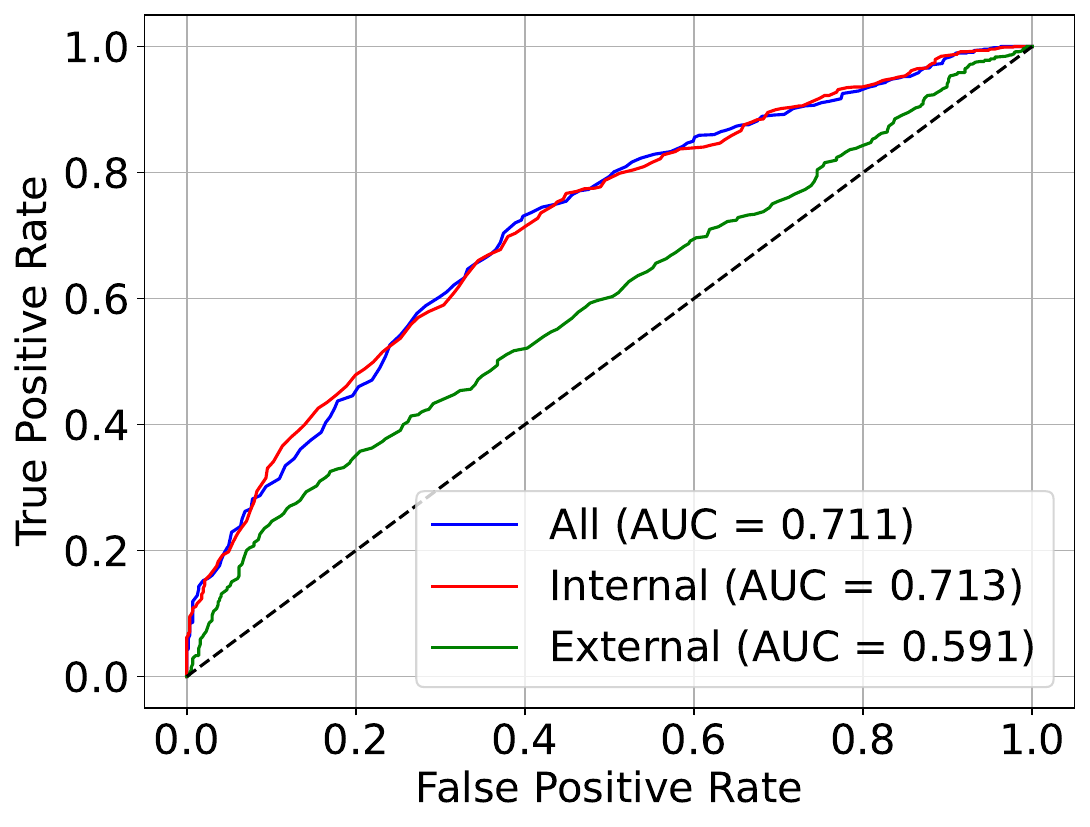}
    \subcaption{Qwen 2.5 3B}
    \label{fig:roc_curve_qwen25_3b_mmlu}
\end{minipage}

\vspace{1em}

\caption{ROC curves comparing classifiers trained on last-layer features (external) versus internal-layer features, across six models on MMLU.}
\label{fig:roc_curves_all_models_mmlu}
\end{figure*}

\subsection{Hidden States}
\label{apn:hidden_states}

\begin{figure*}[ht!]
\centering
% First row
% \includegraphics[width=\linewidth]{figs/layerwise_classifier_metrics_norm_area_under_ROC_curve.png}
\begin{subfigure}{0.49\linewidth}
    \includegraphics[width=\linewidth]{figs/updated_hidden_states/layerwise_auc_norm.png}
    \caption{TriviaQA, with z-score normalization}
  \end{subfigure}
  \hfill
  % ------------ right panel -----------
  \begin{subfigure}{0.49\linewidth}
    \includegraphics[width=\linewidth]{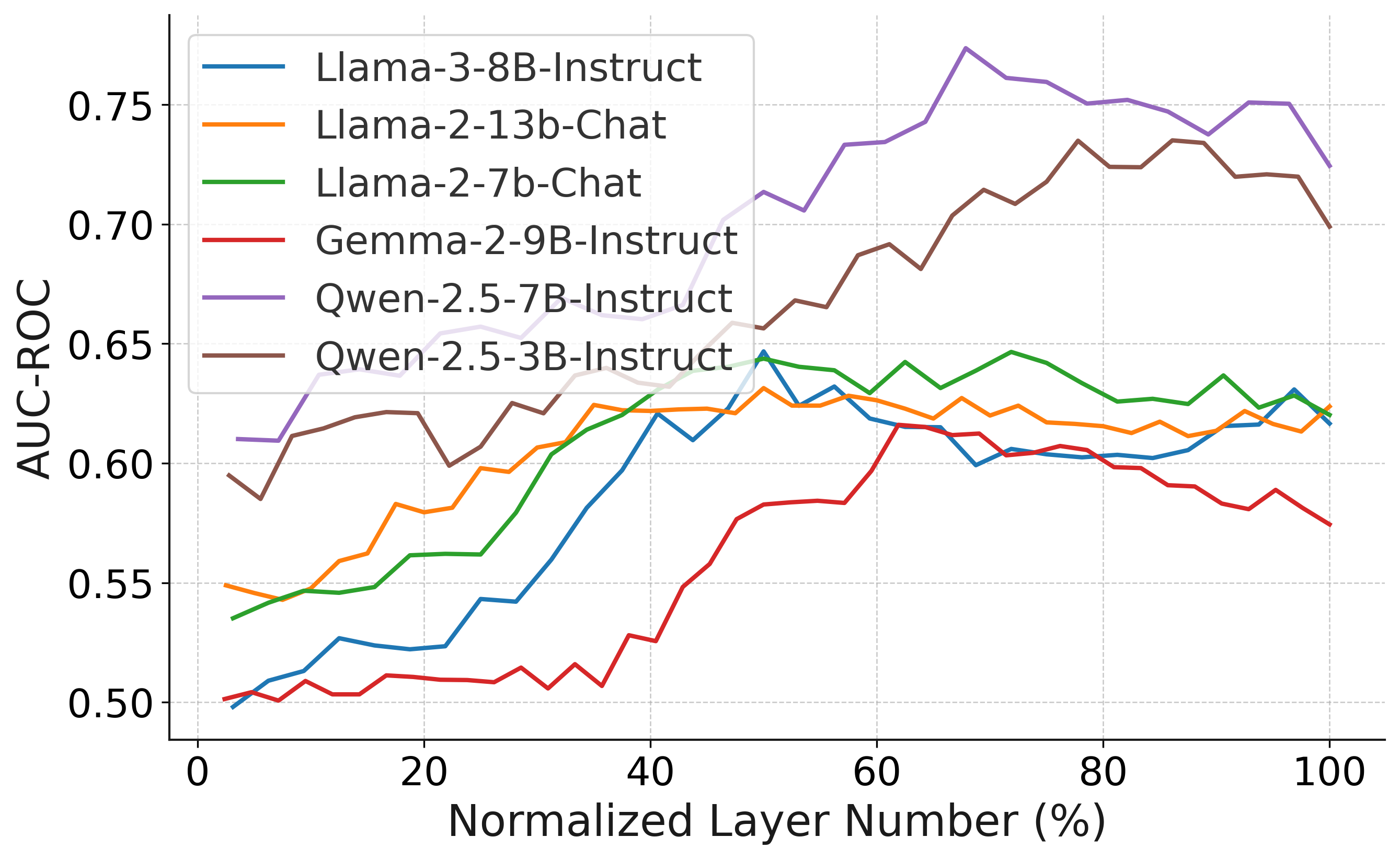}
    \caption{TriviaQA, without normalization}           % produces “(b)”
  \end{subfigure}
\begin{subfigure}{0.49\linewidth}
    \includegraphics[width=\linewidth]{figs/mmlu_hs/mmlu_layerwise_auc_norm.png}
    \caption{MMLU, with z-score normalization}
  \end{subfigure}
  \hfill
  % ------------ right panel -----------
  \begin{subfigure}{0.49\linewidth}
    \includegraphics[width=\linewidth]{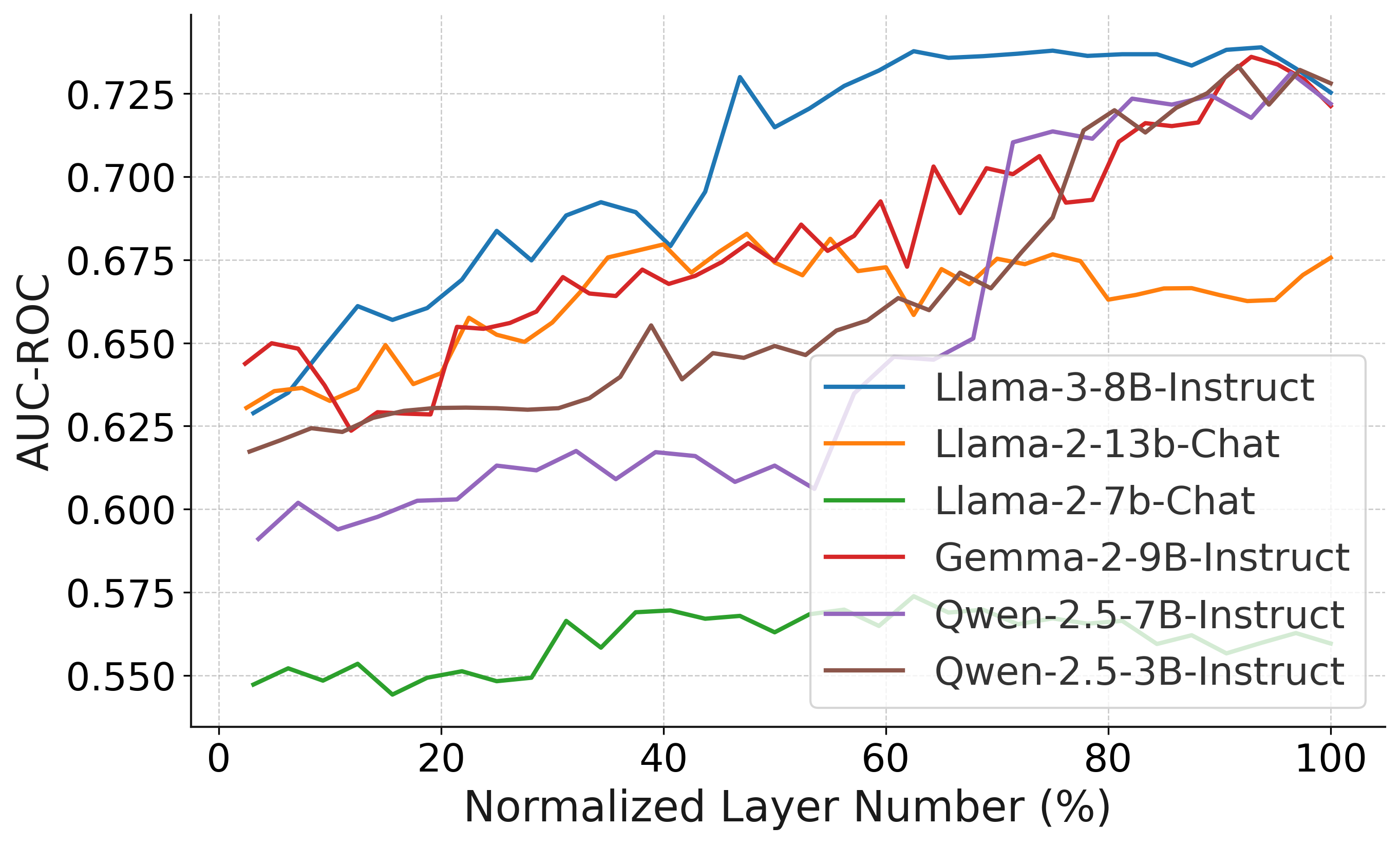}
    \caption{MMLU, without normalization}           % produces “(b)”
  \end{subfigure}
\caption{Area under ROC curve for random forest classifiers trained on z-score normalized hidden states of each layer. Performance increases with layer depth, suggesting that later layers refine and consolidate decision-relevant signals. Normalization has virtually no effect on performance.}
\label{fig:layerwise_hidden_states_auc_roc_norm_vs_no_norm}
\end{figure*}

\paragraph{Effect of Z-Score Normalization}
We obtain means and variances per hidden state dimension for z-score normalization on auxiliary subsets of TriviaQA and MMLU to ensure the distribution remains similar to the train and test examples while avoiding leakage of information across examples. To evaluate the impact of z-score normalization on classifier performance, we compare models trained on hidden state features with and without normalization and find virtually identical performance (see \cref{fig:layerwise_hidden_states_auc_roc_norm_vs_no_norm}). This suggests that the classifier is relatively insensitive to the absolute magnitude of hidden state feature values across different dimensions and examples.

\section{Additional Results for RQ2}

% \subsection{PKS and ECS Score Distributions Across Examples}
% \label{appn:res_analysis_rq2}

% \Cref{fig:pks_ecs_histogram} and \cref{fig:pks_ecs_histogram_gemma2_9b} show the histogram distributions of PKS and ECS scores for LLaMA 3 8B and Gemma 2 9B, respectively. For LLaMA 3 8B, PKS remains relatively stable across different context types, while ECS clearly reflects context shifts when transitioning from correct to incorrect or irrelevant contexts. The other models, including LLaMA 2 7B, Qwen 2.5 7B, and Qwen 2.5 3B, exhibit distributions similar to LLaMA 3 8B. However, Gemma 2 9B stands out due to its poor performance in \cref{tab:main_results_rq2}. These histograms help explain why: while its PKS distribution remains stable, its ECS distribution shows a pronounced spike in the middle, reducing ECS's ability to discriminate between different context types.

\subsection{Non-Applicability of ECS on Gemma 2 9B}
\label{appn:res_analysis_rq2}

% \Cref{fig:pks_ecs_histogram} and \cref{fig:pks_ecs_histogram_gemma2_9b} show the histogram distributions of PKS and ECS scores for LLaMA 3 8B and Gemma 2 9B, respectively. For LLaMA 3 8B, PKS remains relatively stable across different context types, while ECS clearly reflects context shifts when transitioning from correct to incorrect or irrelevant contexts. The other models, including LLaMA 2 7B, Qwen 2.5 7B, and Qwen 2.5 3B, exhibit distributions similar to LLaMA 3 8B. 
Gemma 2 9B employs sliding-window attention on every even-numbered layer, which makes it infeasible to implement ECS in a way that correctly tracks the top-k attention scores. Consequently, we exclude Gemma 2 9B from our analysis for RQ2.

% \begin{figure*}[t]
%   \centering
%   % ------------ left panel ------------
%   \begin{subfigure}{\linewidth}
%     \includegraphics[width=\linewidth]{figs/ecs_pks_distribution/llama3_8b_pks_distribution.pdf}
%     \caption{PKS}
%   \end{subfigure}
%   % \hfill
%   % ------------ right panel -----------
%   \begin{subfigure}{\linewidth}
%     \includegraphics[width=\linewidth]{figs/ecs_pks_distribution/llama3_8b_ecs_distribution.pdf}
%     \caption{ECS}           % produces “(b)”
%   \end{subfigure}
%   \caption{Histogram distributions of PKS and ECS scores for LLaMA 3 8B on all TriviaQA examples.}
% \label{fig:pks_ecs_histogram}
% \end{figure*}

% \begin{figure*}[t]
%   \centering
%   % ------------ left panel ------------
%   % \begin{subfigure}{\linewidth}
%   %   \includegraphics[width=\linewidth]{figs/ecs_pks_distribution/gemma2_9b_pks_distribution.pdf}
%   %   \caption{PKS}
%   % \end{subfigure}
%   % \hfill
%   % ------------ right panel -----------
%   % \begin{subfigure}{\linewidth}
%     \includegraphics[width=\linewidth]{figs/ecs_pks_distribution/gemma2_9b_ecs_distribution.pdf}
%     % \caption{ECS}           % produces “(b)”
%   % \end{subfigure}
%   % \caption{Histogram distributions of PKS and ECS scores for Gemma 2 9B on all TriviaQA examples.}
%   \caption{Distribution of ECS scores for Gemma 2 9B across examples from TriviaQA. ECS scores are averaged across layers and tokens. }
% \label{fig:pks_ecs_histogram_gemma2_9b}
% \end{figure*}

\subsection{Examples with PKS and ECS Scores}
\label{appn:examples_pks_ecs}

\begin{figure*}
    \centering
    \includegraphics[width=\linewidth]{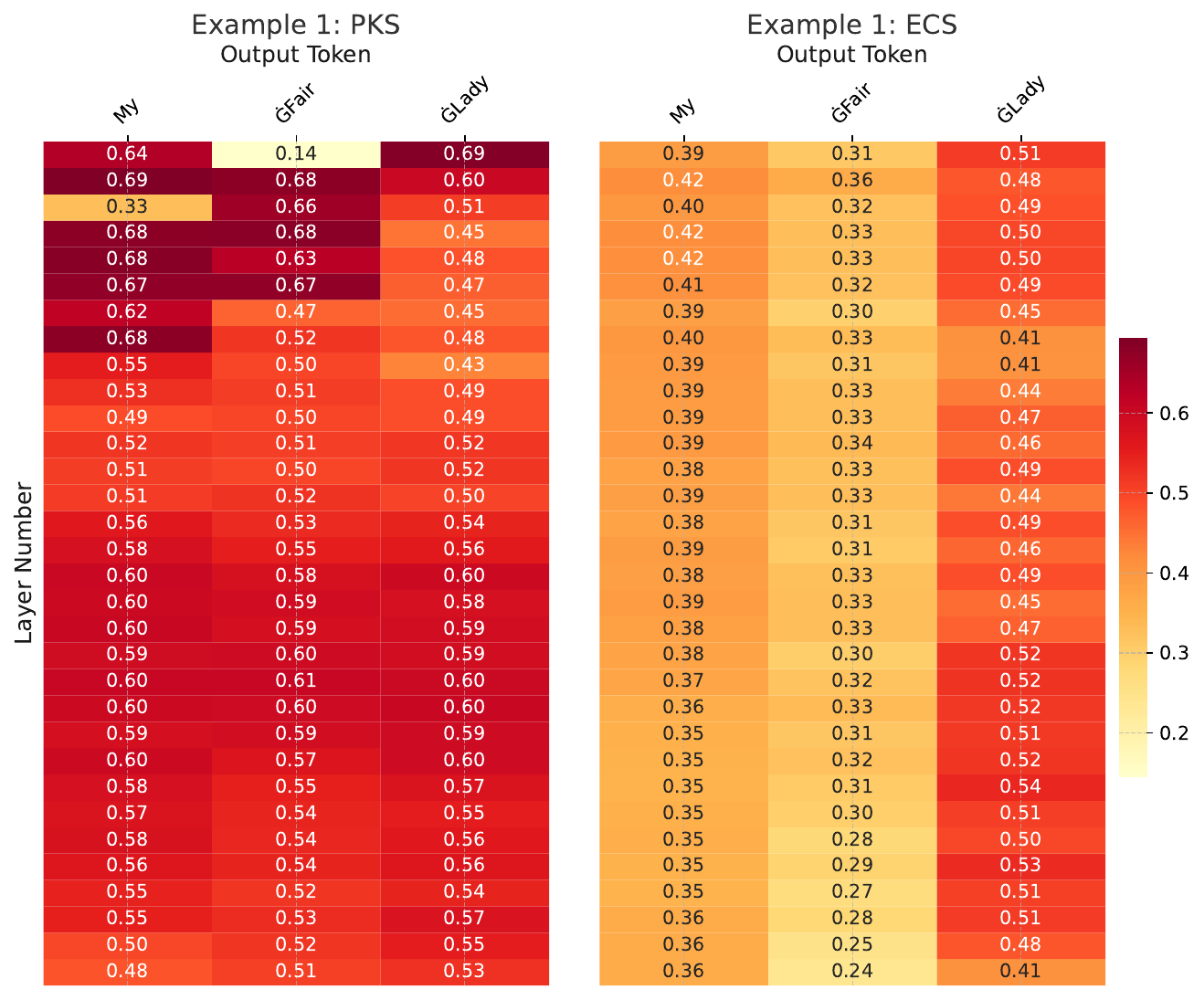}
    \caption{Heatmap example 1 from TriviaQA of PKS and ECS scores for LLaMA-3-8B, illustrating layer-wise token importance. Question: "\textit{Which musical featured the song The Street Where You Live?}" Context: "\textit{"On the Street Where You Live" is a song with music by Frederick Loewe and lyrics by Alan Jay Lerner from the 1956 Broadway musical My Fair Lady. It is sung in the musical by the character Freddy Eynsford-Hill, originally portrayed by John Michael King. In the 1964 film version, the song was performed by Bill Shirley, dubbing for Jeremy Brett. The most popular single was recorded by Vic Damone in 1956, reaching \#4 on the Billboard charts and \#6 on Cash Box magazine's chart, and it was a \#1 hit in the UK in 1958.In 1955, Damone had one song on the charts, "Por Favor," which peaked at \#73, but he starred in Hit the Deck and Kismet. In 1956, he moved to Columbia Records, achieving success with hits like "On the Street Where You Live" from My Fair Lady and "An Affair to Remember." His albums on Columbia included That Towering Feeling, Angela Mia, Closer Than a Kiss, This Game of Love, On the Swingin' Side, and Young and Lively. Lyrics describe the narrator's thrill on the street where a loved one lives, highlighting the emotional impact of such proximity. The content is administered by SME and used here for educational purposes under fair use. If concerns arise about unauthorized use, contact the poster. This adheres to the Copyright Act's fair use principles for criticism, comment, news reporting, teaching, scholarship, and research, emphasizing non-profit, educational intentions.}"}
    \label{fig:ag_redeep_1}
\end{figure*}

\begin{figure*}
    \centering
    \includegraphics[width=\linewidth]{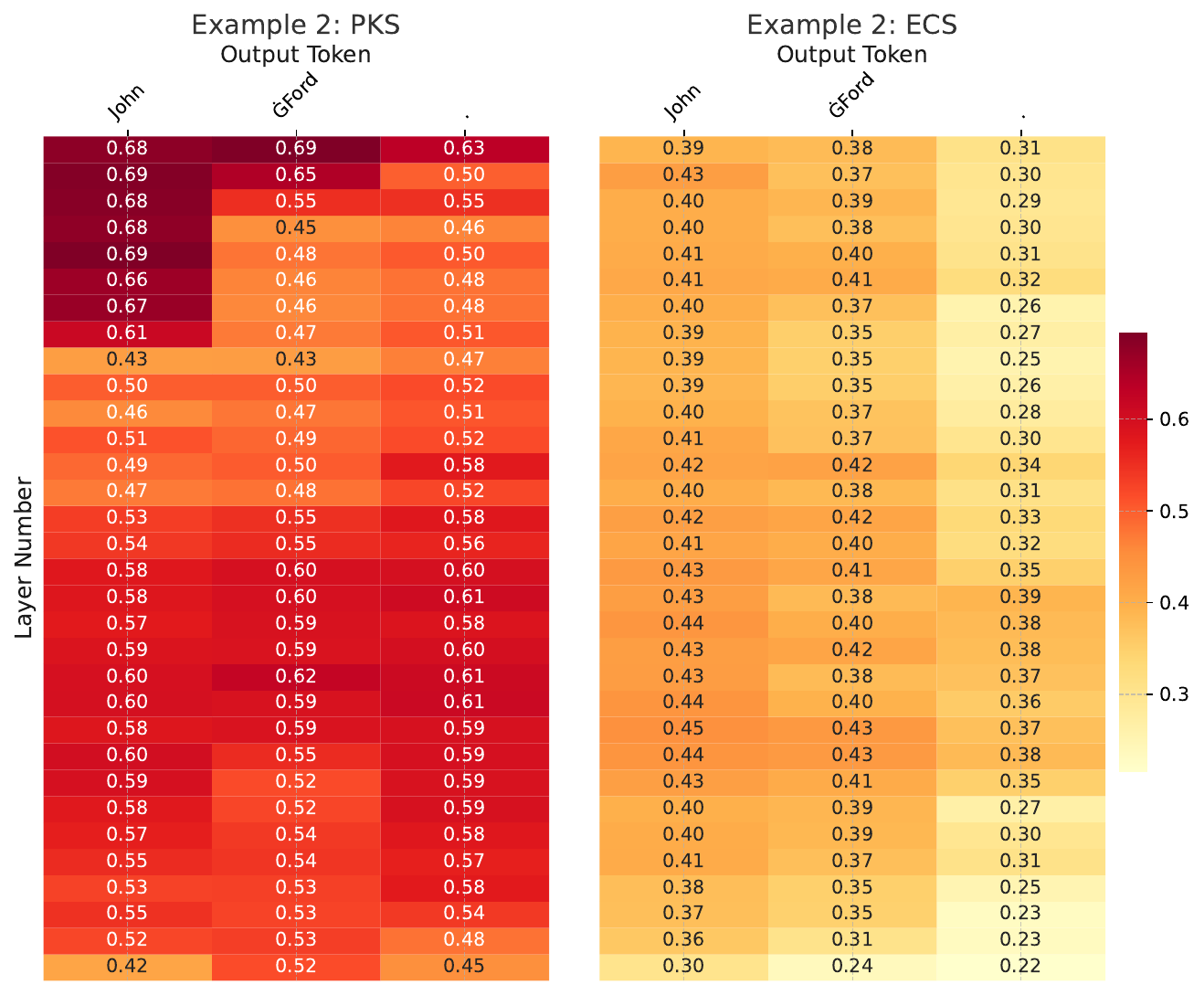}
    \caption{Heatmap example 2 from TriviaQA of PKS and ECS scores for LLaMA-3-8B, illustrating layer-wise token importance. Question: "\textit{Who directed the classic 30s western Stagecoach?}" Context: "\textit{Stagecoach, directed by John Ford (1895-1973), is a quintessential Western notable for its complex characters and Monument Valley setting. John Wayne, portraying "The Ringo Kid," catapulted to stardom in this film. The narrative follows a diverse group of travelers on a stagecoach from Tonto to Lordsburg, facing dangers from an Apache uprising led by Geronimo. Central characters include a falsely accused outlaw Ringo Kid, played by Wayne—seeking to avenge his family's murder—and a prostitute named Dallas, portrayed by Claire Trevor. Other passengers include a drunken doctor (Thomas Mitchell, whose performance won an Academy Award), a gentleman gambler, an embezzling banker, a pregnant army wife, and others, each contributing to a microcosm of society. Ford's direction, coupled with Dudley Nichols and Ben Hecht's script, ensures tight plotting and memorable character arcs, many based on Ernest Haycox's short story "Stage to Lordsburg" and Guy de Maupassant’s "Boule de Suif." Ford's handling allowed for minimal screen time yet deep character development. The film features an intense climax with a chase/fight scene that includes Yakima Canutt's pioneering stunts, echoing Spielberg's Raiders of the Lost Ark years later. Stagecoach is celebrated for blending mythic Western landscapes with a poignant social allegory, reflecting on issues like prejudice and redemption. While technical aspects, like some cinematography, show their age, the film remains a paragon from the era, emblematic of Ford's work and establishing recurring Western motifs. It earned seven Oscar nominations, securing wins for Best Supporting Actor and Best Score. Stagecoach's impact persists, marking a pivotal moment in film history and solidifying John Ford's legacy as a master director, influencing numerous directors and films in subsequent years. Whether or not it is perceived as a perfect film, it stands as a significant cultural artifact and vital viewing for any student of cinema.}"}
    \label{fig:ag_redeep_2}
\end{figure*}

\begin{figure*}
    \centering
    \includegraphics[width=\linewidth]{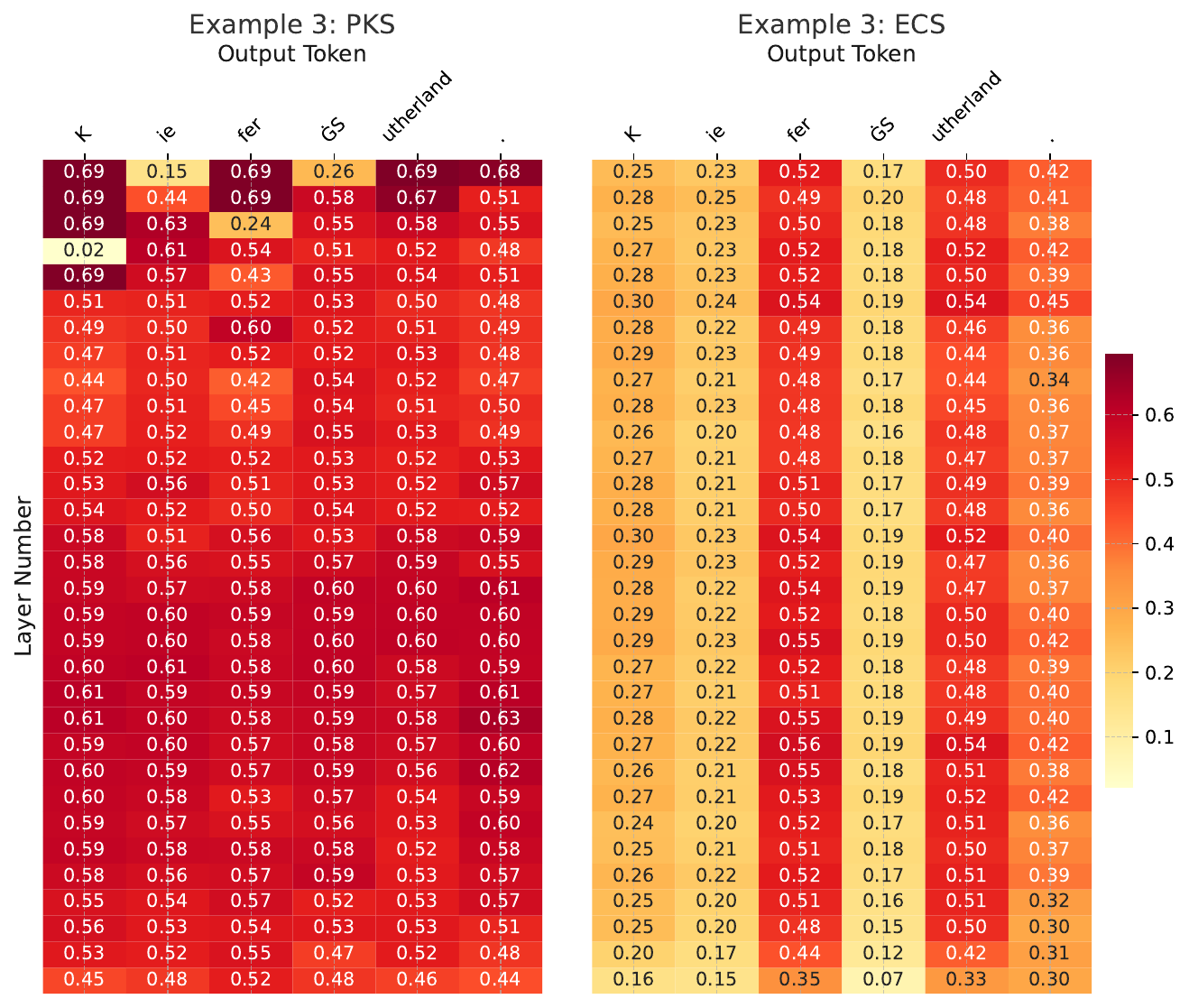}
    \caption{Heatmap example 3 from TriviaQA of PKS and ECS scores for LLaMA-3-8B, illustrating layer-wise token importance. Question: "\textit{Who was born first, Kiefer Sutherland or Christian Slater?}" Context: "\textit{Young Guns II (1990) follows Billy the Kid and his gang as wanted outlaws. The story unfolds with Pat Garrett, a former partner of Billy’s, being paid to kill him by cattle baron John Chisum. The movie, directed by Christopher Cain, explores themes of betrayal and redemption as Pat Garrett, who has plans to go respectable, is conflicted about turning against his former friend. Garrett seizes the opportunity to become Sheriff offered by the Governor, who believes in hiring a thief to catch one. This decision to capture Billy opens up great western adventure, with Pat grappling between loyalty and duty. The nuanced performance of William Petersen as Garrett contrasts well with Emilio Estevez reprising his role as the charismatic Billy the Kid. Lou Diamond Phillips elevates his role as Chavez, delivering a performance that is more spiritual and wise than in the first Young Guns film. The talented cast also includes Kiefer Sutherland, Christian Slater, Balthazar Getty, and Alan Ruck. Though depicted as close friends, the real-life association between Pat Garrett and Billy the Kid was less intimate; their familiarity stemmed from mutual patronage of a saloon. Despite this inaccuracy, the film presents an engaging exploration of the dynamics between Garrett and Billy. Young Guns II is a remarkable sequel to 1988's Young Guns, offering a compelling mix of action and moral questions. The film's strong character portrayals, particularly by Estevez and Phillips, enhance its rich narrative of western adventure and friendship against a backdrop of historical myth. It's an exciting film experience that shouldn’t be missed.}"}
    \label{fig:ag_redeep_3}
\end{figure*}

We illustrate how PKS and ECS scores evolve as the model generates output tokens across layers, using several examples shown in Figures~\ref{fig:ag_redeep_1},~\ref{fig:ag_redeep_2}, and~\ref{fig:ag_redeep_3}.

% \subsection{Statistical Tests}

% The p-values for the method performance comparisons in \cref{tab:main_results_rq2} all reach the lower bound detectable by the permutation tests, which is 0.0004.

% \begin{table*}[t]
% \centering
% \resizebox{\textwidth}{!}{
% \begin{tabular}{llccccccc}
% \toprule
% Context Comparison & Differentiator & LLaMA 3 8B & LLaMA 2 7B & Gemma 2 9B & Qwen 2.5 7B & Qwen 2.5 3B & Average \\
% \midrule
% Correct $>$ Incorrect & $\Delta\Omega^{(\text{prompt w/ A})}$ vs. $\Delta\Psi$ & 0.0004 & 0.0004 & 0.0004 & 0.0004 & 0.0004 & 0.0004 \\
% & $\Delta\Omega^{(\text{prompt w/o A})}$ vs. $\Delta\Psi$ & 0.0004 & 0.0004 & 0.0007 &0.0004  & 0.0004 & 0.0004 \\
% \midrule
% Correct $>$ Irrelevant & $\Delta\Omega^{(\text{prompt w/ A})}$ vs. $\Delta\Psi$ & 0.0005 & 0.0004 & 0.0004 & 0.0004 & 0.0004 & 0.0004 \\
% & $\Delta\Omega^{(\text{prompt w/o A})}$ vs. $\Delta\Psi$ & 0.0004 & 0.0004 & 0.0004 & 0.0004 & 0.0588 & 0.5839 \\
% \bottomrule
% \end{tabular}
% }
% \caption{p-values from permutation tests (5,000 samples) for the comparisons in \cref{tab:main_results_rq2}.}
% \label{tab:permutation_rq2}
% \end{table*}

\section{Computational Costs}

All experiments were conducted using two A6000 GPUs, with a total compute time of under 1,000 hours.

\end{document}